\documentclass[11pt]{article}

\usepackage[final]{acl}

\usepackage{times}
\usepackage{latexsym}

\usepackage[T1]{fontenc}

\usepackage[utf8]{inputenc}

\usepackage{microtype}

\usepackage{inconsolata}

\usepackage{graphicx}

\title{The Flaw of Averages: \\Measuring Benchmark-Level Distributional Robustness}

\author{
  Arda Uzunoğlu \and
  Tianjian Li \and
  Daniel Khashabi \\
  Department of Computer Science, Johns Hopkins University \\
  \texttt{\{auzunog1,tli104,dkhasha1\}@jhu.edu}
}

\usepackage{mathtools}
\usepackage{amsthm}
\usepackage{amssymb}
\usepackage{wrapfig}
\usepackage{subcaption}
\usepackage{hyperref}
\usepackage{url}
\usepackage[many]{tcolorbox}
\usepackage{graphicx}
\usepackage{amsmath} 
\usepackage{xcolor}
\usepackage{enumitem}
\usepackage{booktabs}
\usepackage{siunitx}
\usepackage{multirow}
\usepackage{needspace}
\usepackage{soul}

\definecolor{llamagreen}{HTML}{00bf63}
\definecolor{qwenpurple}{HTML}{5e17eb}
\definecolor{olmoorange}{HTML}{e76f51}
\definecolor{gemmapink}{HTML}{ff6f91}
\definecolor{phiblue}{HTML}{457b9d}
\tcbuselibrary{listings, breakable, skins, theorems, documentation, hooks}
\usepackage[capitalize,noabbrev]{cleveref}
\theoremstyle{plain}

\theoremstyle{definition}

\theoremstyle{remark}

\newcommand{\method}{{\textsc{Harmony}}}

\usepackage[disable,textsize=tiny]{todonotes}
\usepackage{lineno}

\definecolor{darkblue}{rgb}{0, 0, 0.5}
\hypersetup{colorlinks=true, citecolor=darkblue, linkcolor=darkblue, urlcolor=darkblue}

\begin{document}
\maketitle
\begin{abstract}
Benchmarks are central to measuring progress in language models, but aggregate scores can obscure substantial variation across subdomains, making models appear broadly competent despite concentrated strengths and weaknesses. We study this issue as \emph{benchmark-level distributional robustness}: whether aggregate scores faithfully reflect performance across benchmark subdomains. We operationalize this notion with benchmark \method{}, an entropy-based measure of how uniformly model performance is distributed across subdomains. Measuring \method{} on 19 language model benchmarks across five model families, we find substantial variation in benchmark-level distributional robustness. Low-\method{} benchmarks are more likely to yield aggregate scores that overstate broad competence, whereas high-\method{} benchmarks provide more representative summaries of model capability. Rebalancing benchmarks by pruning overrepresented subdomains to increase \method{} substantially shifts aggregate scores for low-\method{} benchmarks, but leaves high-\method{} benchmarks comparatively stable. For example, while BoolQ remains comparatively stable as \method{} increases, PubMedQA, which evaluates performance in a medically consequential domain, exhibits substantial, often statistically significant, shifts in aggregate accuracy. Together, these findings show that aggregate scores can misrepresent broad competence when performance is unevenly distributed. We therefore recommend reporting benchmark \method{} alongside aggregate accuracy as a diagnostic of benchmark representativeness when interpreting claims about broad model competence.
\end{abstract}

\section{Introduction}

Benchmarks are central to measuring and shaping scientific progress in language models, forming a feedback loop with model development \citep{hardy2024marketinginformationvalueai, yao2025secondhalf}. As benchmarks guide model design, stronger models expose benchmark limitations. In this reciprocal process, it is essential that benchmark evaluation remains representative: aggregate scores should reflect genuine capabilities rather than evaluation artifacts. However, whether these scores are actually representative of broad competence \citep{swayamdipta2020datasetcartographymappingdiagnosing, sainz2023nlpevaluationtroubleneed} has received less attention than the algorithmic advances that benchmarks are used to evaluate \citep{brown2020languagemodelsfewshotlearners, ouyang2022traininglanguagemodelsfollow, deepseekai2025deepseekr1incentivizingreasoningcapability}.

Motivated by this gap, we study evaluation through the lens of what we call \emph{benchmark-level distributional robustness}: whether an aggregate score remains a representative summary of performance across the subdomains a benchmark is intended to cover. This perspective complements prior work on benchmark robustness, which has identified concerns including redundancy \citep{polo2024tinybenchmarksevaluatingllmsfewer, perlitz2024efficientbenchmarkinglanguagemodels}, uneven data distributions \citep{huang2025mathperturbbenchmarkingllmsmath}, and whether benchmark gains are trustworthy \citep{singh2025leaderboardillusion} or support the claims commonly drawn from them \citep{ruan2024observational, heineman2025signalnoiseframeworkreducing}. More specifically, we isolate a failure mode in which aggregate benchmark scores misrepresent broad competence by obscuring concentrated strengths and weaknesses. This perspective is internal to the benchmark itself, complementing both prior work on robustness to distribution shift and worst-group failure \citep{sagawa2020distributionallyrobustneuralnetworks, koh2021wildsbenchmarkinthewilddistribution} and efforts to assess benchmark design more systematically \citep{reuel2024betterbenchassessingaibenchmarks, gebru2021datasheetsdatasets}.

\begin{figure*}
    \centering
    \includegraphics[width=0.98\linewidth,trim=0cm 2.7cm 0cm 2.7cm,clip=true]{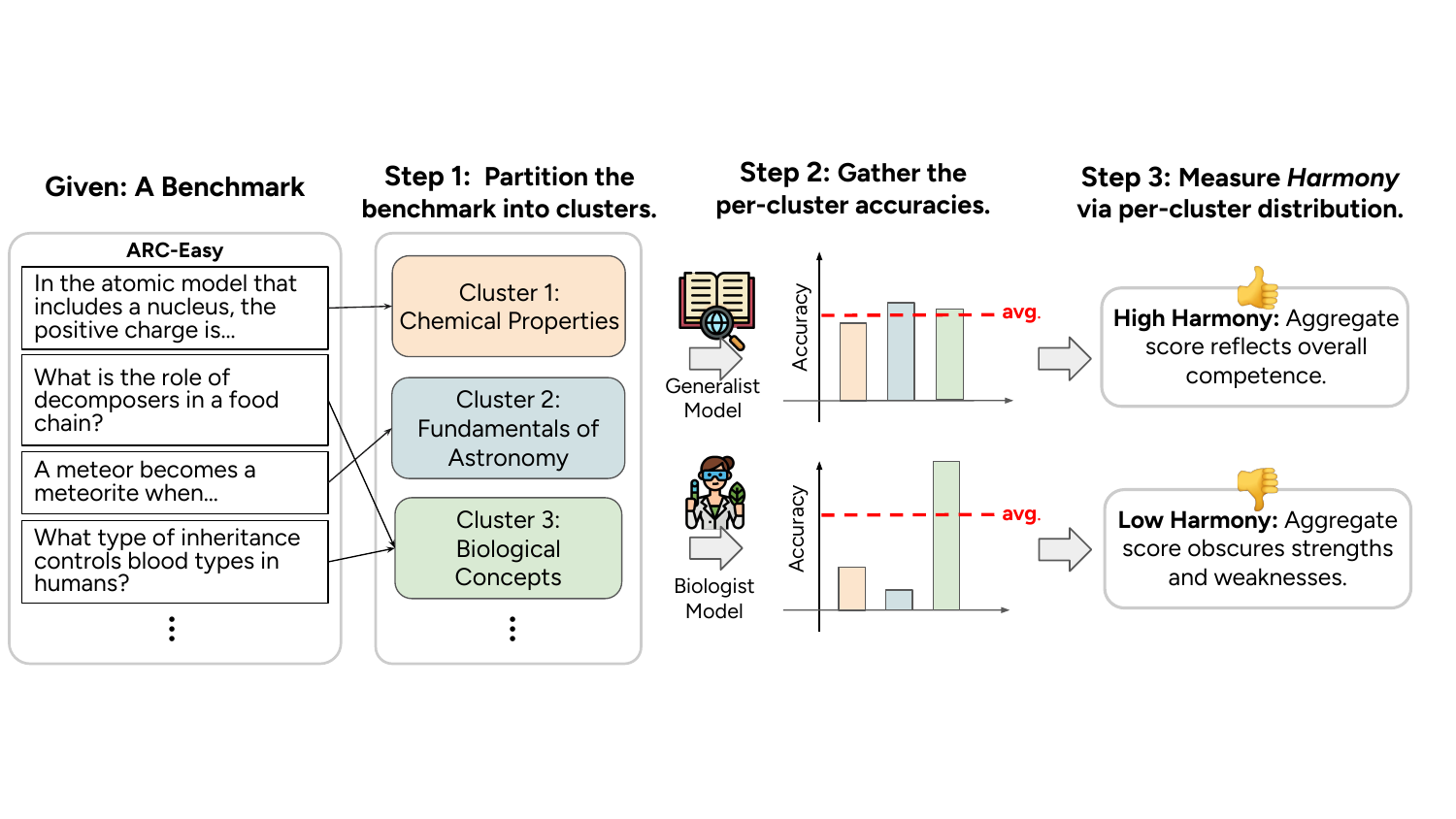}
    \caption{\textbf{Pipeline of evaluating \method{} for a given benchmark.} Step 1: We partition the benchmark into semantic clusters (subdomains). 
    Step 2: We gather each model's performance on every cluster. Step 3: We compute \method{}, the uniformity of the distribution of performance across subdomains. High \method{} indicates that aggregate benchmark scores more faithfully summarize competence across subdomains, whereas low \method{} indicates that aggregate scores may be dominated by only part of the benchmark.}
    \label{fig:teaser}
\end{figure*}

To study \emph{benchmark-level distributional robustness}, we introduce benchmark \method{}, a benchmark-level diagnostic that measures how uniformly a model’s performance is distributed across a benchmark’s subdomains. As illustrated in \autoref{fig:teaser}, we partition each benchmark into semantic subdomains, measure model performance on each, and compute \method{} for each benchmark-model pair. While subdomain-level breakdowns are informative, they do not by themselves quantify whether an aggregate score is representative across subdomains or enable compact comparison across benchmarks and models. High \method{} indicates that aggregate scores are more representative of broad competence across subdomains, whereas low \method{} suggests that aggregate scores may be driven disproportionately by a subset of subdomains. Using \method{}, we analyze a range of benchmarks and models to assess benchmark-level distributional robustness (\S\ref{sec:main:analyses}, \S\ref{sec:beyond}). We recommend reporting benchmark \method{} alongside aggregate accuracy when interpreting benchmark results, as a diagnostic of whether the benchmark’s score is representative across subdomains.

Our contributions are:

\begin{itemize}[leftmargin=1.2em,itemsep=1pt,topsep=1pt]
    \item \textbf{A benchmark-centric perspective on distributional robustness.} We formalize benchmark-level distributional robustness and show that aggregate benchmark scores can systematically misrepresent broad competence when performance is uneven across subdomains.
    \item \textbf{A diagnostic metric of benchmark representativeness across subdomains.} We introduce \method{}, an entropy-based measure of how uniformly performance is distributed across a benchmark’s subdomains, and argue that it should be reported alongside aggregate accuracy as a diagnostic of benchmark representativeness.
    \item \textbf{A large-scale empirical map.} We map 19 MCQA benchmarks and MMLU subtasks into a cross-model \method{} mean-variance plane across 36 models from five families.
\end{itemize}

\section{Benchmark \method{}}
\label{sec:benchmark_harmony}
\subsection{Preliminaries and Notation}
\label{subsec:preliminaries}
Let $\mathcal{B}=\{(x_i,y_i)\}_{i=1}^N$ be a benchmark consisting of $N$ input-output pairs $(x,y)$. Our goal is to understand the \emph{underlying distribution} of $\mathcal{B}$ by inducing a semantic partition $\mathcal{G}=\{A_1,\ldots,A_k\}$ of $\mathcal{B}$, where $A_i \subseteq \mathcal{B}$, $A_i \cap A_j = \emptyset$ for $i \neq j$, and $\bigcup_{i=1}^k A_i = \mathcal{B}$. The partition is guided by a similarity function $\mathcal{S}:\mathcal{X}\times\mathcal{X}\to(0,1]$ that measures the semantic similarity between data points $x_i,x_j\sim\mathcal{B}$. Lastly, let $f$ be a model and let $\Psi(f;A_i)$ denote a measure of performance (e.g., accuracy, pass rate, or judge score) for $f$ computed on a subdomain $A_i \subseteq \mathcal{B}$.

\subsection{\method{}: A Measure of Performance Uniformity}
\label{subsec:harmony}


\paragraph{Intuition.} Consider a biology benchmark spanning microbiology, animal biology, and plant biology. If one subdomain dominates and a model performs well only there, the aggregate score may misleadingly suggest broad biological competence. Even when subdomains are balanced, large performance differences across them can still make the average uninformative. Our goal is therefore to quantify, for each model, whether benchmark performance is broadly distributed across subdomains or concentrated in only part of the benchmark. We capture this with \method{}, a measure of the uniformity of subdomain-level performance. One could instead report the full vector of subdomain accuracies, which is useful descriptively, but that vector does not by itself provide a single measure of whether the aggregate score is representative or support straightforward comparison. Therefore, \method{} is intended as a compact diagnostic. Since \method{} depends jointly on subdomain prevalence and model-specific variation, it is not merely a measure of dataset balance. Instead, it asks whether the aggregate score is representative for that model: high \method{} means that aggregate accuracy reflects subdomain-level performance well, whereas low \method{} means that it does not, whether because of benchmark imbalance (e.g., overrepresented subdomains) or genuine heterogeneity in model capability across tasks. By aggregating \method{} across models, we then assess whether a benchmark reliably supports claims of broad competence.

\paragraph{Formal definition of \method{}.}
Given a partition $\mathcal{G}_f=\{A_i\}_{i=1}^k$, \emph{\method{}} measures how uniformly performance is distributed across the subdomains in this partition. Unlike dispersion statistics such as the variance of subdomain accuracies, which quantify the magnitude of variation across subdomains, \method{} is designed to characterize how performance is distributed across them. For each subdomain $A_i$, let $w_i = \frac{|A_i|}{|\mathcal{B}|}$ denote its relative size and let $\mu=\sum_{i=1}^k w_i \Psi(f;A_i)$ be the weighted mean performance. We convert deviations of $\Psi(f;A_i)$ from $\mu$ into smooth proximity scores via a Gaussian kernel:
\[
K_i \;=\; \exp\!\Big(-\big(\tfrac{\Psi(f;A_i)-\mu}{b}\big)^2\Big),
\]
where $b>0$ is a bandwidth parameter.\footnote{We set $b$ by a robust scale of $\{\Psi(f;A_i)\}$. Let $\tilde a=\mathrm{median}_i\,\Psi(f;A_i)$ and $\mathrm{MAD}=\mathrm{median}_i|\Psi(f;A_i)-\tilde a|$, then $b=\max\{0.02,\;1.4826\cdot\mathrm{MAD}\}$.} We then define \emph{performance masses}
\[
p_i \;=\; \frac{w_i K_i}{\sum_{j=1}^k w_j K_j}, \qquad \sum_{i=1}^k p_i = 1,
\]
and compute \method{} as the normalized Shannon entropy
\[
H(\mathcal{G}_f) \;=\; -\frac{1}{\log k}\sum_{i=1}^k p_i \log\!\big(p_i+\varepsilon\big)\in[0,1],
\]
with a small $\varepsilon=10^{-12}$ for numerical stability. Equivalently,
\[
p_i \propto w_i \exp\!\left(
-\left(\frac{\Psi(f;A_i)-\mu}{b}\right)^2
\right),
\]
which is a weighted softmax over negative squared deviations from the benchmark mean. Subdomains whose performance lies far from the benchmark mean receive exponentially smaller mass, which lowers entropy. Thus, \method{} is high when performance is broadly and evenly distributed across subdomains, and low when performance is concentrated in only a few of them. We defer the further discussion of \method{} design choices to App.~\ref{app:harmony_desing_choices}.

\paragraph{Interpreting \emph{\method{}}.}
Let $\Pi$ be a partitioning rule that maps a benchmark-model pair $(\mathcal{B}, f)$ to a partition $\mathcal{G}_f(\mathcal{B})=\Pi(\mathcal{B};f)$ using $\mathcal{S}$. We define the per-model \method{} of $\mathcal{B}$ as
\[
H_{\mathcal{B}}(f)\;:=\;H\!\big(\mathcal{G}_f(\mathcal{B})\big)\in[0,1].
\]
Given a model set $\mathcal{F}$, we summarize $\mathcal{B}$ by the cross-model mean and variance
\begin{equation}\label{eq:mu_and_sigma}
\begin{aligned}
\mu_H(\mathcal{B}) &= \mathbb{E}_{f\sim\mathcal{F}}\!\left[ H_{\mathcal{B}}(f) \right], \\
\sigma_H^2(\mathcal{B}) &= \mathrm{Var}_{f\sim\mathcal{F}}\!\left( H_{\mathcal{B}}(f) \right).
\end{aligned}
\end{equation}
Since the induced partition depends on the model, benchmark-level distributional robustness is a property of a benchmark relative to a model set $\mathcal{F}$. In particular, higher $\mu_H(\mathcal{B})$ means that aggregate performance is, on average, more representative of subdomain-level performance, while lower $\sigma_H^2(\mathcal{B})$ means that this representativeness is more consistent across models. We therefore interpret benchmarks with high mean \method{} and low variance as more distributionally robust, in the sense that their aggregate scores more reliably summarize broad competence across subdomains (rather than robustness to external shift). However, this notion is comparative rather than absolute: all else equal, $\mathcal{B}_1$ is preferred to $\mathcal{B}_2$ when it achieves higher mean \method{} and lower variance. Accordingly, \method{} should not be interpreted as a measure of benchmark quality overall, but as a diagnostic of one specific aspect of it (i.e. representativeness). Lastly, we note that we use model-specific partitions because benchmark-level distributional robustness concerns whether a benchmark’s aggregate score is representative for a particular model, which depends on the subdomain structure relevant to that model’s behavior. A fixed human-defined partition instead measures uniformity with respect to a prespecified ontology, which may be unavailable, incomplete, or too coarse. By inducing partitions from predictive similarity, we recover the subdomains that govern a model’s performance on the benchmark, so that \method{} evaluates whether aggregate scores are representative of model behavior rather than of an externally imposed taxonomy.

\subsection{Inducing Benchmark Subdomains}
\label{subsec:methodology}
To compute \method{}, we require a semantic partition of the benchmark into subdomains. Because such subdomains are typically not given a priori, we infer them from model-aware semantic similarity. Specifically, we define \textbf{predictive similarity}, a similarity measure between data points based on the divergence of their predictive distributions under a model, and induce $\mathcal{G}_f$ via spectral clustering on the resulting affinity matrix $\mathcal{S}$.

\paragraph{Predictive Similarity.} We define predictive similarity to measure how similarly a model $f$ behaves on two inputs at the level of its distributions. For \(x_i, x_j \sim \mathcal{B}\), the predictive similarity is
\begin{multline}
S(x_i, x_j) =
\exp\!\Bigg(-\tfrac{\tau}{2}\Big[
D_{\mathrm{KL}}\big(\bar{p}_f(x_i)\,\|\,\bar{p}_f(x_j)\big) \\
+\; D_{\mathrm{KL}}\big(\bar{p}_f(x_j)\,\|\,\bar{p}_f(x_i)\big)
\Big]\Bigg).
\label{eq:predictive:similarity}
\end{multline}
where $D_{\mathrm{KL}}(\cdot \,\|\, \cdot)$ denotes the KL divergence, $\tau$ is a scaling factor chosen as the reciprocal of the median symmetric divergence, and the averaged predictive distribution is given by $\bar{p}_f(x_i) \;=\; \frac{1}{T} \sum_{t=1}^{T} p_f(x_i, y_{i}^{<t}),$ with $y_{i}^{<t}$ denoting the ground-truth prefix up to token $t-1$.\footnote{For $t>1$, we condition on the ground-truth answer tokens rather than on the model’s own autoregressive predictions, ensuring that accumulated model errors do not affect the similarity measure.} We defer further discussion of predictive similarity to App.~\ref{sec:pred_sim_app}.

\paragraph{Clustering.} Given the predictive similarity matrix $S\in(0,1]^{N\times N}$, we induce the partition of a benchmark via spectral clustering \citep{NIPS2001_801272ee}. We treat $S$ as a precomputed affinity, form the symmetric normalized Laplacian
$L \;=\; D^{-1/2}(D-S)D^{-1/2}$ with $D \;=\; \mathrm{diag}(S\mathbf{1}),$
compute the $k$ eigenvectors of $L$ associated with its smallest eigenvalues, and apply $k$-means in this spectral embedding to obtain a partition $\mathcal{G}=\{A_1,\ldots,A_k\}$. To choose the number of subsets, we sweep $2 \leq k \leq 20$ and select the value maximizing the silhouette score $s(k)\in[-1,1]$ as a data-driven measure of cluster compactness and separation \citep{ROUSSEEUW198753}. The selected $k$ is not significantly associated with \method{} ($r=0.274$, $p=0.282$; App.~\ref{app:k_dependence}). The induced subdomains are human interpretable, as we assign concise labels to clusters using a bottom-up labeling procedure and provide representative examples in Apps.~\ref{sec:hier_labeling} and~\ref{app:example_clusters}.


Although predictive similarity requires access to model predictive distributions, this is only needed for model-specific partition induction. Given any fixed partition, including model-agnostic partitions induced from embeddings, \method{} requires only subdomain level performance and is therefore applicable to black-box or proprietary models. We validate this setting in App.~\ref{app:minilm_sanity}.

\subsection{Validating the Induced Subdomains}
\label{subsec:empirical_val}

Because \method{} depends on the induced subdomain partition, validating the partitioning step is essential. To this end, we introduce RedundantQA, a synthetic four-domain\footnote{Biology, History, Economics, Popular Culture.} MCQA benchmark with a ground-truth underlying structure. Each item pairs a reference question with two \emph{true-similar} paraphrases (same underlying knowledge) and two \emph{false-similar} distractors (high lexical overlap, different answers). This design cleanly separates semantic from lexical similarity while allowing controlled variation in the underlying data distribution. See Appendix~\ref{sec:redundantqa} for construction and validation details, along with representative examples.
\begin{figure}
  \centering
  \begin{subfigure}[b]{0.48\linewidth}
    \centering
    \caption{RedundantQA}
    \label{fig:validating_methodology_red}
    \includegraphics[width=\linewidth]{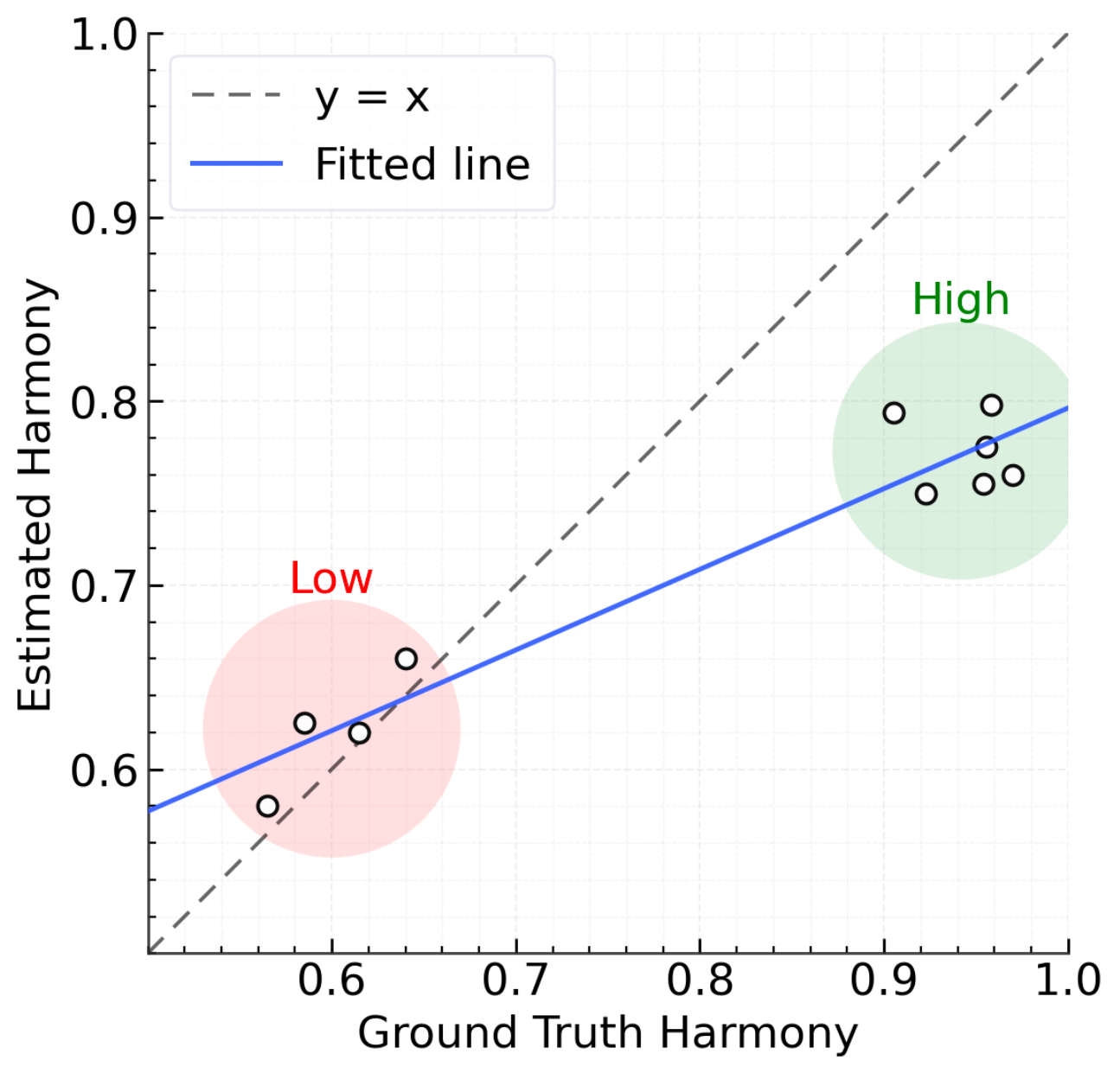}
  \end{subfigure}
  \hfill
  \begin{subfigure}[b]{0.48\linewidth}
    \centering
    \caption{MMLU subtasks.}
    \label{fig:validating_methodology_mmlu}
    \includegraphics[width=\linewidth]{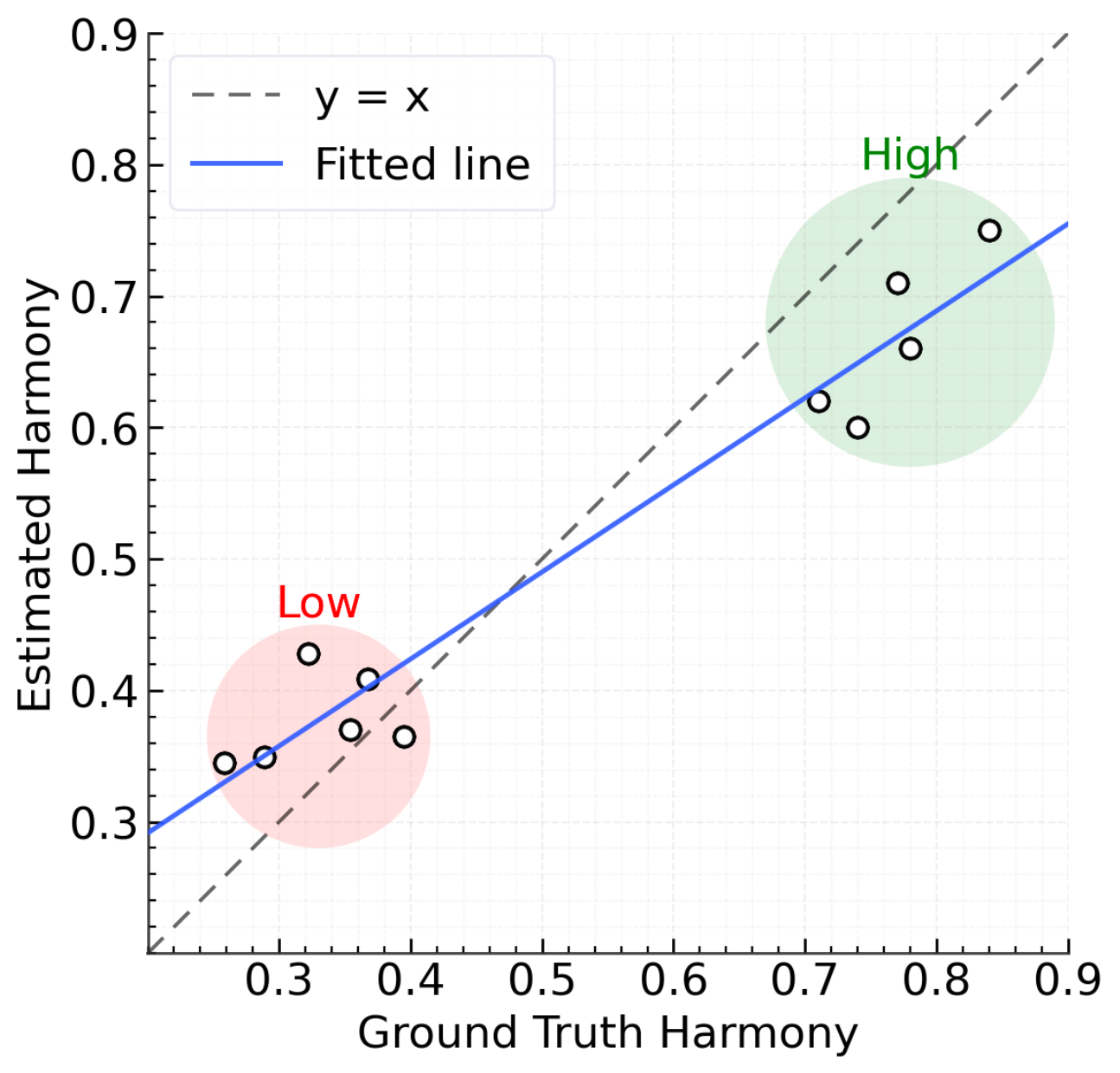}
  \end{subfigure}
  \caption{\textbf{Validation of our approach on (a) RedundantQA and (b) MMLU subtasks.} Estimated \method{} strongly correlates with the ground truth and clearly separates \textcolor{red}{low} from \textcolor{green!75!black}{high} \method{} variants. Each dot represents one variant averaged across five random seeds. Full results are provided in Fig.~\ref{fig:baseline_validations} and~\ref{fig:baseline_validations_mmlu}.}
  \label{fig:validating_methodology}
\end{figure}
We validate both the induced partitions and the resulting \method{} estimates on controlled variants of RedundantQA and on a compilation of MMLU high school subtasks. As shown in Fig.~\ref{fig:validating_methodology}, \method{} estimated by our method exhibits a strong positive correlation with ground-truth \method{} and clearly separates low- and high-\method{} regimes.
\begin{figure*}[!t]
  \centering

  \begin{subfigure}{0.49\linewidth}
    \centering
    \includegraphics[width=\linewidth]{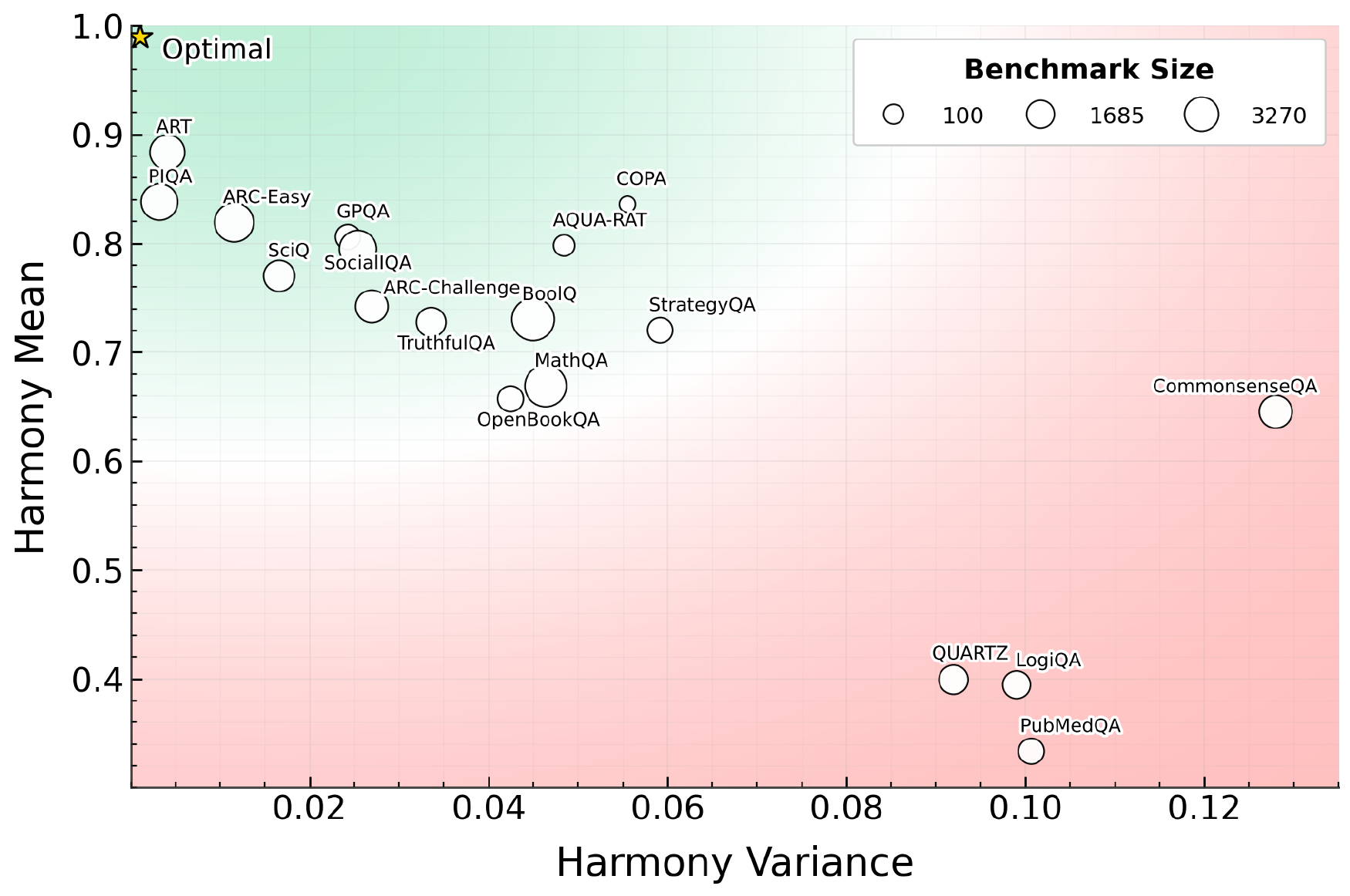}
    \caption{MCQA Benchmarks.}
    \label{fig:non_mmlu_tasks}
  \end{subfigure}
  \begin{subfigure}{0.49\linewidth}
    \centering
    \includegraphics[width=\linewidth]{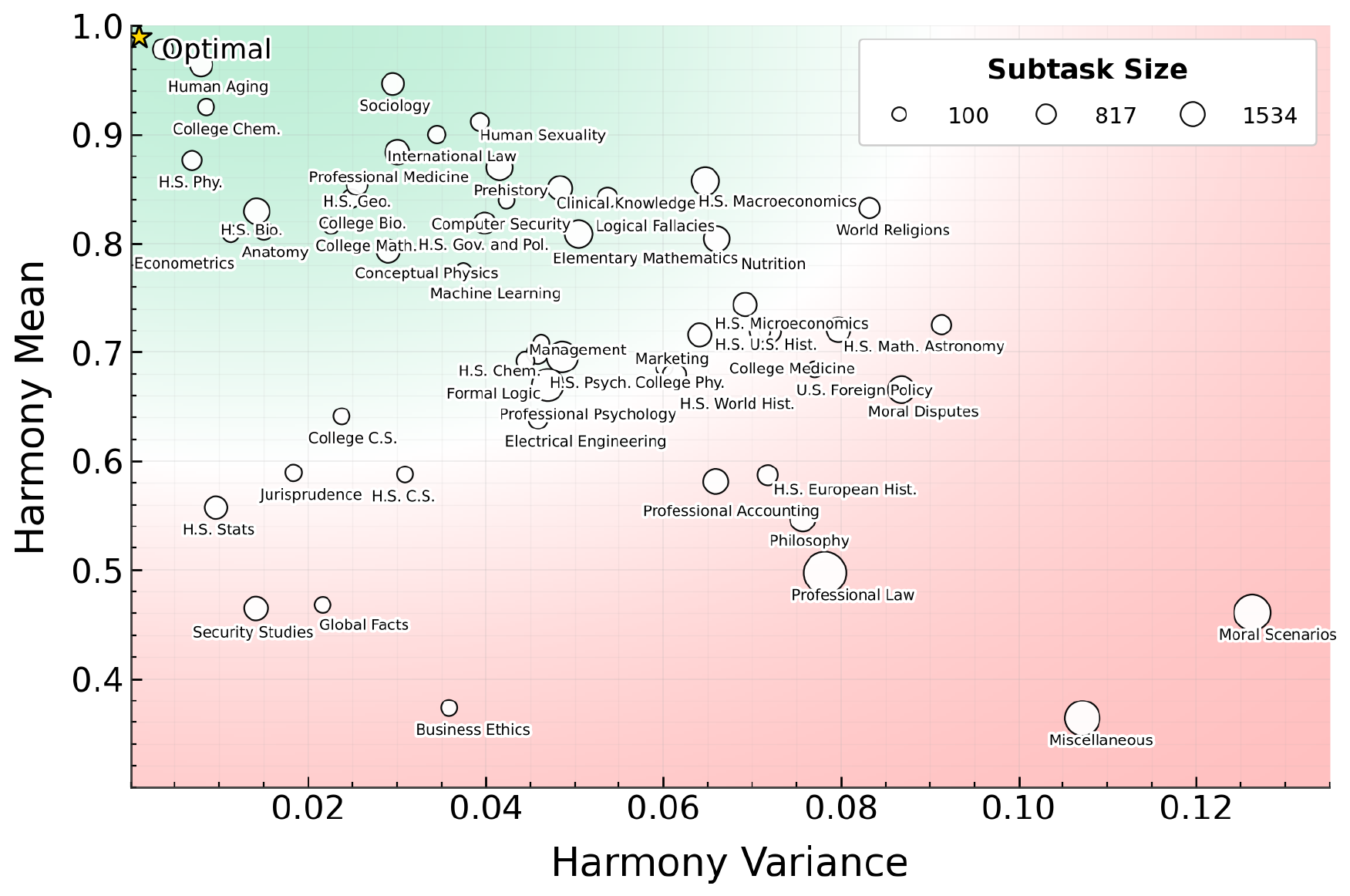}
    \caption{MMLU Subtasks} \label{fig:benchmarks_overview}
  \end{subfigure}
  \caption{\textbf{Mean-variance plane for \method{} across (a) MCQA benchmarks and (b) MMLU subtasks.}
    Each point represents a benchmark or a subtask plotted by the \method{} mean ($\mu_H(\mathcal{B})$) and variance ($\sigma_H^2(\mathcal{B})$) over 36 models. Upper-left (high mean, low variance) indicates \textcolor{green!75!black}{higher} \emph{benchmark-level distributional robustness}; rightward (higher variance) and downward (lower mean) shifts signal \textcolor{red}{diminished} robustness. In (a), \method{} mean and variance show a strong negative Pearson correlation ($r=-0.785$ with $p=0.0001$), while in (b) the correlation is more moderate ($r = -0.404$ with $p = 0.0025$).}
  \label{fig:combined_benchmark_fig}
  \vspace{-1em}
\end{figure*}
We further validate predictive similarity along three axes and defer the details to Appendix~\ref{sec:pred_sim_app}: (i) discrimination of semantic vs.\ lexical similarity (App.~\ref{subsec:sem_vs_lex}), (ii) recovery of ground-truth domains on RedundantQA and MMLU (App.~\ref{subsec:recover_dom}), and (iii) fidelity of \method{} estimates under distributional shifts (App.~\ref{subsec:capture_quality}).

\section{Main Analysis: How Harmonious Are the Benchmarks?}
\label{sec:main:analyses}

We now turn to the central question we investigate: how robustly do benchmarks evaluate model competence across subdomains? To answer this, we compute \method{} for each benchmark-model pair as described in \S\ref{sec:benchmark_harmony}, and then aggregate these scores across models to place each benchmark in the mean-variance plane defined by $(\mu_H(\mathcal{B}), \sigma_H^2(\mathcal{B}))$ in Eq.~\ref{eq:mu_and_sigma}. This representation lets us compare benchmarks by both their average distributional robustness and the stability of that robustness across models. We describe the experimental setup (\S\ref{subsec:exp_setup}) and then interpret the resulting benchmark landscape (\S\ref{subsec:mapping}).

\subsection{Experimental Setup}
\label{subsec:exp_setup}

We conduct evaluations using \texttt{lm-evaluation-harness}\footnote{https://github.com/EleutherAI/lm-evaluation-harness} across a wide range of model sizes from five families: \texttt{Llama 3} \citep{grattafiori2024llama}, \texttt{Qwen3} \citep{yang2025qwen3}, \texttt{Gemma 3} \citep{gemmateam2025gemma3technicalreport}, \texttt{Phi-3} \citep{abdin2024phi3technicalreporthighly}, and \texttt{OLMo 2} \citep{olmo20252olmo2furious} (see App.~\ref{sec:model_list} for the full list). Our setup spans 19 MCQA benchmarks covering a broad range of capabilities:

\begin{itemize}[left=0pt, itemsep=1pt]
  \item \textbf{Reasoning}: 
  ARC \citep{clark2018think}, ART \citep{Bhagavatula2020Abductive}, BoolQ \citep{clark2019boolq}, CommonsenseQA \citep{talmor2018commonsenseqa}, COPA \citep{roemmele2011choice}, LogiQA \citep{liu2020logiqachallengedatasetmachine}, PIQA \citep{Bisk2020}, QUARTZ \citep{quartz}, SocialIQA \citep{sap-etal-2019-social}, StrategyQA \citep{geva2021didaristotleuselaptop}.
  \item \textbf{Math Problem-Solving}: 
  AQUA-RAT \citep{ling2017program}, MathQA \citep{amini2019mathqainterpretablemathword}.
  \item \textbf{World Knowledge}: 
  GPQA \citep{rein2023gpqa}, MMLU \citep{hendrycks2020measuring}, OpenBookQA \citep{mihaylov2018suit}, PubMedQA \citep{jin2019pubmedqa}, SciQ \citep{SciQ}.
  \item \textbf{Truthfulness}: 
  TruthfulQA \citep{lin2021truthfulqa}.
\end{itemize}

For evaluation, we use average token-level log-likelihood over answer options, selecting the option with the highest average log-probability. We follow each benchmark’s evaluation protocol as implemented in the harness, use zero-shot evaluation by default, and report accuracy. We focus on MCQA benchmarks because (i) the discrete label space $\mathcal{Y}$ yields unambiguous ground truth and exact accuracy,\footnote{In free-response settings, multiple plausible intermediate steps can lead to the same answer, complicating evaluation.} and (ii) evaluation is automatic and does not rely on a judge, thereby avoiding the grading variance common in free-form scoring \citep{chiang2024chatbot}. At the same time, \method{} is not limited to MCQA: App.~\ref{app:beyond_mcqa} demonstrates its application to open-ended mathematical reasoning and instruction following evaluation.
\begin{figure*}[t]
    \centering
    \includegraphics[width=0.97\linewidth]{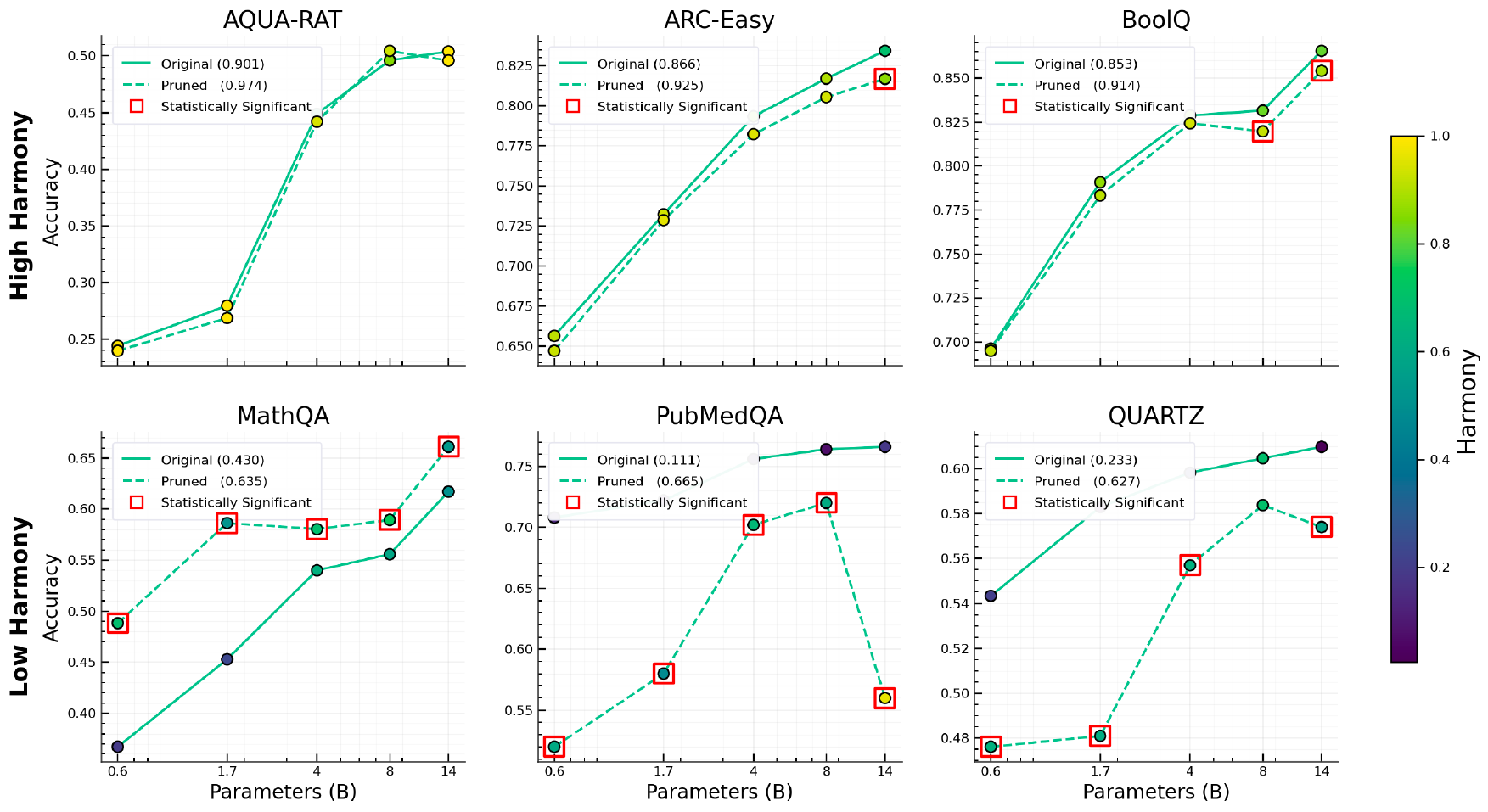}
    \caption{\textbf{Balancing benchmarks via pruning.} We remove overly similar items. The \emph{top row} shows more harmonious benchmarks, where accuracy remains stable as \method{} increases. The \emph{bottom row} shows less harmonious benchmarks, where \method{} rises and accuracy shifts significantly. Model-averaged \method{} values for the original and pruned benchmarks are reported in parentheses in the legends.}
    \label{fig:pruning}
\end{figure*}
\subsection{Benchmark Landscape in the \method{} Mean-Variance Plane}
\label{subsec:mapping}
We analyze the benchmarks with \method{} by inducing model-specific partitions, computing a \method{} score for each benchmark-model pair, and then placing each benchmark in a two-dimensional plane whose axes are the mean and variance of these per-model scores.

Fig.~\ref{fig:non_mmlu_tasks} and Fig.~\ref{fig:benchmarks_overview} position, respectively, the benchmark suite and the MMLU subtasks in the cross-model mean-variance plane of \method{}. The vertical axis is \(\mu_H(\mathcal{B})\), the average uniformity of performance across subdomains, and the horizontal axis is \(\sigma_H^2(\mathcal{B})\), the cross-model variability of that property. Benchmarks in the \emph{upper-left} region (high mean, low variance) tend to provide more representative and more stable aggregate evaluations across models. By contrast, low mean indicates that performance is skewed across subdomains on average, while high variance indicates that this robustness is model-dependent. Thus, upward and leftward directions in the plane correspond to more robust evaluation, whereas downward and rightward movement signals greater concentration of performance on a few subdomains and less stable conclusions across models. See App.~\ref{app:bench_size_confound},~\ref{app:prun_rat_confound}, and~\ref{app:k_dependence} for a confounder analysis covering benchmark size, pruning-rate sensitivity, and dependence on the number of clusters $k$. App.~\ref{app:minilm_sanity} further shows that the MCQA benchmark landscape is qualitatively and quantitatively similar when using fixed model-agnostic MiniLM partitions.

\section{Further Analyses}
\label{sec:beyond} 

In this section, we pursue two controlled analyses. First, we show how less harmonious benchmarks can distort model evaluation by making aggregate scores less representative of performance across subdomains. Second, we examine whether this fragility is especially pronounced for larger models or for models trained on more tokens.
\begin{figure*}[ht]
    \centering
    \includegraphics[width=\linewidth]{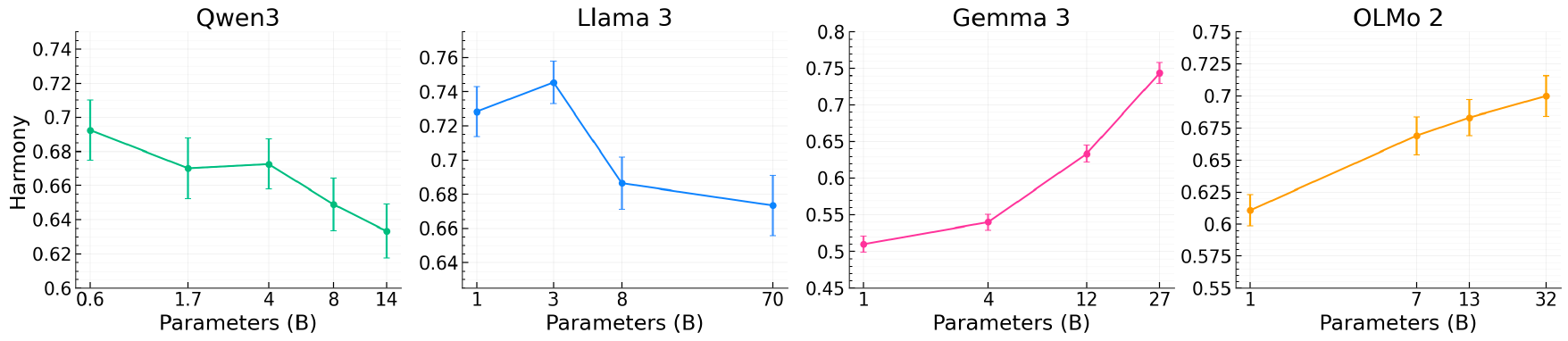}
    \caption{\textbf{Model size vs.\ \method{}.} Scaling trends are \emph{family-specific}: Qwen and Llama show negative correlations, while Gemma and OLMo show positive correlations (larger models perform more uniformly). Thus, parameter count alone is not predictive of performance uniformity. Y-axis shows each model’s average \method{} over all benchmarks.}
    \label{fig:parameter_sl}
    \vspace{-1em}
\end{figure*}
\subsection{Effects of Benchmark Rebalancing}
\label{sec:pruning}
\label{subsec:balancing_subsets}
We now study whether less harmonious benchmarks are more vulnerable to unrepresentative aggregate evaluation. To do so, we prune benchmarks using predictive similarity to remove overly similar items. The pruning ratio is set inversely proportional to benchmark \method{}, so that high-\method{} benchmarks receive minimal pruning while low-\method{} benchmarks are pruned more aggressively.\footnote{Specifically, we use the formula \(p=\mathrm{clip}_{[0.05,0.5]}\!\big(0.05+(0.5-0.05)\big(\tfrac{1-\mathrm{clip}(H;\,0.1,\,1)}{1-0.1}\big)^{1.5}\big)\).} This procedure reduces benchmark skewness and lets us examine how aggregate accuracy changes under a rebalancing intervention that typically increases \method{}.

As shown in Fig.~\ref{fig:pruning}, accuracies on high-\method{} benchmarks remain stable under pruning, with changes that are not statistically significant despite increased \method{}. By contrast, low-\method{} benchmarks are fragile: pruning substantially increases \method{} and is accompanied by statistically significant accuracy shifts (see App.~\ref{sec:details_of_stat_sig}). Under this rebalancing intervention, per-subdomain accuracies often tighten around the benchmark mean, making the aggregate score a more representative summary of performance across subdomains. Overall, these results suggest that low-\method{} benchmarks are more vulnerable to aggregate evaluations that are less representative of broad competence, whereas high-\method{} benchmarks tend to be more stable under pruning. While Fig.~\ref{fig:pruning} shows this pattern for the Qwen3 family, App.~\ref{sec:extended_balancing} provides the full results for other model families.

\subsection{Effects of Model Scale and Token Budget}
\label{subsec:subsec6.2}
We next examine how \method{} varies with model size and pre-training token budget.

\paragraph{Model parameters.} We find that the relationship between model parameter count and \method{} is \textbf{family-specific}. As shown in Fig.~\ref{fig:parameter_sl}, within-family comparisons reveal a negative correlation for the Qwen and Llama families, indicating that larger models in these families concentrate performance more on a few subdomains. In contrast, the Gemma and OLMo families exhibit a positive correlation between model size and \method{}, with larger models distributing performance more evenly across benchmark subdomains. Parameter count alone is therefore not a reliable predictor of performance uniformity.

\begin{figure}
    \centering
    \includegraphics[width=\linewidth]{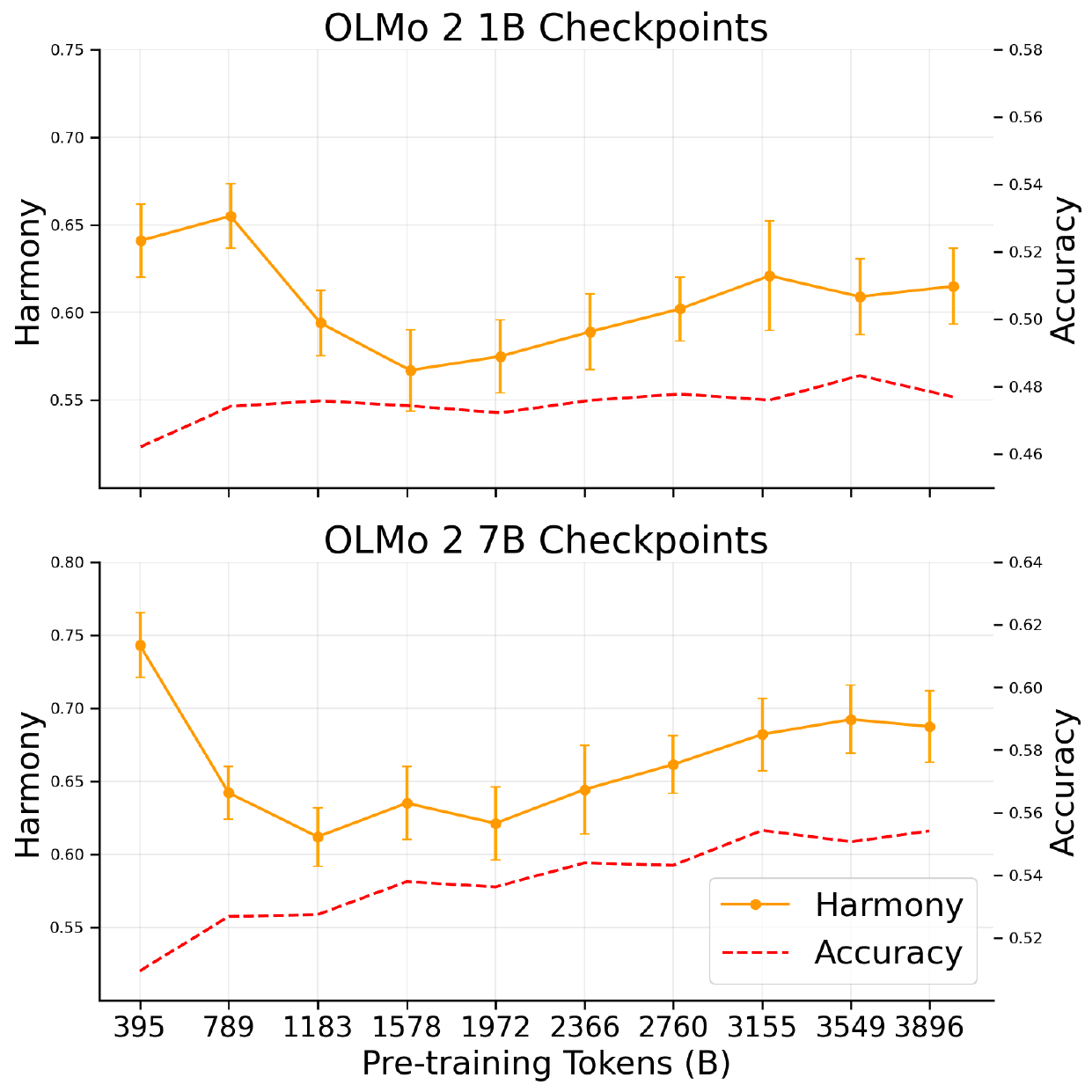}
    \caption{\textbf{Pre-training tokens vs.\ \method{}.} For OLMo 2 1B/7B, \method{} dips then steadily rises with more pre-training tokens while aggregate accuracy improves slightly.}
    \label{fig:token_sl}
    \vspace{-1em}
\end{figure}
\paragraph{Pre-training tokens.} Under a fixed architecture, increasing the pre-training token budget leads to a rise in \method{} after an initial dip. We examine this by tracking OLMo2 1B and OLMo2 7B across increasing token budgets. As shown in Fig.~\ref{fig:token_sl}, \method{} dips early and then rises steadily, while aggregate accuracy changes only slightly across checkpoints. Thus, the distribution of performance becomes more uniform even when the aggregate score remains nearly unchanged. Additional pre-training is therefore associated with a more even spread of performance across subdomains.

We emphasize that the trends reported here are empirical rather than causal, and leave the mechanisms underlying them to future work.

\section{Related Work}
\paragraph{Distributional Robustness.}
A large literature studies robustness to changes in the data distribution, spanning distributionally robust optimization, worst-group and group-shift robustness, invariant learning, and domain generalization \citep{duchi2020, sagawa2020distributionallyrobustneuralnetworks, arjovsky2020, koh2021wildsbenchmarkinthewilddistribution}. Across these lines of work, a common concern is that strong average performance can mask fragile behavior on minority groups, under reweightings of the training distribution, or after shifts from train to test environments \citep{duchi2020, sagawa2020distributionallyrobustneuralnetworks, recht2019imagenetclassifiersgeneralizeimagenet}. Our work is adjacent in spirit to this literature, but differs in its object of analysis. Prior work typically asks whether a model remains reliable under external distribution shift \citep{sagawa2020distributionallyrobustneuralnetworks, arjovsky2020, koh2021wildsbenchmarkinthewilddistribution}. By contrast, we ask whether a benchmark's aggregate score is representative of performance across its internal subdomains. In this sense, our proposal is a diagnostic for the validity of aggregate benchmark summaries. Hence, our perspective is aligned with evidence that benchmark conclusions can be sensitive to modest changes in evaluation data and test-set construction \citep{recht2019imagenetclassifiersgeneralizeimagenet, miller2020accuracy}.

\paragraph{Assessing Benchmark Robustness.}
Beyond proposing new tasks, a growing body of work examines the robustness and reliability of benchmarks themselves. One line of research studies how evaluation sets behave under dynamic or adversarial conditions, including dynamic benchmarks \citep{kiela2021dynabenchrethinkingbenchmarkingnlp, chiang2024chatbot} and adversarially constructed or perturbed versions \citep{nie2020adversarialnlinewbenchmark, croce2021robustbenchstandardizedadversarialrobustness}. Closely related are concerns about contamination and overfitting to public benchmarks \citep{deng2024investigatingdatacontaminationmodern, golchin2024timetravelllmstracing, roberts2023datacontaminationlenstime, dong-etal-2024-generalization}, as well as redundancy, annotation artifacts, and label errors that can distort benchmark conclusions \citep{northcutt2021pervasivelabelerrorstest, gema2025mmlu, polo2024tinybenchmarksevaluatingllmsfewer, perlitz2024efficientbenchmarkinglanguagemodels}. A complementary line of work develops frameworks and documentation practices for systematic benchmark assessment \citep{reuel2024betterbenchassessingaibenchmarks, mazumder2023dataperfbenchmarksdatacentricai, gebru2021datasheetsdatasets}, while work on measurement validity asks whether benchmark scores support the inferential claims drawn from them \citep{Yin2019UnderstandingTE, chouldechova2024sharedstandardvalidmeasurement, salaudeen2025measurementmeaningvaliditycenteredframework}. Other studies investigate whether benchmark gains are trustworthy, what reported scores actually capture, and how reliably comparative rankings can be stabilized \citep{Ott_2022, dehghani2021benchmarklottery, singh2025leaderboardillusion, madaan2024quantifyingvarianceevaluationbenchmarks, evalarena, heineman2025signalnoiseframeworkreducing}. Our work contributes to this literature by foregrounding a benchmark-level distributional perspective. Instead of treating a benchmark score as a single scalar summary, we ask whether that summary is actually representative of performance across the benchmark’s internal subdomains.

\paragraph{Distributional Approaches to Efficient Evaluation.}
A separate line of work uses distributional structure in benchmarks to reduce evaluation cost while preserving comparative validity. Examples include adaptive reliability metrics \citep{perlitz-etal-2024-efficient}, Item Response Theory based methods for modeling item informativeness \citep{rodriguez-etal-2021-evaluation, Tatsuoka1971StatisticalTO}, constructing compact but faithful benchmark subsets \citep{polo2024tinybenchmarksevaluatingllmsfewer}, and usable-information based analyses of evaluative signal \citep{ethayarajh2022understandingdatasetdifficultymathcalvusable}. Although these methods differ in technique, they share the goal of exploiting benchmark structure to obtain cheaper evaluations that preserve rankings or other comparative conclusions. In contrast, we do not aim to compress evaluation, but to test whether a benchmark's aggregate score is itself a representative summary of broad competence across subdomains.

\section{Conclusion}
We introduce \method{}, an entropy-based measure of how uniformly performance is distributed across a benchmark’s subdomains, and use it to study \emph{benchmark-level distributional robustness}. Mapping 19 MCQA benchmarks across five model families in the \method{} mean-variance plane reveals substantial variation in how reliably aggregate scores reflect broad competence across subdomains. Benchmarks with high mean \method{} and low cross-model variance yield more representative and stable evaluations, whereas low mean or high variance signal that aggregate conclusions may be fragile or model-dependent.

Our controlled analysis shows that higher-\method{} benchmark variants tend to yield more stable aggregate evaluation. We further find that scaling trends in \method{} are family-specific and parameter count alone does not reliably predict performance uniformity, although larger pre-training budgets are associated with more uniform performance on average. Overall, our results show that aggregate benchmark scores can misrepresent broad competence when performance is uneven across subdomains. We therefore recommend reporting \method{} alongside accuracy to support claims about broad model competence. We view \method{} as complementary to, rather than a substitute for, subdomain-level analysis, as subdomain-level results reveal where strengths and weaknesses lie, while \method{} summarizes whether aggregate benchmark performance is broadly representative across them.

\section*{Ethical Considerations} 
This work evaluates publicly available benchmarks and introduces \textit{RedundantQA}, a synthetic dataset generated from author-written seeds and LLM generations with no human subjects, personal data, or sensitive attributes collected. All examples were screened to avoid offensive content and verbatim copyrighted material. We respect the licenses of all benchmarks and models used. We will release our code and RedundantQA with documentation of construction, intended use, and limitations.

\method{} is intended to complement (and \emph{not replace}) standard accuracy and robustness analyses. Potential risks include misinterpretation of \method{} or benchmark rebalancing to mask undesired failure modes. We therefore report both \method{} and accuracy and encourage transparent, multi-metric evaluation. Our experiments rely on inference with open-source models. The authors declare no conflicts of interest or external sponsorship that could bias this work.


\section*{Limitations}

Our study has several limitations. First, we focus on multiple-choice benchmarks, where accuracy and predictive distributions are easy to compute, although we additionally validate \method{} on two non-MCQA tasks. Free-form generation requires reliable scoring and may introduce judge-related uncertainty. Second, \method{} depends on the induced subdomain partition. Although we validate predictive similarity and include model-agnostic sanity checks, the clusters may still be sensitive to the similarity metric, clustering procedure, and choice of \(k\). Third, \method{} is a scalar diagnostic rather than a complete measure of benchmark quality: high \method{} does not rule out contamination, annotation artifacts, narrow coverage, or real-world mismatch, and low \method{} does not make a benchmark unusable. Finally, our empirical findings are conditioned on the benchmarks and open model families studied, and the pruning experiments should be viewed as diagnostic interventions rather than prescriptions for benchmark construction.

\section*{Acknowledgments}
AU is supported by JHU PURA (the Provost’s Undergraduate Research Award) and Pistritto Research Fellowship. DK is in part supported by Defense Advanced Research Projects Agency (DARPA) under Contract No. HR001125C0304, ONR grant (N0001424-1-2089) and JHU Provost Discovery Award (2025–2027). Any opinions, findings and conclusions or recommendations expressed in this material are those of the author(s) and do not necessarily reflect the views of DARPA. We acknowledge the use of computational resources on the Johns Hopkins Data Science and AI Institute (DSAI) cluster.


\bibliography{custom}

\appendix

\section{RedundantQA}
\label{sec:redundantqa}
To rigorously evaluate the discriminative power of similarity metrics, we construct RedundantQA, a controlled benchmark designed to disentangle genuine semantic similarity from superficial lexical overlap. Each set in RedundantQA consists of a reference question accompanied by two \textit{true-similar} and two \textit{false-similar} questions. The true-similar questions are paraphrases that evaluate the same underlying knowledge as the reference, while differing in surface form.\footnote{E.g., variations in vocabulary, syntax, or phrasing.} In contrast, the false-similar questions exhibit high lexical similarity to the reference but target distinct conceptual content. This design ensures that strong similarity metrics must go beyond surface-level cues, rewarding semantic alignment while ignoring spurious correlations.

In this section, we detail the construction (\ref{subsec:gen_redundantqa}) and validation (\ref{subsec:val_redundantqa}) of RedundantQA, and showcase examples (\ref{subsec:examples_redundantqa}).

\subsection{Construction}
\label{subsec:gen_redundantqa}
We construct RedundantQA through a two‐phase pipeline followed by strict validation: \textbf{(i) Seed Set Selection.}  We begin by manually authoring three high‐quality reference questions across four domains (Biology, Economics, Popular Culture, History). For each reference question, we also craft two paraphrases that target the same underlying knowledge (\emph{true‐similar}) and two distractors that share surface tokens but probe different concepts (\emph{false‐similar}). \textbf{(ii) Generative Expansion.} Using the seed sets as in-context learning examples, we prompt \texttt{Gemini-2.0-flash} \citep{deepmind2023gemini} to generate 100 sets that consist of one reference question, two true-similar questions, and two false-similar questions for each domain. For different domains, we use a fixed template (Listing \ref{box:gen-prompt}) with domain‐specific examples. This pipeline yields a large, automatically generated candidate pool.

\begin{tcolorbox}[
    breakable,
    title={Prompt for Generating RedundantQA (Biology)},
    colback=gray!5!white,
    colframe=black!75,
    fonttitle=\bfseries,
    label={box:gen-prompt}
]
\tiny
Come up with question sets. Each set must contain:
\begin{itemize}
    \item A reference question,
    \item Two same-meaning questions: These should require the same factual answer and test the same biological concept as the reference question, but they should use different wording, phrasing styles, and sentence structures.
    \item Two distractor questions: These should look superficially very similar to the reference question but evaluate a different knowledge or skill with different answers than the reference question.
\end{itemize}

\textbf{Notes:}
\begin{itemize}
    \item Focus on the domain of biological knowledge.
    \item Same-meaning questions should preserve deep semantic equivalence but vary stylistically. These must have the same answer.
    \item Distractor questions should maximize shallow textual similarity (e.g., shared nouns, verbs, syntactic patterns) while changing the underlying meaning. So, distractor questions should trick an incapable similarity measure into thinking they are similar.
\end{itemize}

\textbf{Examples:}

\textbf{Set 1}
\begin{itemize}
    \item \textit{Reference Question:} What organ pumps blood throughout the human body?
    \item \textit{Same-meaning Question 1:} Which organ circulates blood to deliver oxygen and nutrients?
    \item \textit{Same-meaning Question 2:} What body system structure maintains blood flow across the body?
    \item \textit{Distracting Question 1:} What organ removes carbon dioxide from the blood?
    \item \textit{Distracting Question 2:} What organ transports nutrients through the blood?
\end{itemize}

\textbf{Set 2}
\begin{itemize}
    \item \textit{Reference Question:} What process converts glucose into energy in cells?
    \item \textit{Same-meaning Question 1:} Which process produces ATP from sugar molecules?
    \item \textit{Same-meaning Question 2:} What pathway transforms glucose into usable cellular energy?
    \item \textit{Distracting Question 1:} What process stores glucose in cells?
    \item \textit{Distracting Question 2:} What process breaks down proteins for energy?
\end{itemize}

\textbf{Set 3}
\begin{itemize}
    \item \textit{Reference Question:} What type of blood vessel carries blood away from the heart?
    \item \textit{Same-meaning Question 1:} Which vessels transport oxygenated blood from the heart?
    \item \textit{Same-meaning Question 2:} What structures move blood outward from the heart?
    \item \textit{Distracting Question 1:} What type of blood vessel brings blood to the heart?
    \item \textit{Distracting Question 2:} What blood vessel type filters blood in the kidneys?
\end{itemize}
\end{tcolorbox}

\subsection{Validation}
\label{subsec:val_redundantqa}
We validate each set generated by \texttt{Gemini-2.0-flash} through a two‐stage pipeline: (a) an automated and simple consistency check using \texttt{Gemini-2.0-flash} to confirm that true‐similar paraphrases produce identical answers while false‐similar distractors yield divergent ones (using Listing \ref{box:valid-prompt}); and (b) a manual review by expert annotators to correct any misclassifications, formatting issues, or errors introduced during automated filtering. After the validation step, we obtain 71, 39, 72, and 73 sets from Biology, Economics, Culture, and History domains respectively, with each set consisting of one reference question, two true-similar questions, and two false-similar questions.

This procedure yields a benchmark in which effective similarity metrics must discriminate semantic equivalence from mere lexical coincidence.
\begin{tcolorbox}[
    float=htb,
    title={Prompt for Validating RedundantQA},
    colback=gray!5!white,
    colframe=black!75,
    fonttitle=\bfseries,
    label={box:valid-prompt}
]
\small
Do the following questions have the same answer? Output only yes or no. \\
Question 1: REFERENCE\_QUESTION \\
Question 2: TRUE\_SIM\_1
\end{tcolorbox}

\subsection{Examples}
\label{subsec:examples_redundantqa}
We provide examples from RedundantQA across all four domains in Table \ref{tab:examples_redundantqa}.
\begin{table*}[ht]
\centering
\resizebox{\textwidth}{!}{
\scriptsize
\begin{tabular}{|p{2.7cm}|p{4.2cm}|p{2.9cm}|p{3.1cm}|}
\hline
\textbf{Biology} & \textbf{Economics} & \textbf{History} & \textbf{Popular Culture} \\
\hline
\begin{tabular}[t]{@{}l@{}}
\textbf{Reference} \\
What process converts\\ glucose into energy in cells? \\
A: Cellular respiration \\
B: Photosynthesis \\
C: Osmosis \\
D: Transcription \\[6pt]
\textbf{True Similar} \\
Which process produces\\ ATP from sugar molecules? \\
A: Cellular respiration \\
B: Photosynthesis \\
C: Osmosis \\
D: Transcription \\[6pt]
\textbf{False Similar} \\
What process stores\\ glucose in cells? \\
A: Glycogenolysis \\
B: Gluconeogenesis \\
C: Glycogenesis \\
D: Glycolysis
\end{tabular} &
\begin{tabular}[t]{@{}l@{}}
\textbf{Reference} \\
How does increased government\\ spending affect aggregate demand? \\
A: Increases it. \\
B: Decreases it. \\
C: Has no effect. \\
D: Only affects aggregate supply. \\[6pt]
\textbf{True Similar} \\
What happens to total demand in economy\\ when the government increase its spending? \\
A: Increases it. \\
B: Decreases it. \\
C: Has no effect. \\
D: Only affects aggregate supply. \\[6pt]
\textbf{False Similar} \\
How does increased government\\ spending affect government debt? \\
A: Increases it. \\
B: Decreases it. \\
C: Has no effect. \\
D: Only affects short-term debt.
\end{tabular} &
\begin{tabular}[t]{@{}l@{}}
\textbf{Reference} \\
Who was the first president\\ of the United States? \\
A: George Washington \\
B: Abraham Lincoln \\
C: Thomas Jefferson \\
D: John Adams \\[6pt]
\textbf{True Similar} \\
Who assumed leadership as\\ America's first head of state? \\
A: George Washington \\
B: Abraham Lincoln \\
C: Thomas Jefferson \\
D: John Adams \\[6pt]
\textbf{False Similar} \\
Who was the first vice\\ president of the United States? \\
A: John Adams\\
B: Thomas Jefferson\\
C: Alexander Hamilton\\
D: James Madison
\end{tabular} &
\begin{tabular}[t]{@{}l@{}}
\textbf{Reference} \\
Who played Iron Man in the\\ Marvel Cinematic Universe? \\
A: Robert Downey Jr.\\
B: Chris Evans\\
C: Hugh Jackman\\
D: Tobey Maguire\\[6pt]
\textbf{True Similar} \\
Which actor portrayed\\ Tony Stark in the MCU? \\
A: Robert Downey Jr.\\
B: Chris Evans\\
C: Hugh Jackman\\
D: Tobey Maguire\\[6pt]
\textbf{False Similar} \\
Who played Captain America in\\ the Marvel Cinematic Universe? \\
A: Chris Evans\\
B: Chris Pratt\\
C: John Krasinski\\
D: Matt Damon
\end{tabular} \\
\hline
\end{tabular}}
\caption{Example sets across all domains in RedundantQA.}
\label{tab:examples_redundantqa}
\end{table*}

\section{Predictive Similarity}
\label{sec:pred_sim_app}

\subsection{Alternative Baselines}
\label{subsec:alt_baselines}
To better situate our predictive similarity metric relative to existing work, we first position it against a diverse set of alternative similarity measures from the literature. These baselines span surface-form (n-gram), embedding-based, gradient-based, model-output-based similarities that are commonly used in prior work. In contrast, to the best of our knowledge, predictive similarity is the only metric that explicitly leverages the full probability distributions produced by the models under evaluation.

In this section, we describe the alternative baselines compared against predictive similarity across a range of controlled settings.

\paragraph{Bigram.}
We compute an $n$-gram-overlap Jaccard similarity matrix. For each text $x_i$, we lowercase and split on whitespace, then form the set $G_i$ of contiguous bigrams. The pairwise similarity is $S_{ij} = \dfrac{|G_i \cap G_j|}{|G_i \cup G_j|}$
with $S_{ii}=1$ for all $i$.\footnote{Note that if a text has fewer than $n$ tokens, its $n$-gram set is empty. In such a case, the pairwise similarity is set to be 1. However, this is not observed in practice.} The resulting $S \in [0,1]^{N\times N}$ is symmetric and measures surface-form overlap.

\paragraph{BERTScore F1.} We use BERTScore \citep{zhang2020bertscoreevaluatingtextgeneration} to measure semantic similarity between pairs of texts by comparing their contextualized token embeddings. Tokens are greedily matched via cosine similarity to compute precision and recall, and the final sentence-sentence score is the F1 aggregate, where
$F_1 = \tfrac{2PR}{P+R}$. We treat this F1 value as the pairwise similarity, which yields a symmetric matrix $S \in [-1,1]^{N\times N}$. We employ \texttt{Roberta$_\text{Large}$} \citep{liu2019robertarobustlyoptimizedbert} for obtaining the contextualized token embeddings.

\paragraph{Input Embeddings Cosine Similarity.}
We map each input to a single vector and measure pairwise similarity via cosine in embedding space. We use two variants: (i) for each example, we take the last-token hidden state from the model under evaluation, $\ell_2$-normalize it, and set $S_{ij}=\hat{h}_i^\top \hat{h}_j$, (ii) we encode each input with a frozen sentence-embedding model, normalize the embeddings, and compute the same cosine-based matrix. In both cases, we obtain a symmetric matrix $S \in [-1,1]^{N\times N}$. For the sentence-embedding variant, we use \texttt{MiniLM}\footnote{https://huggingface.co/sentence-transformers/all-MiniLM-L6-v2} \citep{wang2020minilmdeepselfattentiondistillation} and \texttt{gte-Qwen2-7B-instruct}\footnote{https://huggingface.co/Alibaba-NLP/gte-Qwen2-7B-instruct} \citep{li2023towards}; for the model-under-evaluation variant, we use \texttt{microsoft/Phi-3-mini-4k-instruct}\footnote{https://huggingface.co/microsoft/Phi-3-mini-4k-instruct} \citep{abdin2024phi3technicalreporthighly}, which yields the best performance in Appendix \ref{subsec:sem_vs_lex}.

\paragraph{Input and Output Embeddings Cosine Similarity.}
We represent each input-output pair (e.g., question+answer) as a single vector by taking the last-token hidden state of the concatenated sequence from the model under evaluation. We $\ell_2$-normalize these vectors and define pairwise similarity via cosine with $S_{ij}=\hat{h}_i^\top \hat{h}_j$. The resulting $S \in [-1,1]^{N\times N}$ is symmetric and reflects similarity over both the question and its associated answer. We use \texttt{microsoft/Phi-3-mini-4k-instruct} \citep{abdin2024phi3technicalreporthighly} for obtaining the hidden states, as it yields the best performance in Appendix \ref{subsec:sem_vs_lex}.

\paragraph{G-Vendi.} Following \cite{jung2025prismaticsynthesisgradientbaseddata}, we quantify the diversity of per–example gradients via a sketch-based spectral entropy. For each example, we form a compact count-sketch of the gradient of the (negative) log-probability of the correct answer under a proxy LM, yielding a matrix $G\!\in\!\mathbb{R}^{N\times d}$. We compute $C=\frac{1}{N}G^\top G$ and its eigenvalues $\{\lambda_i\}$. Let $p_i=\lambda_i/\sum_j \lambda_j$; the \emph{G-Vendi} score is the exponential Shannon entropy of this spectrum:
\[
\mathrm{G\text{-}Vendi}\;=\;\exp\!\Big(-\sum_i p_i \log p_i\Big),
\]
which acts as an effective rank where higher values indicate gradients spread across more orthogonal directions and lower values indicate concentration in a low dimensional subspace. For pairwise similarity, we $\ell_1$-normalize the sketch rows of $G$ and take their dot products to obtain a symmetric similarity matrix with unit diagonal. Following the original implementation, we employ \texttt{Qwen2.5-0.5B-Instruct} \citep{qwen2025qwen25technicalreport} as the proxy model.

\begin{table}
\centering
\scriptsize
\sisetup{
  detect-weight=true,
  detect-family=true,
  round-mode=places,
  round-precision=1,
  table-format=2.1,
}
\begin{tabular}{l S}
\toprule
\textbf{Method} & {\textbf{Duplicate Catch Ratio} ($\uparrow$)} \\
\midrule
\multicolumn{2}{l}{\textit{N-gram \& Token}} \\
Bigram            &  96.3 \\
BERTScore F1      & 100.0 \\
\midrule
\multicolumn{2}{l}{\textit{Embedding-Based}} \\
Input Embeddings$_\text{MiniLM}$        & 100.0  \\
Input Embeddings$_\text{gte-Qwen2}$     & 100.0  \\
Input Embeddings$_\text{phi3}$          & 100.0  \\
Input+Output Embeddings$_\text{phi3}$   & 100.0  \\
\midrule
\multicolumn{2}{l}{\textit{Literature}} \\
G-Vendi  & 100.0  \\
CORRS$_\text{Llama}$ &  100.0  \\
CORRS$_\text{all}$   & 100.0 \\
IRT Representation &  100.0 \\
\midrule
Predictive Similarity & 100.0 \\
\bottomrule
\end{tabular}
\caption{\textbf{Validation of our metric implementations.} All metrics other than bigram similarity perfectly detect exact duplicate questions, satisfying the minimum requirement to ensure implementation accuracy.}
\label{tab:nec_condition_wrap}
\end{table}

\paragraph{CORRS.} Following \cite{vivek2024anchorpointsbenchmarkingmodels}, given a bank of source models, we map each input $i$ to a vector $v_i \in \mathbb{R}^M$ whose $m$-th entry is the logit of the probability that model $m$ assigns to the correct choice. Then, the similarity between two examples is defined as the Pearson correlation of these vectors with $S_{ij}=\mathrm{corr}(v_i, v_j)$. The resulting $S \in [-1,1]^{N\times N}$ is symmetric and represents the cross-model agreement in correct class confidence across inputs. We instantiate the source bank using the Llama model family \citep{grattafiori2024llama} and the full set of models used in our experiments.

\paragraph{IRT Representation.} Following \cite{polo2024tinybenchmarksevaluatingllmsfewer}, from a bank of source models, we form a binary response matrix $Y\in\{0,1\}^{L\times N}$ whose $(\ell,i)$ entry indicates whether model $\ell$ answered example $i$ correctly. We then fit a $d$-dimensional IRT model with per-example parameters $(\alpha_i\in\mathbb{R}^d,\ \beta_i\in\mathbb{R})$ and per-model ability vectors $\theta_\ell\in\mathbb{R}^d$, using
\[
\Pr(Y_{\ell i}=1)=\sigma\!\big(-\,\theta_\ell^\top \alpha_i + \beta_i\big).
\]
Here, optimization alternates between gradient updates for $\theta$ (with $\ell_2$ regularization and recentring) and logistic regressions to update $(\alpha_i,\beta_i)$. Finally, we obtain the embedding $E_i=\big[\alpha_i;\beta_i\big]\in\mathbb{R}^{d+1}$ and define pairwise similarity by cosine similarity as $S_{ij}=\tfrac{E_i^\top E_j}{\|E_i\|\,\|E_j\|}$, which yields a symmetric matrix $S \in [-1,1]^{N\times N}$. We use $d=200$ and instantiate the source bank using the full set of models used in our experiments.

As open-source implementations of efficient evaluation metrics are unavailable, we re-implement them following the specifications in prior work. We then validate our implementations via a sanity check in which each metric was tasked with detecting verbatim duplicate questions. As shown in Table~\ref{tab:nec_condition_wrap}, all metrics (with the exception of bigram similarity) achieve perfect performance, satisfying the minimum requirement to ensure implementation accuracy.

\subsection{Measuring Semantic Similarity}
\label{subsec:sem_vs_lex}
An effective similarity measure for uncovering underlying data distributions must exhibit strong discriminative power, reliably identifying semantically similar data points while rejecting distractors. We empirically validate that predictive similarity meets this criterion, as it consistently distinguishes true semantic matches from misleading surface-level overlaps in RedundantQA.

\begin{table*}[t]
\centering
\resizebox{\linewidth}{!}{%
\scriptsize
\sisetup{
  detect-weight=true,
  detect-family=true,
  round-mode=places,
  round-precision=1,
  table-format=2.1,
}
\begin{tabular}{
    l
    S S S S S
    S S S S S
}
\toprule
\multirow{2}{*}{\textbf{Method}}
  & \multicolumn{5}{c}{\textbf{True Similar} ($\uparrow$)}
  & \multicolumn{5}{c}{\textbf{False Similar} ($\downarrow$)} \\
\cmidrule(lr){2-6} \cmidrule(lr){7-11}
  & {\textbf{Biology}} & {\textbf{Economics}} & {\textbf{Culture}} & {\textbf{History}} & {\textbf{All}}
  & {\textbf{Biology}} & {\textbf{Economics}} & {\textbf{Culture}} & {\textbf{History}} & {\textbf{All}} \\
\midrule
\textit{N-gram \& Token} \\
Bigram
  &  1.4 &  0.0 &  6.9 &  7.0 &  4.5
  & 70.0 & 67.6 & 30.6 & 47.9 & 53.7 \\
BERTScore F1
  &  8.6 &  0.0 & 12.5 & 21.1 & 12.4
  & 74.3 & 83.8 & 33.3 & 54.9 & 60.3 \\
\midrule
\textit{Embedding-Based} \\
Input Embeddings$_\text{MiniLM}$
  & 42.9 & 24.3 & \underline{62.5} & \underline{66.2} & \underline{54.1}
  & 27.1 & 51.4 &  5.6 & 12.7 & 21.1 \\
Input Embeddings$_\text{gte-Qwen2}$
  & 18.6 & 18.9 & 23.6 & 39.4 & 26.9
  & 38.6 & 51.4 & 26.4 & 22.5 & 33.5 \\
Input Embeddings$_\text{phi3}$
  & 21.4 &  5.4 & 16.7 & 36.6 & 22.7
  & 47.1 & 75.7 & 37.5 & 31.0 & 45.5 \\
Input+Output Embeddings$_\text{phi3}$
  & 51.4 & \underline{62.2} & 40.3 & 56.3 & 52.9
  & 22.9 & 27.0 & 16.7 & 15.5 & 20.2 \\
\midrule
\textit{Literature} \\
G-Vendi
  & \underline{62.9} & 37.8 & 36.1 & 46.5 & 47.9
  &  5.7 &  2.7 &  6.9 &  5.6 &  5.8 \\
CORRS$_\text{Llama}$
  &  2.9 &  8.1 &  9.7 & 14.1 &  9.1
  &  1.4 &  0.0 &  1.4 &  1.4 &  1.2 \\
CORRS$_\text{all}$
  & 35.7 & 13.5 & 59.7 & 59.2 & 47.5
  &  1.4 &  5.4 &  0.0 &  1.4 &  1.7 \\
IRT Representation
  &  1.4 &  0.0 &  1.4 &  0.0 &  0.8
  &  1.4 &  0.0 &  0.0 &  2.8 &  1.2 \\
\midrule
Predictive Similarity
  & \textbf{80.0} & \textbf{86.5} & \textbf{66.6} & \textbf{73.2} & \textbf{77.7}
  & 1.4 &  2.7 &  0.0 &  4.2 &  2.1 \\
\bottomrule
\end{tabular}
}
\caption{Proportion of identified true-similar ($\uparrow$) and false-similar ($\downarrow$) pairs by method and domain.}
\label{tab:suf_condition_restyled}
\end{table*}

As shown in Table~\ref{tab:suf_condition_restyled}, predictive similarity achieves the highest retrieval of true semantic matches across all domains, while maintaining one of the lowest rates of false matches. This indicates that it captures semantic equivalence without being misled by superficial lexical similarity. In contrast, embedding-based and bigram baselines suffer from high false positives, conflating surface-level resemblance with meaning. Metrics from the efficient evaluation literature show stronger performance but still fall short of predictive similarity. Overall, these results highlight the unique discriminative advantage of predictive similarity in measuring semantic similarity.

\subsection{Inducing the Semantic Partition}
\label{subsec:recover_dom}

\begin{table}
  \centering
  \scriptsize
  \setlength{\tabcolsep}{4pt}
  \renewcommand{\arraystretch}{0.95}
  \resizebox{\linewidth}{!}{%
  \begin{tabular}{lcccccc}
    \toprule
    \textbf{Method} & \multicolumn{2}{c}{\textbf{\texttt{RedundantQA}}} &
                      \multicolumn{2}{c}{\textbf{\texttt{RedundantQA-Ref}}} &
                      \multicolumn{2}{c}{\textbf{\texttt{MMLU-HS}}} \\
     & ARI & NMI & ARI & NMI & ARI & NMI \\
    \midrule
    \textit{N-gram \& Token} \\
    Bigram & -0.3 & 2.5 & -0.4 & 6.7 & 0.1 & 6.2  \\
    BERTScore F1 & -0.2 & 1.6 & 2.2 & 8.6 & 1.4 & 7.9 \\
    \midrule
    \textit{Embedding-Based} \\
    Input Embeddings$_\text{MiniLM}$      & 55.8 & 64.4 & 59.4 & 68.4  & 47.4 & 56.8 \\
    Input Embeddings$_\text{gte-qwen2-7b-instruct}$ & 28.6 & 34.9 & 36.3 & 45.9 & 27.1 & 34.4 \\
    Input Embeddings$_\text{phi3}$        & 27.8 & 37.1 & 33.0 & 45.6 & 25.1 & 31.6 \\
    Input+Output Embeddings$_\text{phi3}$ & 57.1 & 67.0 & 58.9 & 70.3 & 51.3 & 59.6 \\
    \midrule
    \textit{Literature} \\
    G-Vendi & 6.5 & 11.3 & 4.7 & 10.9 & 3.2 & 9.2 \\
    CORRS$_\text{all}$ & 6.1 & 8.2 & 4.8 & 8.2 & 2.8 & 8.4 \\
    IRT Representation & 2.1 & 2.7 & 0.4 & 1.9 & 0.4 & 1.5 \\
    \midrule
    Predictive Similarity$_\text{Qwen3 1.7B}$ & 60.4 & 70.4 & 62.5 & 76.5 & 55.4 & 62.1 \\
    \bottomrule
  \end{tabular}%
  }
  \caption{Validation of our partition induction method on RedundantQA and MMLU. Adjusted Rand Index (ARI) and Normalized Mutual Information (NMI) (higher is better for both) are shown for different methods on \texttt{\textbf{RedundantQA}}, \texttt{\textbf{RedundantQA-Ref}}, and \texttt{\textbf{MMLU-HS}}. Predictive similarity consistently achieves the best domain recovery.}
  \label{tab:redundantqa_results}
\end{table}

A core requirement for our work is that the similarity function should induce a semantic partition of the data. We evaluate this property by inducing cluster assignments from each metric and measuring agreement with the ground-truth domain labels in RedundantQA and its reference-only subset (RedundantQA-Ref) using Adjusted Rand Index (ARI) and Normalized Mutual Information (NMI). In addition, we apply the same protocol to MMLU high school subtasks, testing whether clusters recover canonical subject domains (e.g., computer science, biology, physics). Finally, we validate the top-performing similarity baselines from MMLU and RedundantQA on GPQA and M3Exam \citep{zhang2023m3exammultilingualmultimodalmultilevel}.

\begin{table}
  \centering
  \scriptsize
  \setlength{\tabcolsep}{4pt}
  \renewcommand{\arraystretch}{0.95}
  \resizebox{\linewidth}{!}{%
  \begin{tabular}{lcccc}
    \toprule
    \textbf{Method} &
    \multicolumn{2}{c}{\textbf{\texttt{M3Exam}}} &
    \multicolumn{2}{c}{\textbf{\texttt{GPQA}}} \\
    & ARI & NMI & ARI & NMI \\
    \midrule
    Input Embeddings$_\text{MiniLM}$          & 60.8          & 65.5          & 58.4          & 64.3          \\
    Input+Output Embeddings$_\text{Phi3}$     & 61.7          & 68.1          & 59.3          & 66.1          \\
    Predictive Similarity$_\text{Qwen3 1.7B}$ & \textbf{64.7} & \textbf{71.1} & \textbf{63.1} & \textbf{69.9} \\
    \bottomrule
  \end{tabular}%
  }
  \caption{Further validation of our partition induction method on \texttt{\textbf{GPQA}} and \texttt{\textbf{M3Exam}}. Predictive similarity achieves the best domain recovery on both benchmarks.}
  \label{tab:gpqa_m3exam_results}
\end{table}

As shown in Table~\ref{tab:redundantqa_results} and~\ref{tab:gpqa_m3exam_results}, predictive similarity achieves the highest agreement on all three sets, while strong embedding-based baselines are competitive yet consistently behind. By contrast, token/ngram measures and metrics from the efficient evaluation literature fail to recover domain structure, indicating that they are unreliable for semantic grouping. Taken together with the pairwise retrieval evidence, these results show that predictive similarity not only discriminates true semantic matches from distractors, but also organizes instances into compact and consistent clusters. This behavior is precisely what enables a cluster-centric analysis of benchmarks, yielding low within-cluster variance and high between-cluster separation.

\subsection{Capturing Benchmark \method{}}
\label{subsec:capture_quality}

\begin{figure*}[t]
    \centering
    \includegraphics[width=\linewidth]{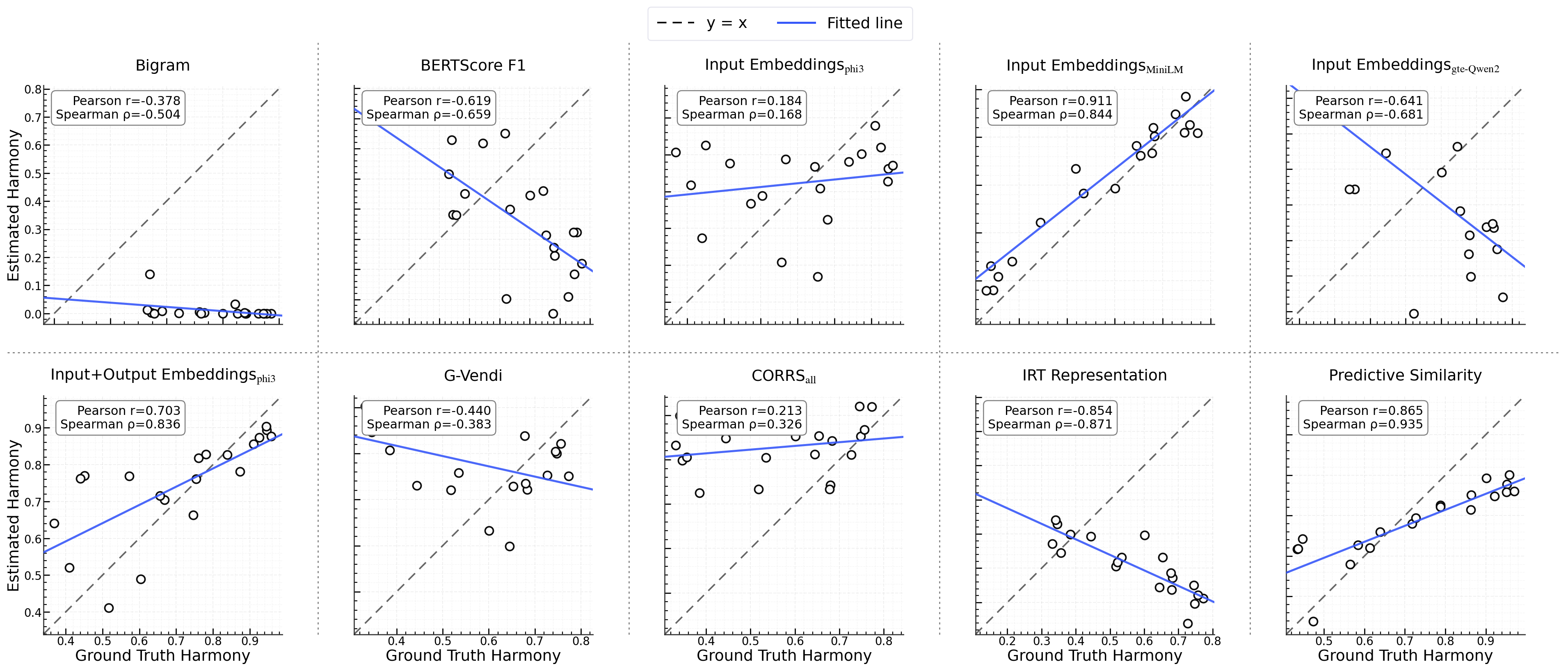}
    \caption{\textbf{Validation of our partition induction method on RedundantQA}. Predictive similarity achieves the strongest correlation between the ground truth \method{} and estimated \method{}, while the \texttt{MiniLM} input-embedding baseline is a close runner-up.}
    \label{fig:baseline_validations}
\end{figure*}

We evaluate all alternative similarity baselines from Appendix \ref{subsec:alt_baselines} in the identical setting of \S\ref{subsec:empirical_val}. For each baseline, we induce clusters from its similarity matrix, compute \method{} $H(\mathcal{G})$, and report correlations between the ground-truth \method{} and its counterpart computed from the partition induced by each baseline. We additionally validate the top-performing similarity baselines from MMLU and RedundantQA on GPQA and M3Exam.

\begin{figure*}
    \centering
    \includegraphics[width=\linewidth]{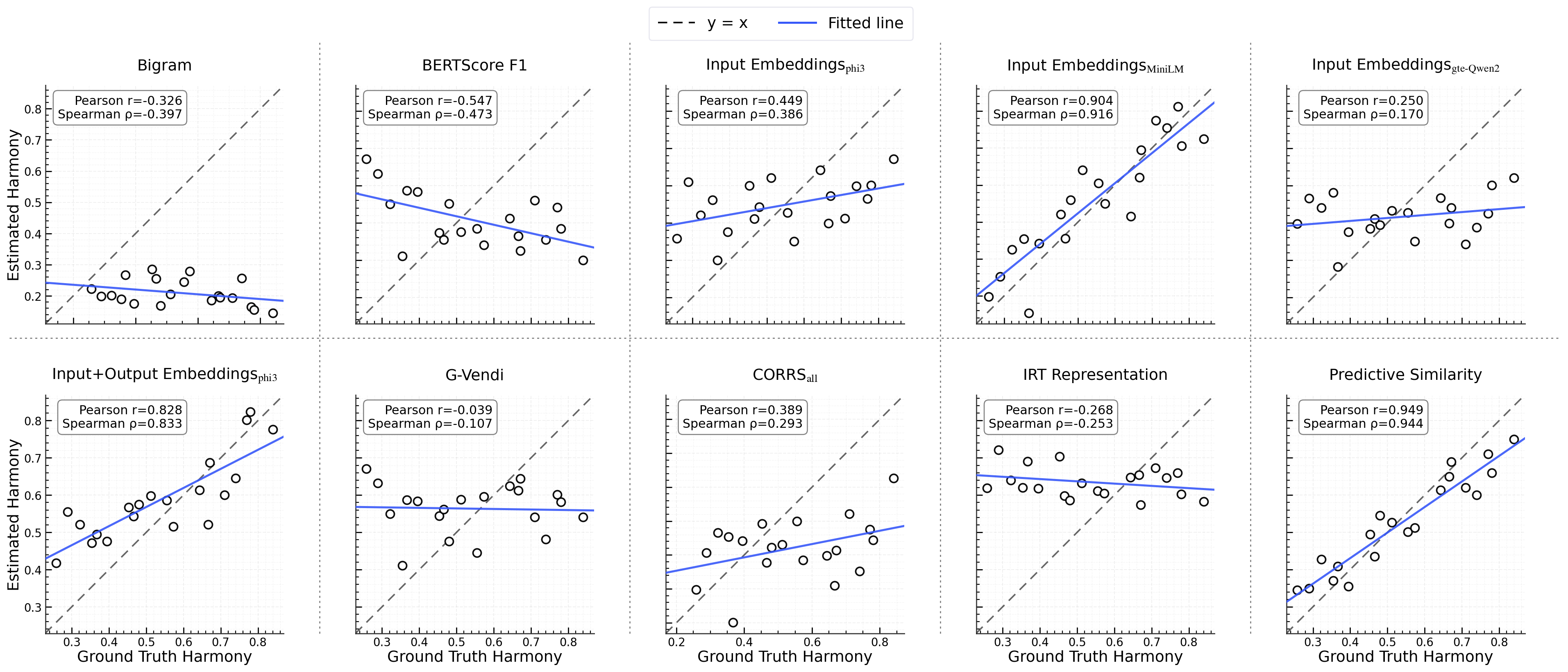}
    \caption{\textbf{Validation of our partition induction method on MMLU}. Similar to the results on RedundantQA, predictive similarity achieves the strongest correlation between the ground truth \method{} and estimated \method{}, while the \texttt{MiniLM} input-embedding baseline is a close runner-up.}
    \label{fig:baseline_validations_mmlu}
\end{figure*}

\begin{figure*}[t]
    \centering
    \begin{subfigure}[b]{0.48\linewidth}
        \centering
        \includegraphics[width=\linewidth]{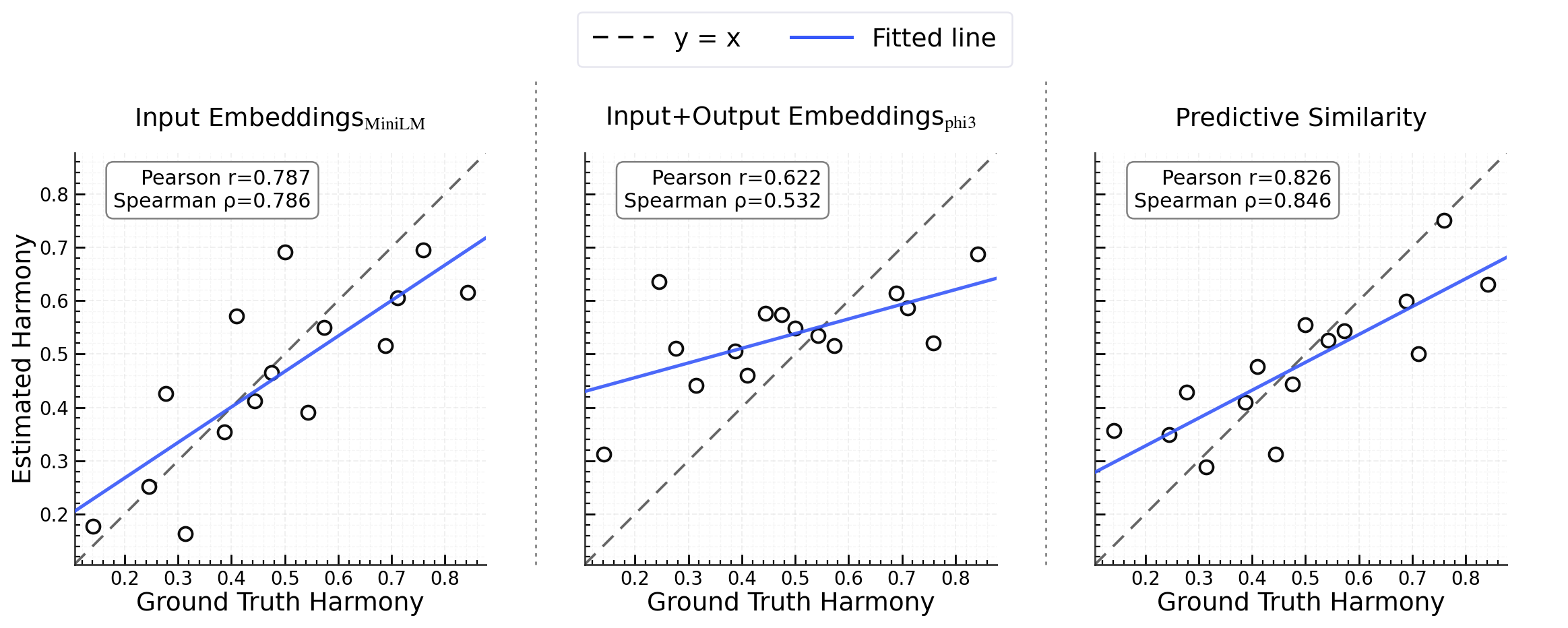}
        \caption{GPQA}
        \label{fig:baseline_validation_gpqa}
    \end{subfigure}
    \hfill
    \begin{subfigure}[b]{0.48\linewidth}
        \centering
        \includegraphics[width=\linewidth]{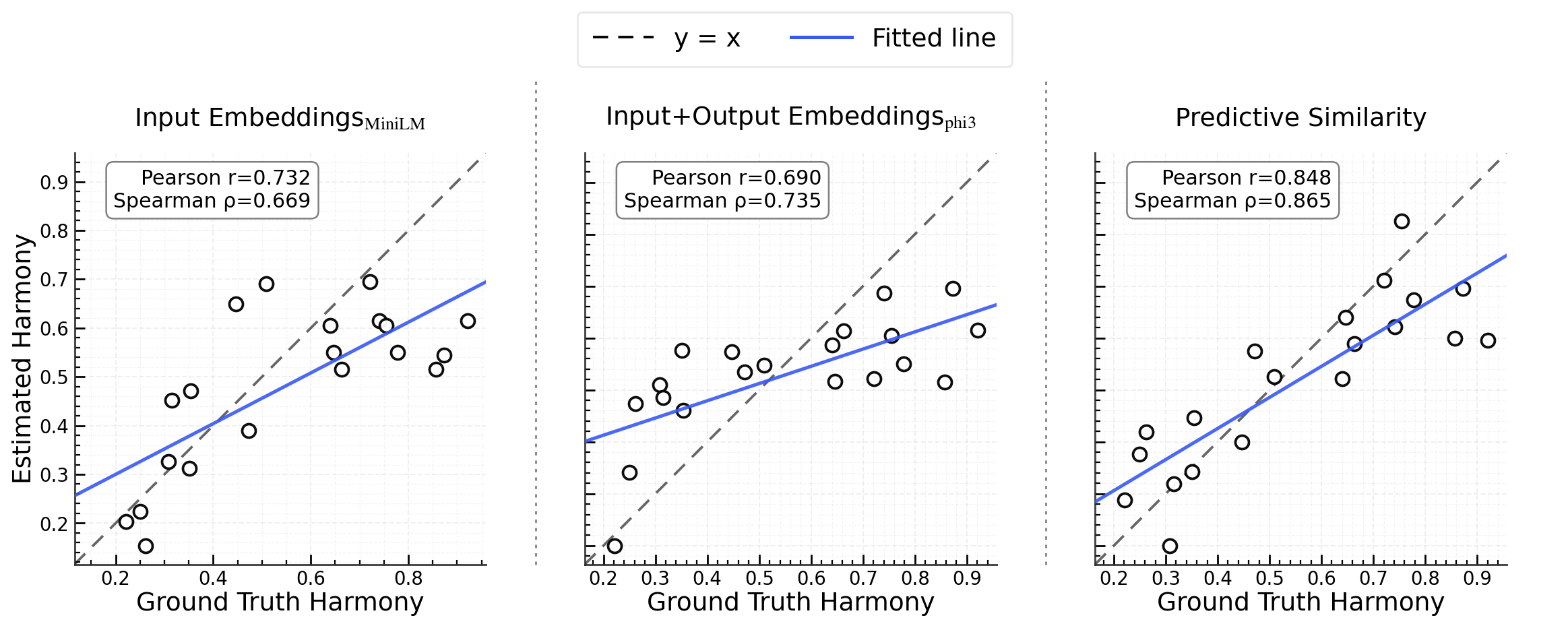}
        \caption{M3Exam}
        \label{fig:baseline_validation_m3exam}
    \end{subfigure}
    \caption{\textbf{Validation of our partition induction method on GPQA and M3Exam}. Similar to the results on RedundantQA and MMLU, predictive similarity achieves the strongest correlation between the ground truth \method{} and estimated \method{}.}
    \label{fig:baseline_validation_gpqa_m3exam}
\end{figure*}

\begin{figure*}
    \centering
    \includegraphics[width=\linewidth]{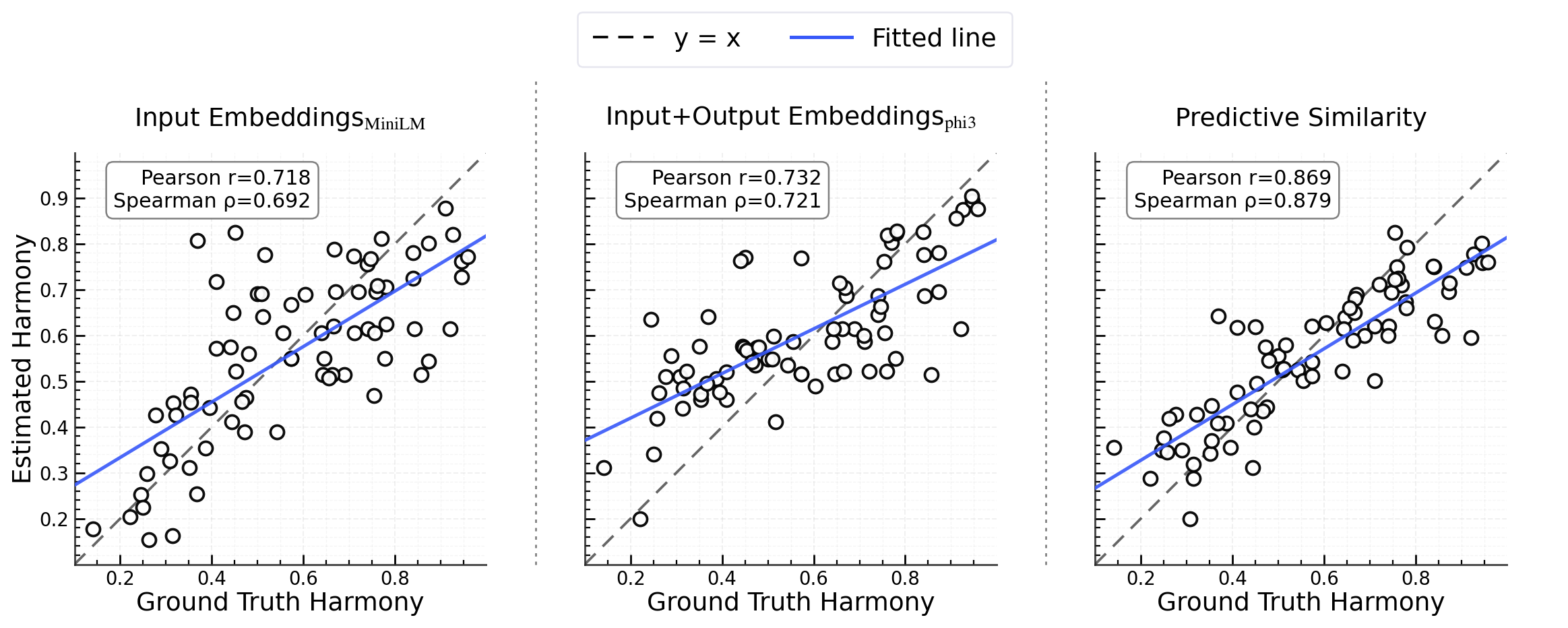}
    \caption{\textbf{Validation of our partition induction method on a combination of RedundantQA, MMLU, GPQA, and M3Exam}. Similar to the individual results, predictive similarity achieves the strongest correlation between the ground truth \method{} and estimated \method{}.}
    \label{fig:baseline_validation_combined}
\end{figure*}

As shown in Fig. \ref{fig:baseline_validations}, \ref{fig:baseline_validations_mmlu} and \ref{fig:baseline_validation_gpqa_m3exam}, predictive similarity achieves the strongest correlation with ground-truth entropy. Similar to prior validation experiments (App. \ref{subsec:sem_vs_lex}, \ref{subsec:recover_dom}), embedding-based baselines are the next-best performers but consistently lag behind, whereas token-and-$n$-gram overlap measures perform substantially worse. These results establish predictive similarity as the most reliable similarity metric choice for capturing the benchmark dynamics.

Lastly, Figure \ref{fig:baseline_validation_combined} pools all validation points from RedundantQA, MMLU, GPQA, and M3Exam, complementing the per-benchmark results in Figure \ref{fig:baseline_validations}, \ref{fig:baseline_validations_mmlu}, and \ref{fig:baseline_validation_gpqa_m3exam}, where predictive similarity consistently achieves the strongest alignment with ground-truth \method{} across imbalance variants. When we combine these experiments into a single pool, the relationship remains strongly positive (Pearson $r=0.869$ and Spearman $\rho=0.879$), showing that our estimator preserves \method{} ordering not only within each benchmark family but also across benchmarks. This supports our claim that estimated \method{} provides a benchmark-agnostic signal of distributional robustness.

\subsection{Consistency of Predictive Similarity Across Models}
\label{subsec:consistency_pred_sim}
As defined in \S\ref{subsec:methodology}, predictive similarity induces, for each benchmark $\mathcal{B}$ and model $f$, a symmetric similarity matrix $S^{(f,\mathcal{B})} \in (0,1]^{N_\mathcal{B} \times N_\mathcal{B}}$ over the $N_\mathcal{B}$ items of $\mathcal{B}$. In this section, we ask whether these model-specific neighborhoods are idiosyncratic. To quantify cross-model consistency, we correlate the upper-triangular entries of the corresponding similarity matrices using both rank-based (Spearman) and linear (Pearson) correlations:
\[
\begin{aligned}
r_{\mathrm{S}}^{\mathcal{B}}(f_1,f_2)
&=
\rho_{\mathrm{S}}\!\big(\mathrm{vec}(S^{(f_1,\mathcal{B})}),\ \mathrm{vec}(S^{(f_2,\mathcal{B})})\big),\\
r_{\mathrm{P}}^{\mathcal{B}}(f_1,f_2)
&=
\rho_{\mathrm{P}}\!\big(\mathrm{vec}(S^{(f_1,\mathcal{B})}),\ \mathrm{vec}(S^{(f_2,\mathcal{B})})\big).
\end{aligned}
\]

where $\mathrm{vec}(\cdot)$ stacks the upper triangle of a matrix into a vector, $\rho_{\mathrm{S}}$ is Spearman’s rank correlation, and $\rho_{\mathrm{P}}$ is Pearson’s correlation. We report both, as Spearman is invariant to monotone re-scalings and thus robust to calibration differences across models, while Pearson captures linear alignment in similarity magnitudes.

\begin{table*}[t]
\centering
\scriptsize
\setlength{\tabcolsep}{4pt}
\resizebox{\textwidth}{!}{
\begin{tabular}{l*{6}{cc}}
\toprule
& \multicolumn{2}{c}{\textbf{Gemma}} & \multicolumn{2}{c}{\textbf{Llama}} & \multicolumn{2}{c}{\textbf{OLMo}} & \multicolumn{2}{c}{\textbf{Phi}} & \multicolumn{2}{c}{\textbf{Qwen}} & \multicolumn{2}{c}{\textbf{Average}} \\
\cmidrule(lr){2-3}\cmidrule(lr){4-5}\cmidrule(lr){6-7}\cmidrule(lr){8-9}\cmidrule(lr){10-11}\cmidrule(lr){12-13}
\textbf{Benchmark} & \textbf{Spearman} & \textbf{Pearson} & \textbf{Spearman} & \textbf{Pearson} & \textbf{Spearman} & \textbf{Pearson} & \textbf{Spearman} & \textbf{Pearson} & \textbf{Spearman} & \textbf{Pearson} & \textbf{Spearman} & \textbf{Pearson} \\
\midrule
AQUA-RAT      & 61.8 & 62.7 & 63.3 & 65.9 & 47.0 & 49.3 & 49.8 & 50.9 & 76.1 & 76.5 & 59.6 & 61.1 \\
ARC-Challenge & 41.9 & 42.7 & 76.3 & 76.9 & 72.9 & 73.9 & 71.9 & 73.9 & 74.5 & 76.1 & 67.5 & 68.7 \\
ARC-Easy      & 41.6 & 42.0 & 66.6 & 67.1 & 67.9 & 68.7 & 75.3 & 76.3 & 62.7 & 64.0 & 62.8 & 63.6 \\
ART           & 25.5 & 26.4 & 84.1 & 84.6 & 76.6 & 77.9 & 78.6 & 79.8 & 75.7 & 76.8 & 68.1 & 69.1 \\
BoolQ         & 25.5 & 25.9 & 58.3 & 59.8 & 35.3 & 36.3 & 18.2 & 18.9 & 27.5 & 28.9 & 33.0 & 34.0 \\
CommonsenseQA & 33.8 & 35.6 & 33.6 & 34.4 & 62.8 & 64.5 & 42.4 & 43.8 & 39.1 & 39.8 & 42.3 & 43.6 \\
COPA          & 27.9 & 29.5 & 76.7 & 78.4 & 68.7 & 70.2 & 64.8 & 67.2 & 51.4 & 52.8 & 57.9 & 59.6 \\
GPQA          & 77.8 & 78.0 & 53.4 & 51.9 & 64.1 & 65.2 & 82.1 & 83.0 & 85.5 & 86.1 & 72.6 & 72.8 \\
LogiQA        & 55.4 & 57.3 & 77.7 & 79.8 & 67.8 & 67.2 & 70.3 & 72.1 & 80.1 & 80.3 & 70.3 & 71.3 \\
MathQA        & 33.5 & 34.2 & 61.0 & 62.0 & 55.2 & 56.0 & 54.9 & 54.0 & 57.1 & 56.4 & 52.3 & 52.5 \\
OpenBookQA    & 33.2 & 34.0 & 73.9 & 75.0 & 70.1 & 72.3 & 74.0 & 75.5 & 71.0 & 72.5 & 64.4 & 65.9 \\
PIQA          & 38.6 & 39.6 & 78.6 & 80.2 & 77.3 & 79.4 & 67.0 & 68.2 & 74.6 & 76.4 & 67.2 & 68.8 \\
PubMedQA      & 50.5 & 50.0 & 60.8 & 60.8 & 37.8 & 36.8 & 50.6 & 50.6 & 63.0 & 61.1 & 52.5 & 51.9 \\
QUARTZ        & 37.9 & 39.0 & 81.4 & 80.3 & 65.9 & 63.9 & 76.0 & 74.9 & 70.0 & 68.1 & 66.2 & 65.2 \\
SciQ          & 22.0 & 22.5 & 55.7 & 58.3 & 60.3 & 61.4 & 56.0 & 56.0 & 58.0 & 60.3 & 50.4 & 51.7 \\
SocialIQA     & 20.7 & 21.3 & 72.9 & 73.5 & 64.6 & 65.6 & 63.5 & 64.9 & 61.2 & 62.0 & 56.6 & 57.5 \\
StrategyQA    &  3.6 &  3.7 & 48.2 & 49.6 & 20.5 & 20.9 & 21.6 & 21.4 & 27.9 & 28.9 & 24.4 & 24.9 \\
TruthfulQA    & 50.4 & 52.7 & 78.4 & 80.7 & 71.4 & 73.6 & 85.3 & 86.4 & 77.1 & 79.3 & 72.5 & 74.5 \\
\midrule
\textbf{Average} & 37.9 & 38.7 & 66.7 & 67.7 & 60.3 & 61.3 & 61.2 & 62.1 & 62.9 & 63.7 & 57.8 & 58.7 \\
\bottomrule
\end{tabular}}
\caption{\textbf{Cross-model consistency of predictive similarity}: Spearman and Pearson correlations (values $\times 100$) between the upper-triangle entries of affinity matrices, by benchmark (rows) and model family (columns). The rightmost block reports per-benchmark averages across families; the bottom row reports per-family averages across benchmarks; the bottom-right cell shows overall means.}
\label{tab:predsim_consistency_table}
\end{table*}

As shown in Table~\ref{tab:predsim_consistency_table}, predictive similarity neighborhoods exhibit substantial within-family consistency across many benchmarks. Averaging across families yields high per-benchmark means clustered in the mid-$60$s, with notable peaks for \emph{LogiQA} (70.3 Spearman / 71.3 Pearson), \emph{TruthfulQA} (72.5 / 74.5), \emph{PIQA} (67.2 / 68.8), and \emph{ART} (68.1 / 69.1). Family-wise averages further show broad stability for Llama (66.7 / 67.7), Qwen (62.9 / 63.7), and Phi (61.2 / 62.1), with OLMo close behind (60.3 / 61.3) and Gemma lower (37.9 / 38.7). In particular, \emph{ART}, \emph{COPA}, \emph{LogiQA}, \emph{PIQA}, and \emph{TruthfulQA} attain high agreement for Llama, Phi, and Qwen (typically $70$–$85$), indicating that the induced item-item structure is largely task-driven rather than model idiosyncratic. Knowledge-centric \emph{GPQA} is also strong for Phi and Qwen ($82$–$86$). By contrast, \emph{StrategyQA}, \emph{BoolQ}, and, to a lesser extent, \emph{SocialIQA} show weaker agreement particularly for Gemma and Phi, suggesting greater family-specific effects on these benchmarks.

\subsection{Probing the Determinants of Predictive Similarity}
\label{subsec:determinants_pred_sim}
\begin{table}
  \centering
  \scriptsize
  \setlength{\tabcolsep}{6pt}
  \renewcommand{\arraystretch}{1.0}
  \begin{tabular}{lcccc}
    \toprule
    & \multicolumn{2}{c}{\textbf{$S_\text{JS}$}} & \multicolumn{2}{c}{\textbf{$S_\text{KL\texttt{-top-50}}$}} \\
    \cmidrule(lr){2-3}\cmidrule(lr){4-5}
    \textbf{Benchmark}  & Pearson & Spearman & Pearson & Spearman \\
    \midrule
    AQUA-RAT & 96.1 & 95.5 & 97.4 & 97.4 \\
    ARC-Challenge & 95.7 & 95.2 & 98.5 & 98.5 \\
    ARC-Easy & 95.4 & 95.0 & 98.7 & 98.7 \\
    ART & 95.5 & 95.4 & 97.1 & 96.9 \\
    BoolQ & 93.5 & 92.5 & 98.4 & 98.4 \\
    CommonsenseQA & 97.0 & 96.7 & 99.5 & 99.5 \\
    COPA & 94.7 & 95.2 & 95.1 & 94.9 \\
    GPQA & 97.1 & 97.1 & 99.1 & 99.0 \\
    LogiQA & 98.7 & 98.5 & 98.5 & 98.5 \\
    MathQA & 99.1 & 99.4 & 96.4 & 96.3 \\
    OpenBookQA & 97.6 & 97.7 & 97.5 & 97.6 \\
    PIQA & 96.4 & 96.1 & 96.2 & 95.9 \\
    PubMedQA & 96.8 & 96.7 & 99.6 & 99.6 \\
    QUARTZ & 95.8 & 95.8 & 99.8 & 99.8 \\
    SciQ & 93.6 & 93.3 & 98.8 & 98.9 \\
    SocialIQA & 98.2 & 98.2 & 97.0 & 96.8 \\
    StrategyQA & 99.4 & 99.5 & 97.0 & 97.0 \\
    TruthfulQA & 94.6 & 94.2 & 98.1 & 97.7 \\
    \bottomrule
  \end{tabular}
  \caption{\textbf{Agreement between predictive similarity variants.} Per-benchmark correlations between \(S_{\text{KL}}^{(f,\mathcal{B})}\) and (i) \(S_{\text{JS}}^{(f,\mathcal{B})}\) and (ii) \(S_{\text{KL-}\texttt{top-50}}^{(f,\mathcal{B})}\) show that neighborhoods are driven by head probability mass and remain stable under truncation or mixture smoothing.}
  \label{tab:sim_corrs}
\end{table}
\paragraph{Dependence on the Tail.}
We test sensitivity to the probability tail by constructing a truncated variant that re-normalizes mass over the union of the \texttt{top-50} tokens, yielding $S_{\text{KL-\texttt{top-50}}}^{(f,\mathcal{B})}$, and contrasting it with the full $S_{\text{KL}}^{(f,\mathcal{B})}$.

\paragraph{JS vs.\ KL-based Similarity.}
Equation~\ref{eq:predictive:similarity} defines \(S\) as an RBF of the Jeffreys divergence, producing \emph{sharp, tunable} neighborhoods that strongly penalize coverage errors (as near-zeros drive \(J\) higher and \(S\) lower). Jensen-Shannon (JS) instead compares to the mixture \(M=\tfrac12(\bar{p}_f(x_i)+\bar{p}_f(x_j))\), yielding a bounded, tail-robust divergence that is easier to compare across benchmarks. To place JS on the same similarity scale, we apply the same RBF transform from Equation~\ref{eq:predictive:similarity} entrywise, obtaining $S_{\text{JS}}^{(f,\mathcal{B})}$.

Following Appendix \ref{subsec:consistency_pred_sim}, we compute Spearman and Pearson correlations between the upper-triangular entries of $S_{\text{KL}}^{(f,\mathcal{B})}$ and, respectively, $S_{\text{KL-\texttt{top-50}}}^{(f,\mathcal{B})}$ and $S_{\text{JS}}^{(f,\mathcal{B})}$. We report per-benchmark scores where $f$ is $\text{OLMo 2 7B}$. \emph{High agreement} indicates that neighborhoods are driven by high probability mass and remain stable under truncation or mixture smoothing, while \emph{low agreement} indicates sensitivity to tail mismatches or calibration asymmetries that Jeffreys magnifies but JS attenuates.

Across benchmarks, correlations between \(S_{\text{KL}}^{(f,\mathcal{B})}\) and (i) $S_\text{KL\texttt{-top-50}}^{(f,\mathcal{B})}$ and (ii) $S_\text{JS}^{(f,\mathcal{B})}$ variants are uniformly high (typically \(95\text{-}99\%\)) (Table~\ref{tab:sim_corrs}). This indicates that the divergence between probability distributions generated by OLMo~2~7B is governed by head probability mass rather than the tail. Truncation preserves structure nearly perfectly as \(S_{\text{KL\texttt{-top-50}}}\) matches or exceeds \(S_{\text{JS}}\) on most tasks, while \(S_{\text{JS}}\) remains strongly aligned, reflecting robustness to calibration and coverage noise. Modest dips (e.g., COPA, PIQA) suggest settings where tail mismatches or asymmetries matter more, but overall the stability under truncation and mixture smoothing supports that \(S\) captures meaningful, head-driven divergence.

\subsection{Theoretical and Computational Discussions}
\paragraph{A Theoretical Perspective on Predictive Similarity.} We measure similarity via the (symmetrized) $D_{\text{KL}}$ between model predictive distributions because it aligns with how models differ operationally and geometrically. First, $D_{\text{KL}}$ has a clear testing meaning, as it governs optimal error exponents in distinguishing two distributions. Hence, larger $D_{\text{KL}}$ divergence implies that the model would more reliably tell the two inputs apart (by Stein/Chernoff asymptotics) \citep{elements_of_it}. Second, small $D_{\text{KL}}$ guarantees closeness in total variation by Pinsker’s inequality, implying high indistinguishability and hence high similarity for our purposes \citep{elements_of_it}. Third, $D_{\text{KL}}$ is information monotone under coarse-graining, making the measure stable to relabeling or merging answer tokens/options that preserve semantics \citep{it_and_stats}. Finally, locally $D_{\text{KL}}$ induces the Fisher-Rao geometry on the probability simplex, so $\exp(-\tau\,\mathrm{D_\text{KL}})$ behaves like a Gaussian kernel in the natural metric of the model’s predictive space, yielding compact clusters of similar predictive behavior \citep{info_geo}. We use the Jeffreys (symmetrized) form to remove directionality while retaining these properties.

\paragraph{Computational Overhead of Predictive Similarity.} Predictive similarity is a \emph{post hoc} computation, since we operate on the logits already cached from the benchmark evaluation. Hence, no additional model forward passes are required. Given these logits, we convert them to predictive distributions and evaluate the pairwise KL terms that define the similarity in Eq.~\ref{eq:predictive:similarity}. 

The principal cost arises from forming pairwise interactions across $N$ items, which is quadratic in $N$ and linear in the label-space size $D$ (i.e., $O(N^2 \cdot D)$ time). Memory is dominated by storing the evaluation logits ($O(N \cdot D)$) and the similarity matrix ($O(N^2)$). In practice, $D$ corresponds to the size of the vocabulary of a given model and can be large. We therefore view the cost as $O(N^2 D)$ and the memory requirement as $O(N \cdot D)$. When $N$ is large, standard remedies (e.g. blockwise evaluation) reduce peak memory without changing the definition of the metric. Overall, computing predictive similarity adds negligible \emph{inference} overhead and modest \emph{analysis} overhead relative to running the benchmarks themselves.

Compared to alternative baselines discussed in Appendix \ref{subsec:alt_baselines}, predictive similarity is computationally frugal: it reuses cached logits and requires neither additional inference nor any backward passes. By contrast, embedding baselines, as well as BERTScore, invoke separate encoders (extra forward passes), G-Vendi relies on gradients (backward passes), and CORRS/IRT aggregate signals from a bank of models (multiple evaluations per item). While string-based methods such as Bigram are lightweight, they do not leverage model behavior. Thus, predictive similarity offers a favorable trade-off between compute and quality when benchmarks are already being run.

\subsection{Discussion on Model-Specific Similarity}

Our goal is to \emph{evaluate benchmarks}, not to define a single, task-agnostic neighborhood of data points. 
A benchmark can be distributionally robust for one model but not for another. Accordingly, the similarity function used to induce the partition $\mathcal{G}_f$ should be \emph{conditional on the model $f$}. We provide our rationale below.

\paragraph{Evaluation target is $H_{\mathcal{B}}(f)$.}
Section~\ref{sec:benchmark_harmony} defines \method{} \emph{per model}, $H_{\mathcal{B}}(f)$, and then aggregates across $f$ via $(\mu_H,\sigma_H^2)$ in Eq.~\ref{eq:mu_and_sigma}. Using a \emph{global, model-agnostic} similarity collapses distinct predictive neighborhoods into a single partition, implicitly assuming that $\mathcal{G}_f$ is invariant across $f$. This undermines the very statistic we report: two models with the same accuracy profile but different predictive structure could receive the same $H$ under a fixed partition, obscuring model specialties.

\paragraph{Benchmarks are instruments relative to a model.}
A benchmark is a diagnostic instrument for a \emph{given} model: priors, tokenization, calibration, and pre-training exposure all change which items are \emph{similar} from the model’s perspective. Hence, a model can perform uniformly on a benchmark while another one overfits to certain subdomains. Model-specific similarity preserves this relativity, letting robustness vary meaningfully across families.

\section{Design Choices for \method{}}
\label{app:harmony_desing_choices}

We briefly motivate our definition of \method{} and our choice of normalized Shannon entropy. The primary reason is pragmatic: Shannon entropy is the canonical measure of uncertainty (and thus uniformity) in information theory, so its behavior and scale are well understood and easy to interpret.

Recall that we first compute non-negative weights $\pi_i=w_iK_i$ over subdomains $i = 1,\dots,K$ that increase with (i) the mass of the dataset assigned to subdomain $i$ and (ii) how close the model's performance on $i$ is to the benchmark mean. Let
\[
p_i \;=\; \frac{\pi_i}{\sum_{j=1}^K \pi_j}, \qquad \sum_{i=1}^K p_i = 1,
\]
denote the normalized weights as in Section~\ref{subsec:harmony}, so that $p = (p_1,\dots,p_K)$ yields a probability vector that summarizes \emph{where} in the benchmark the model is performing well.

Our design is guided by three criteria:

\begin{enumerate}[label=(\roman*)]
    \item \textbf{Joint coverage and uniformity:} \method{} should increase when more dataset mass lies on subdomains whose performance is close to the benchmark mean, and decrease when performance is concentrated on a few subdomains.
    \item \textbf{Comparability across benchmarks:} \method{} should lie in $[0,1]$ so benchmarks with different numbers of subdomains are directly comparable.
    \item \textbf{Interpretability as robustness:} High \method{} should indicate that the benchmark behaves \emph{as if} many subdomains are near the mean, while low \method{} signals effective concentration on a smaller subset.
\end{enumerate}

We therefore define \method{} as the normalized Shannon entropy:
\[
H(\mathcal{G}_f) \;=\; -\frac{1}{\log k}\sum_{i=1}^k p_i \log\!\big(p_i+\varepsilon\big)\in[0,1],
\]
This choice yields:

\begin{itemize}
    \item \textbf{Boundedness:} $0 \leq H(\mathcal{G}_f) \leq 1$, with $1$ attained only when $p_i = 1/k$ for all $i$, and values approaching $0$ as mass concentrates on a single subdomain.
    \item \textbf{Effective number of well-performing subdomains:} Let $\tilde{H}(p) = -\sum_{i=1}^k p_i \log p_i$ denote the (unnormalized) Shannon entropy. Its exponential $\exp(\tilde{H}(p))$ is often interpreted as the \emph{effective number} of equally weighted categories. Since $H(\mathcal{G}_f) = \tilde{H}(p)/\log k$, this can be read as a normalized effective number, i.e., the effective fraction of subdomains whose performance is close to the benchmark mean.
    \item \textbf{Sensitivity to non-uniformity:} Shannon entropy is strictly concave and Schur-concave, which makes any move from a more uniform to a more concentrated distribution $p = (p_1,\dots,p_k)$ strictly decrease $H(\mathcal{G}_f)$.
\end{itemize}

\subsection{Alternative Uniformity Measures}

\paragraph{Variance and coefficient of variation.}
A natural idea is to measure dispersion of subdomain performances $\{\Psi(f; A_i)\}_{i=1}^k$ via a (weighted) variance
\[
\begin{aligned}
\mathrm{Var}\big(\Psi(f;\cdot)\big)
&=
\sum_{i=1}^k w_i \Bigl(\Psi(f; A_i) - \bar{\Psi}\Bigr)^2,\\
\bar{\Psi}
&=
\sum_{i=1}^k w_i \Psi(f; A_i).
\end{aligned}
\]

However, variance is scale-dependent and not naturally normalized to $[0,1]$, and it does not admit an \emph{effective number of subdomains} interpretation. The coefficient of variation partly addresses scale but becomes unstable when the mean is close to $0$ or $1$, which is common for strong models. In contrast, \method{}  depends only on the probability distribution $p = (p_1,\dots,p_k)$ over subdomains and is automatically normalized.

\paragraph{Gini coefficient and related indices.}
Inequality indices such as the Gini coefficient measure concentration but are typically defined directly on the performance values $\{\Psi(f; A_i)\}_{i=1}^k$, not on a distribution that already combines dataset mass and closeness-to-mean. This makes it harder to interpret them as describing how \emph{dataset mass} is distributed over near-mean vs.\ outlier subdomains, and they lack the effective-number interpretation that entropy has.

\paragraph{Rényi entropies.}
Rényi entropies form a one-parameter family
\[
H_\alpha(p)
\;=\;
\frac{1}{1-\alpha} \log \Bigl(\sum_{i=1}^k p_i^\alpha\Bigr),
\quad \alpha > 0, \alpha \neq 1,
\]
with Shannon entropy recovered as the limit $\alpha \to 1$. Different orders $\alpha$ emphasize different parts of the distribution ($\alpha > 1$ focuses on high-probability subdomains and $\alpha < 1$ emphasizes low-probability ones), effectively introducing a hyperparameter that controls how aggressively \method{} responds to tails and modes. In this sense, choosing a particular Rényi order $\alpha$ is analogous to choosing a norm order (e.g., $\ell_\alpha$) when measuring distances. Therefore, Rényi entropy yields a whole family of alternatives, but also an extra design choice about which order to prefer. We deliberately avoid this additional degree of freedom. Using the Shannon case $\alpha = 1$ gives a single, parameter-free and widely interpretable notion of entropy that treats subdomains according to their weights $p_i$, without further reweighting.

\subsection{Robustness of Harmony}

\method{} is designed to penalize benchmarks that leave small, poorly performing regions unaddressed. Ignoring the smoothing term $\varepsilon$ for clarity, the underlying (unnormalized) Shannon entropy
\[
\tilde{H}(p) = -\sum_{i=1}^k p_i \log p_i
\]
can be viewed as the average information content under $p$. Each subdomain contributes $p_i \log(1/p_i)$, and $\log(1/p_i)$ grows as $p_i$ becomes small. As a result, shifting even modest probability mass away from under-represented, well-performing subdomains (or toward poorly performing ones) leads to a noticeable drop in $H(\mathcal{G}_f)$. In this sense, \method{} is particularly sensitive to \emph{gaps} in coverage, whereas variance-based dispersion measures tend to be dominated by large deviations in high-mass components and can largely ignore small but problematic regions.

Our formulation is, like any scalar index, not completely immune to adversarial benchmark design (e.g., redefining the subdomain partition). However, for a fixed partition and mean performance, increasing $H(\mathcal{G}_f)$ requires genuinely redistributing dataset mass so that a larger share lies in subdomains whose performance is close to the mean, while simple rescalings of scores or adding redundant, near-duplicate subdomains do not help. This makes \method{} behave as an interpretable robustness index with higher values corresponding to benchmarks where performance is spread broadly across the subdomains, rather than confined to certain subdomains.

\section{\method{} with Model-Agnostic Partitions}
\label{app:minilm_sanity}

Our main analysis (\S\ref{sec:main:analyses}) uses model-sensitive predictive-similarity partitions. This choice is intended to capture the subdomain structure relevant to each model's behavior. As a sanity check, we ask whether the resulting MCQA benchmark landscape changes substantially when the partition is instead fixed and model-agnostic. To this end, we repeat the MCQA analysis using partitions induced from \texttt{all-MiniLM-L6-v2} embeddings. For each benchmark, we embed all inputs with MiniLM, compute pairwise cosine similarities, and apply the same clustering procedure used in the main analysis. The resulting partition is fixed for a benchmark and reused across all models.

\begin{figure}[t]
    \centering
    \includegraphics[width=\linewidth]{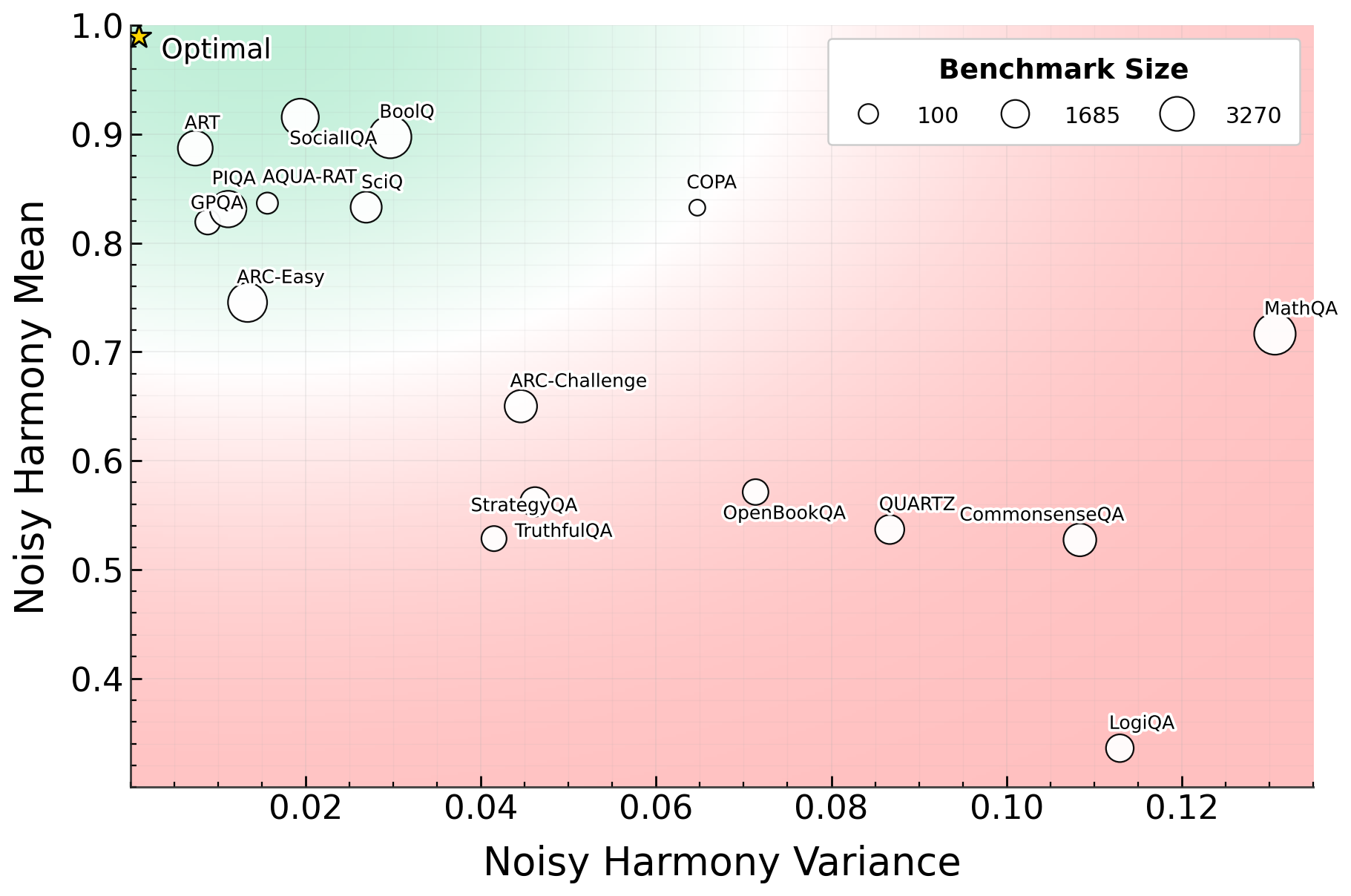}
    \caption{
    \textbf{MCQA benchmark landscape using model-agnostic MiniLM partitions.}
    Each point represents a benchmark plotted by the mean and variance of \method{} across models, using a fixed partition induced from \texttt{all-MiniLM-L6-v2} embeddings. The qualitative structure is similar to the model-sensitive predictive-similarity analysis in Fig.~\ref{fig:non_mmlu_tasks}.
    }
    \label{fig:minilm_mcqa}
\end{figure}

Figure~\ref{fig:minilm_mcqa} shows the resulting mean-variance plane under the MiniLM partitions. Qualitatively, the benchmark landscape remains similar to the model-sensitive analysis in Fig.~\ref{fig:non_mmlu_tasks}, as benchmarks that are comparatively high or low in \method{} under predictive-similarity partitions tend to remain so under the model-agnostic partition. Table~\ref{tab:minilm_sanity} summarizes this agreement quantitatively. Across benchmark-model pairs, \method{} scores based on MiniLM partitions show strong agreement with the model-sensitive scores (Pearson $r=0.897$), with a mean absolute difference of $0.109$. At the benchmark level, the mean \method{} values are also strongly correlated (Pearson $r=0.859$), and the cross-model variances remain reasonably aligned (Pearson $r=0.800$).

Overall, this sanity check suggests that the main MCQA conclusions are not driven solely by the model sensitivity of the partitioning step. While MiniLM partitions do not exactly reproduce the model-sensitive scores, they preserve the broad benchmark-level structure of the analysis.

\begin{table}[t]
\centering
\small
\begin{tabular}{lcc}
\toprule
Comparison level & Pearson $r$ & MAE \\
\midrule
Benchmark-model \method{} & 0.897 & 0.109 \\
Benchmark mean $\mu_H$ & 0.859 & 0.082 \\
Benchmark variance $\sigma_H^2$ & 0.800 & 0.019 \\
\bottomrule
\end{tabular}
\caption{
Agreement between model-sensitive predictive-similarity partitions and fixed model-agnostic MiniLM partitions on MCQA benchmarks. The first row compares individual \method{} scores over all benchmark-model pairs, while the latter rows compare benchmark-level summaries across MCQA benchmarks. 
}
\label{tab:minilm_sanity}
\end{table}

\section{\method{} Beyond Multiple-Choice Evaluation}
\label{app:beyond_mcqa}

We additionally evaluate \method{} on two non-MCQA benchmarks: GSM8K for open-ended mathematical reasoning and IFEval for instruction following. We induce fixed, model-agnostic partitions using MiniLM embeddings, yielding six subdomains for GSM8K and four for IFEval. For GSM8K, we extract and normalize the final numerical answer and use exact match, while for IFEval, we use binary pass or fail judgments from Llama-3.1-70B-Instruct.
\begin{table}[t]
\centering
\small
\resizebox{\columnwidth}{!}{%
\begin{tabular}{lcccc}
\toprule
Model & \multicolumn{2}{c}{IFEval} & \multicolumn{2}{c}{GSM8K} \\
& Score & \method{} & Score & \method{} \\
\midrule
Qwen3-4B      & 76.3 & 0.818 & 79.1 & 0.877 \\
Qwen3-8B      & 80.1 & 0.799 & 83.5 & 0.894 \\
Llama-3.2-3B  & 69.8 & 0.687 & 75.8 & 0.701 \\
Llama-3.1-8B  & 77.9 & 0.750 & 81.1 & 0.824 \\
\bottomrule
\end{tabular}%
}
\caption{\method{} on non-MCQA benchmarks using fixed model-agnostic partitions.}
\label{tab:non_mcqa}
\end{table}

As shown in Table~\ref{tab:non_mcqa}, \method{} provides information complementary to the aggregate score across both benchmarks. For example, Qwen3-8B improves over Qwen3-4B on IFEval in aggregate score (76.3 to 80.1), while \method{} decreases (0.818 to 0.799), indicating that the performance is less uniformly distributed across subdomains. These results demonstrate that \method{} can be applied with task-appropriate evaluation measures beyond multiple-choice accuracy. We evaluate on the full test sets, using base Qwen checkpoints for GSM8K and instruction-tuned Qwen checkpoints for IFEval, while using instruction-tuned Llama checkpoints for both tasks. Generation follows the settings recommended in the corresponding model cards and technical reports.

\section{Model-wise Decomposition of Benchmark \method{}}
\label{app:model_wise_dissected}
\begin{figure}
  \centering
\includegraphics[width=\linewidth]{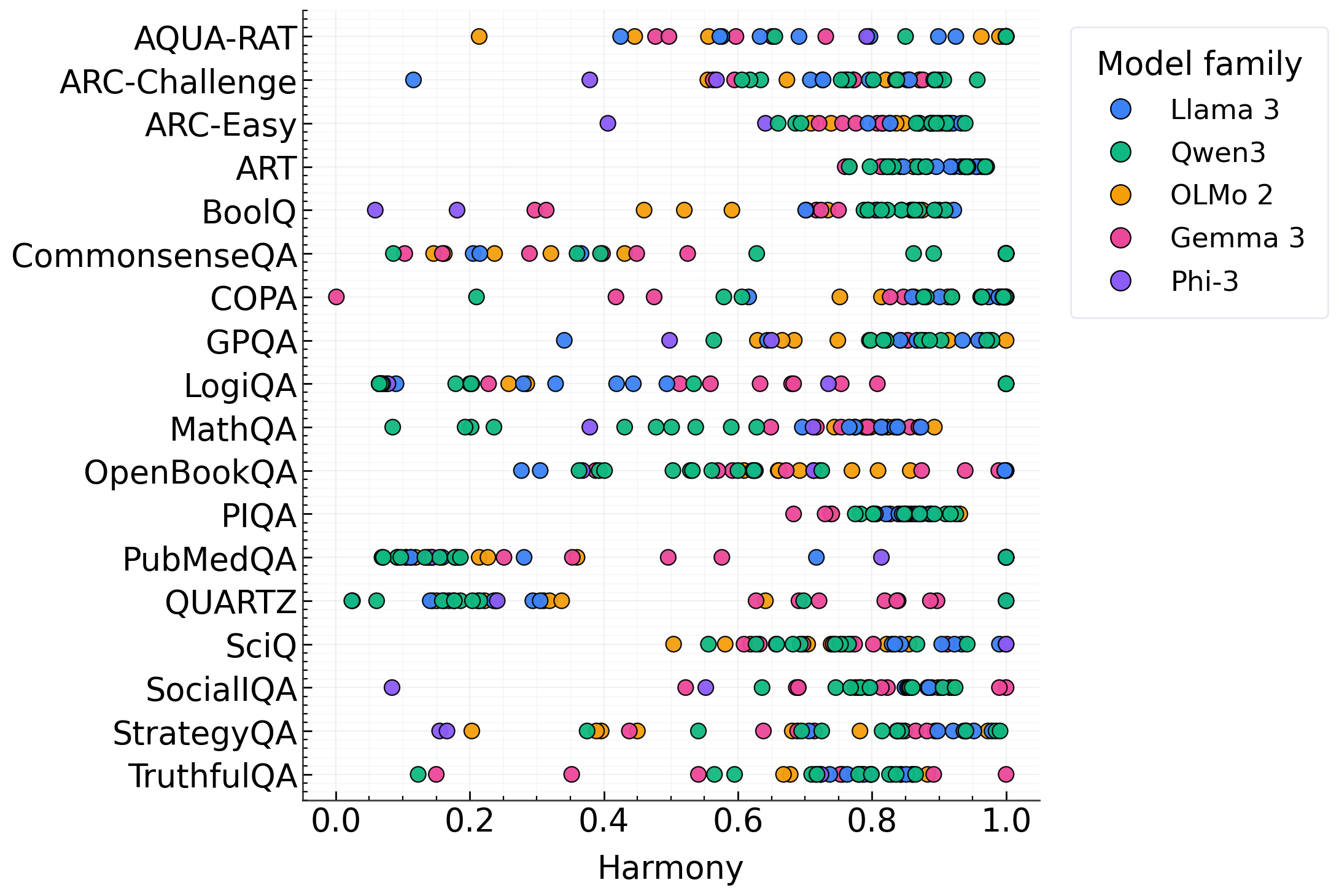}
  \caption{Model-wise decomposition of \method{} for MCQA benchmarks.}
  \label{fig:dissected_non_mmlu}
\end{figure}
In \S\ref{subsec:mapping}, we position each benchmark \(\mathcal{B}\) using the cross-model mean \(\mu_H(\mathcal{B})\) and variance \(\sigma_H^2(\mathcal{B})\). We now resolve this view at the model level. For each benchmark, Fig.~\ref{fig:dissected_non_mmlu} plots the per-model vector \(\{H_{\mathcal{B}}(f)\}_{f\in\mathcal{F}}\), revealing structure that is obscured by aggregation. 

Tight \emph{horizontal} groupings (small spread across models) indicate \emph{model-invariant} distributional balance, where different families assign similar \method{} to the same benchmark, suggesting that aggregate accuracy reflects uniform competence irrespective of architectural or training choices. Conversely, wide horizontal scatter exposes \emph{model-dependent robustness}, as some families concentrate performance on a few subsets (low \method{}), while others distribute performance more evenly (high \method{}).

We note that benchmarks with tight clusters are favorable for cross-family comparison, as accuracy rankings are less likely to be artifacts of benchmark composition. In contrast, wide scatter warns that leaderboard deltas may be driven by subsets that particular families exploit. In such cases, we suggest reporting accuracy alongside the \method{} profiles of the models under evaluation, \(\{H(\mathcal{G}_f)\}_{f}\).

\section{Assessing Benchmark Size as a Confound}
\label{app:bench_size_confound}

A natural concern is that benchmark size might confound our mean-variance characterization of Harmony. In this section, we show analytically that \method{} is invariant to uniform rescaling of benchmark size and empirically that benchmark size does not explain the characteristics observed in the mean-variance plane.

Recall that, given a partition $G_f = \{A_i\}_{i=1}^k$ of benchmark $\mathcal{B}$, \method{} is computed using the relative subset weights $w_i = |A_i| / |\mathcal{B}|$ and a normalized Shannon entropy over proximity-based scores, scaled by $\log k$. Because only the normalized weights $\{w_i\}$ and the relative distribution of performance across subdomains are present in the definition, operations that uniformly scale the benchmark size (e.g., duplicating all items or uniformly sub-sampling) should leave \method{} unchanged in principle. \method{} is therefore designed to reflect how coverage and performance are distributed across semantic subdomains in a manner that is not impacted by the number of examples given that subdomain proportions remain the same.

To assess benchmark size as a potential confound in our setting, we first compute the Pearson correlation between benchmark size and (i) the mean of \method{} across models and (ii) the variance of \method{} across models for the set of 18 MCQA benchmarks in our experimental setup. We obtain:
\[
\mathbb{E}_{f\sim\mathcal{F}}\!\left[ H_{\mathcal{B}}(f) \right]\text{ vs.\ size: } r = -0.310,\quad p = 0.2105
\]
\[
\mathrm{Var}_{f\sim\mathcal{F}}\!\left( H_{\mathcal{B}}(f) \right)\text{ vs.\ size: } r = 0.227,\quad p = 0.3653
\]
The associated $p$-values show no statistically significant evidence of linear dependence between benchmark size and either axis of the mean-variance plane. In particular, larger benchmarks do not exhibit systematic shifts in mean \method{} or in cross-model variance, indicating that benchmark size does not explain the structure observed in Figure~\ref{fig:combined_benchmark_fig}.

\begin{figure}
    \centering
    \includegraphics[width=\linewidth]{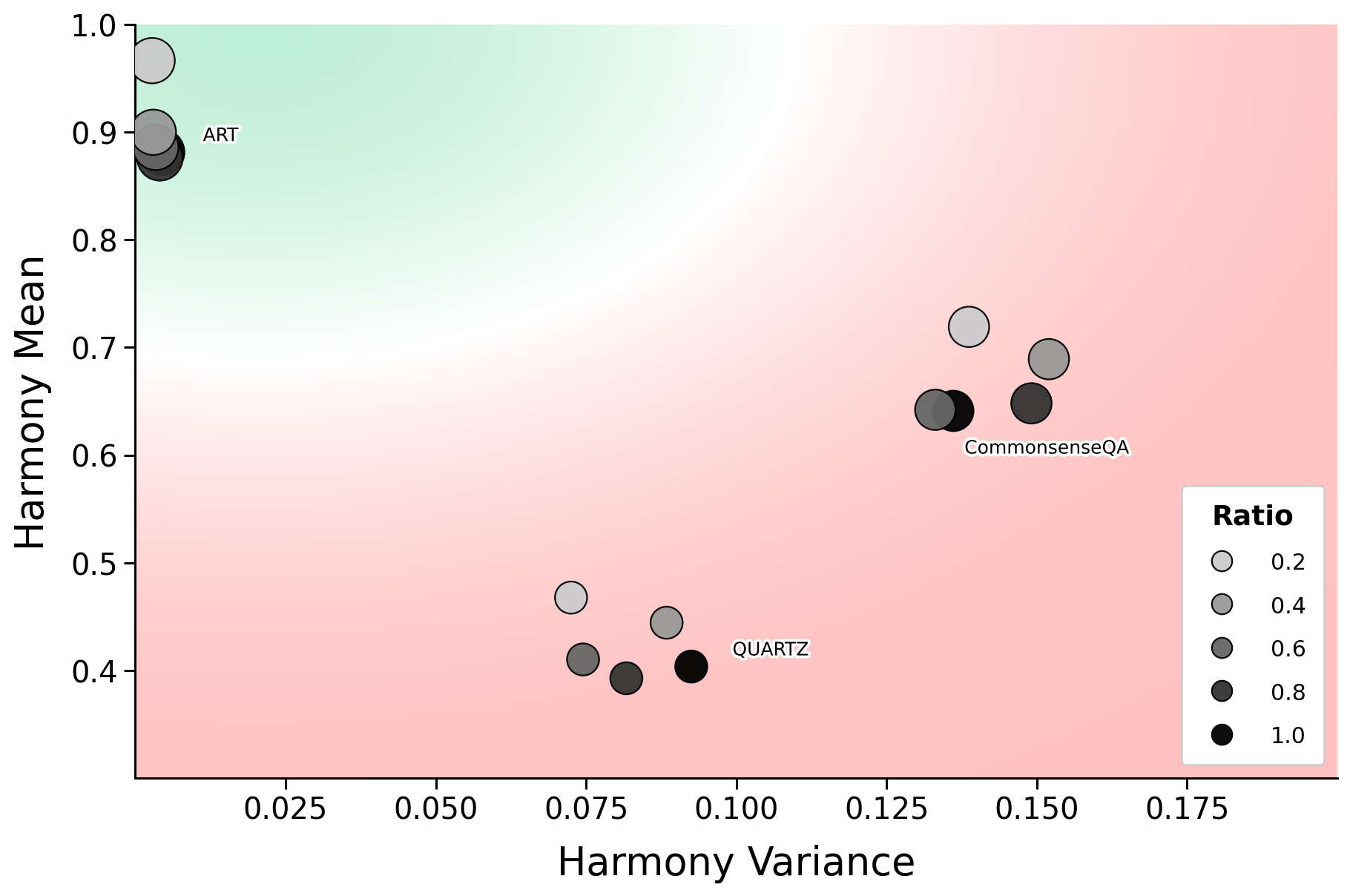}
    \caption{
        \textbf{Effect of benchmark size on \method.} For ART, QUARTZ, and CommonsenseQA, we down-sample at several ratios while preserving subdomain proportions and recompute \method{} across models. The resulting trajectories show only minor fluctuations in both mean and variance.
    }
    \label{fig:bench-size-confound}
\end{figure}

Second, we perform a controlled sub-sampling experiment on three benchmarks selected from different regions of the
mean-variance plane. For each benchmark, we apply several down-sampling ratios uniformly, hence preserving the relative frequencies of semantic subdomains. We then recompute \method{} across models. The resulting trajectories in the mean-variance plane are shown in Figure~\ref{fig:bench-size-confound}. Across all three benchmarks, we observe only minor fluctuations in both the mean and variance of Harmony, and the points remain within the same qualitative regions of the plane.

Taken together, the size-invariance of \method{} under uniform rescaling, the lack of statistically significant correlations
between benchmark size and \method{} statistics, and the stability observed in the sub-sampling experiment support our view that the mean-variance plane captures differences in \emph{distributional robustness}, rather than merely reflecting benchmark size.

\section{Assessing $k$ as a Confound}
\label{app:k_dependence}

Our construction of \method{} relies on a partition of each benchmark into $k$ semantic subdomains. A natural concern is that our choice of $k$ might introduce a confound, if $k$ systematically tracked \method{} in a way that would inflate or suppress our robustness measure. In this section, we empirically study $k$'s relationship to benchmark size and \method.

As described in Sec. \ref{subsec:methodology}, for each benchmark we sweep $k \in \{2, \dots, 20\}$ and select the value that maximizes the silhouette score, a well-established metric of clustering quality~\citep{ROUSSEEUW198753}. This procedure determines $k$ in a fully data-driven manner based on the geometry of the representation space, and $k$ is never hand-tuned to optimize \method.

To quantify how $k$ relates to benchmark size and \method{}, we compute Pearson correlations between (i) benchmark size and the selected $k$, and (ii) $k$ and the \method{} for each benchmark-model pair. We obtain:
\[
\text{\method{} vs.\ } k: \quad r = 0.274,\quad p = 0.2818,
\]
\[
\text{Benchmark size vs.\ } k: \quad r = 0.914,\quad p = 0.0000.
\]
The strong correlation between benchmark size and $k$ confirms the intuitive behavior that larger benchmarks admit more clusters. Crucially, however, there is no statistically significant linear relationship between $k$ and \method: the
correlation coefficient is weak and the associated $p$-value indicates no evidence against the null hypothesis of zero
correlation at standard significance levels.

This distinction matters because our conclusions rely on \method{} rather than $k$ or benchmark size itself. \method{} already accounts for varying cluster sizes through the weights $w_i = |A_i| / |B|$, so differences in the number of clusters primarily reflect how finely the benchmark can be partitioned, not a direct change in the robustness measure. The absence of a statistically significant correlation between $k$ and \method{} indicates that our robustness measure is not spuriously driven by the specific value of $k$, even though $k$ increases with benchmark size as expected.

\section{Assessing Pruning Ratio as a Confound}
\label{app:prun_rat_confound}
\begin{figure*}
    \centering
    \includegraphics[width=\linewidth]{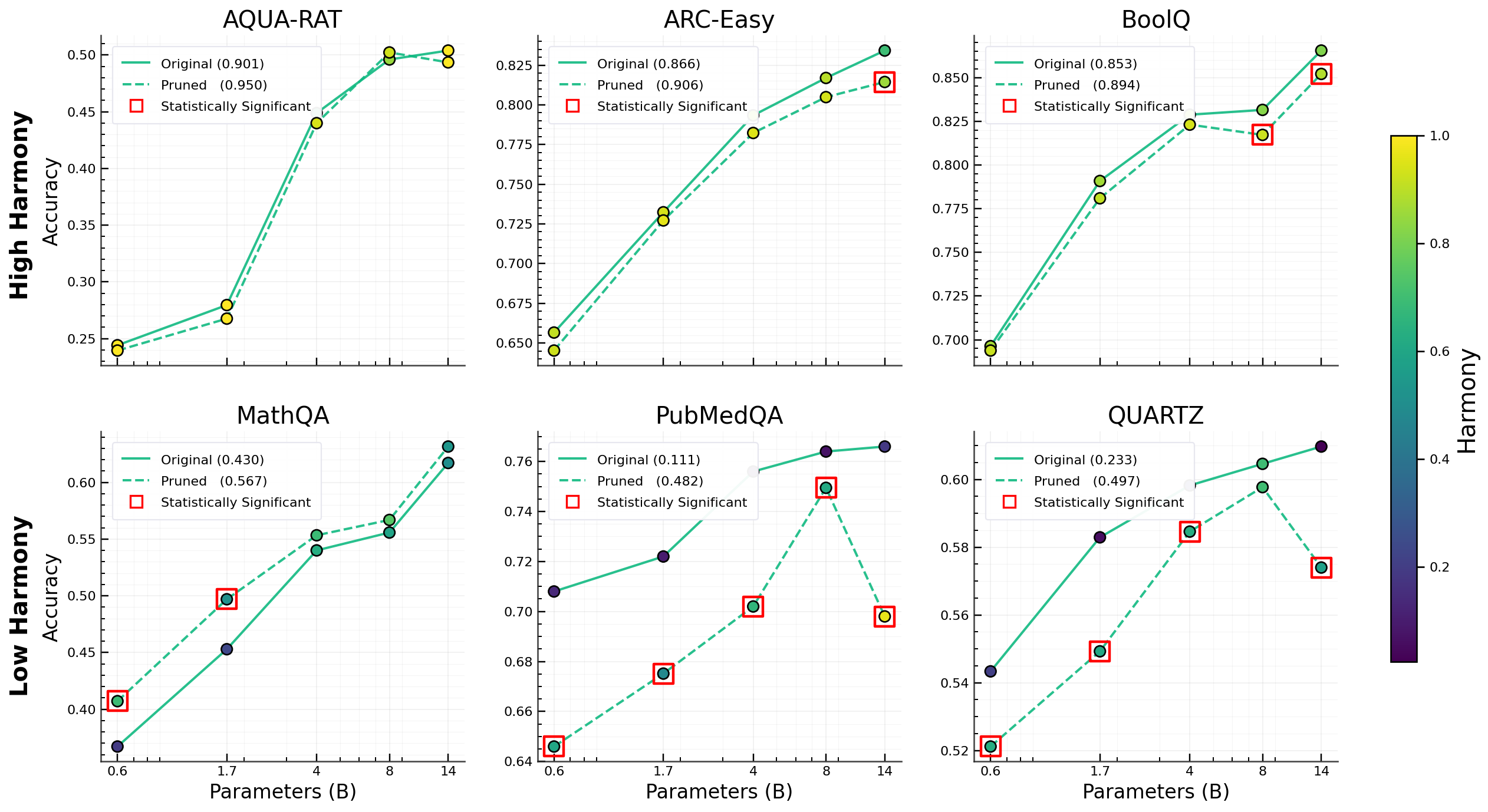}
    \caption{Pruning experiments replicated with pruning ratio $p=0.1$ for all benchmarks.}
    \label{fig:p01}
\end{figure*}

\begin{figure*}
    \centering
    \includegraphics[width=\linewidth]{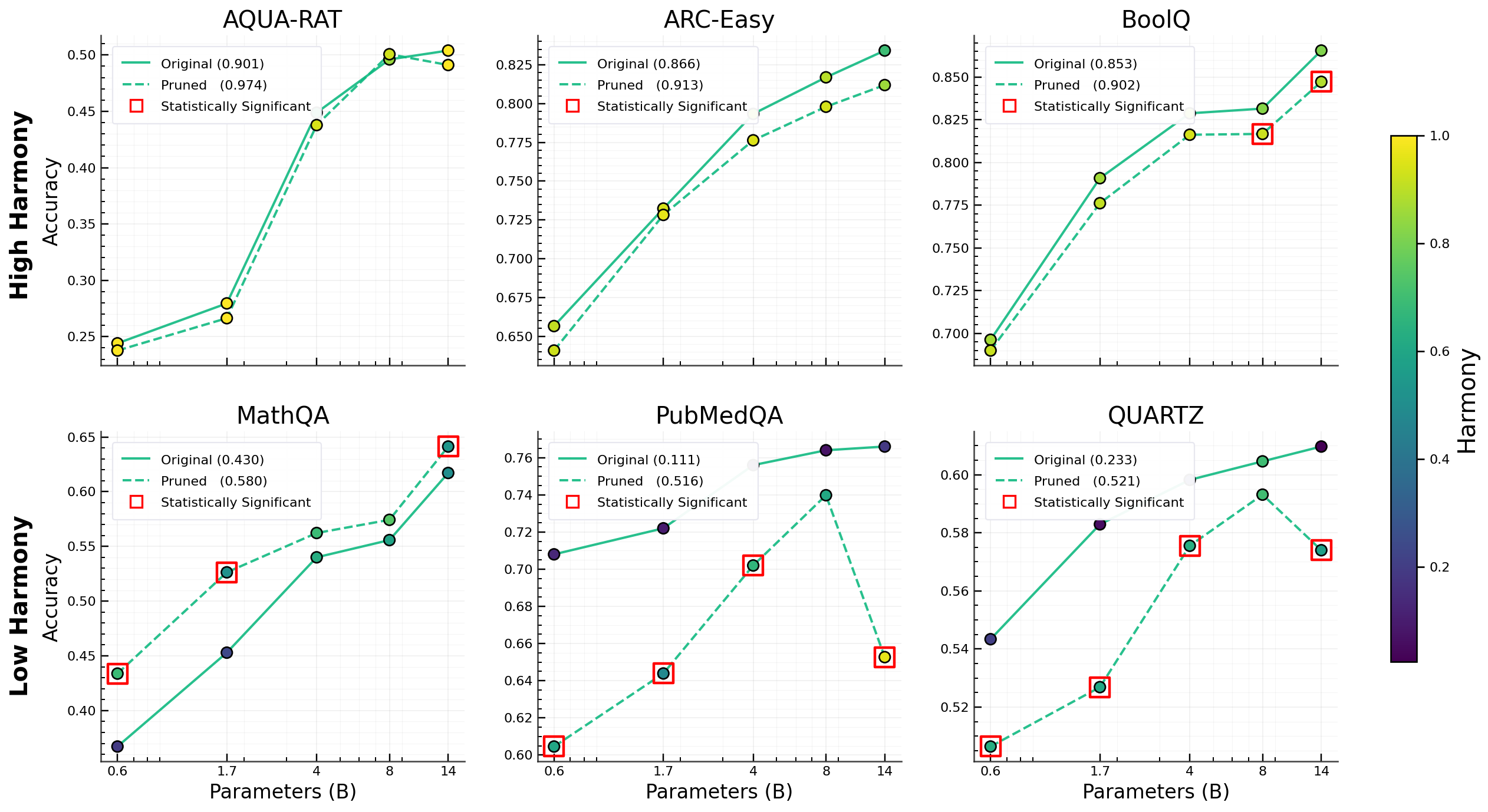}
    \caption{Pruning experiments replicated with pruning ratio $p=0.2$ for all benchmarks.}
    \label{fig:p02}
\end{figure*}

\begin{figure*}
    \centering
    \includegraphics[width=\linewidth]{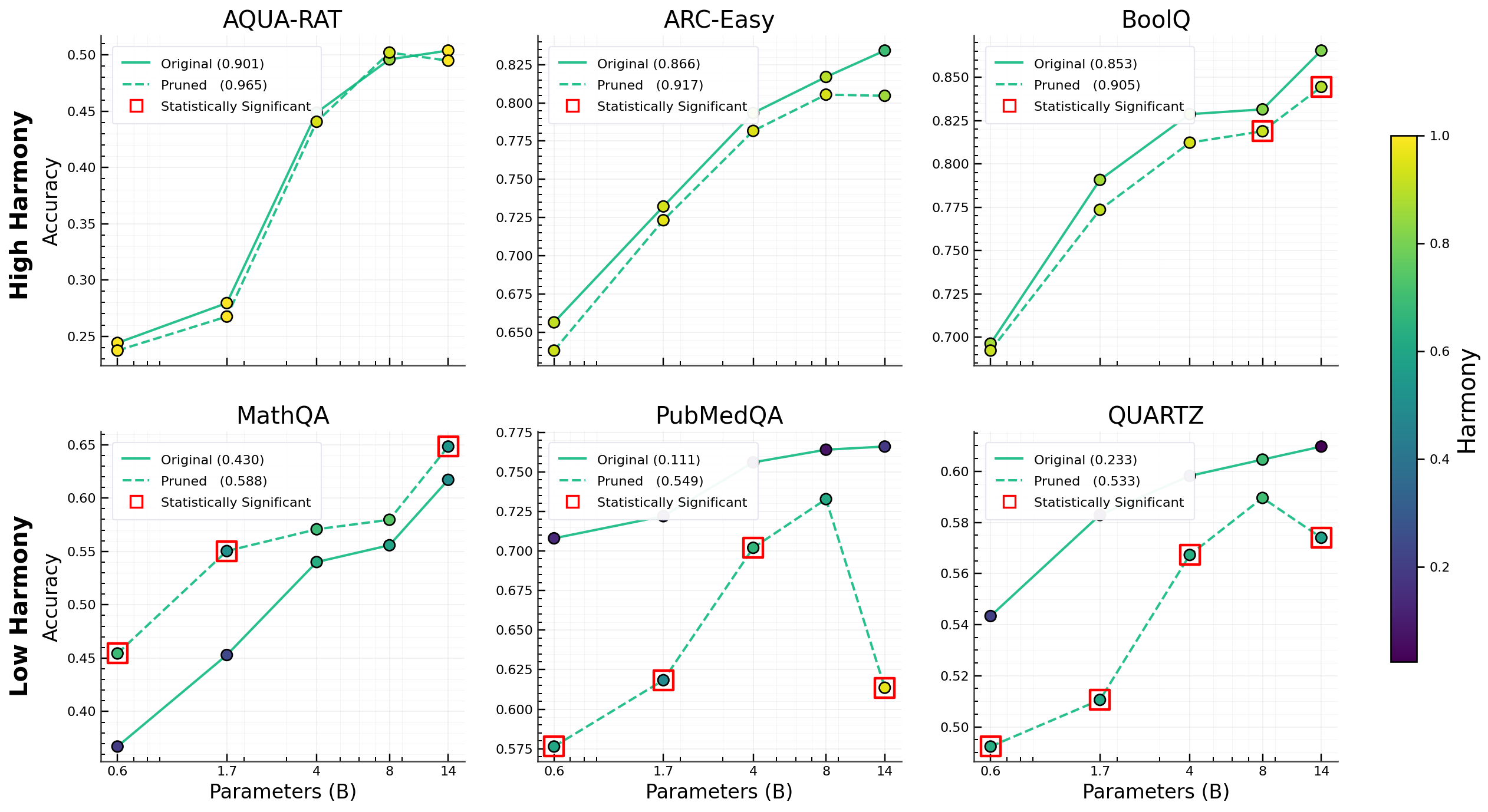}
    \caption{Pruning experiments replicated with pruning ratio $p=0.3$ for all benchmarks.}
    \label{fig:p03}
\end{figure*}

\begin{figure*}
    \centering
    \includegraphics[width=\linewidth]{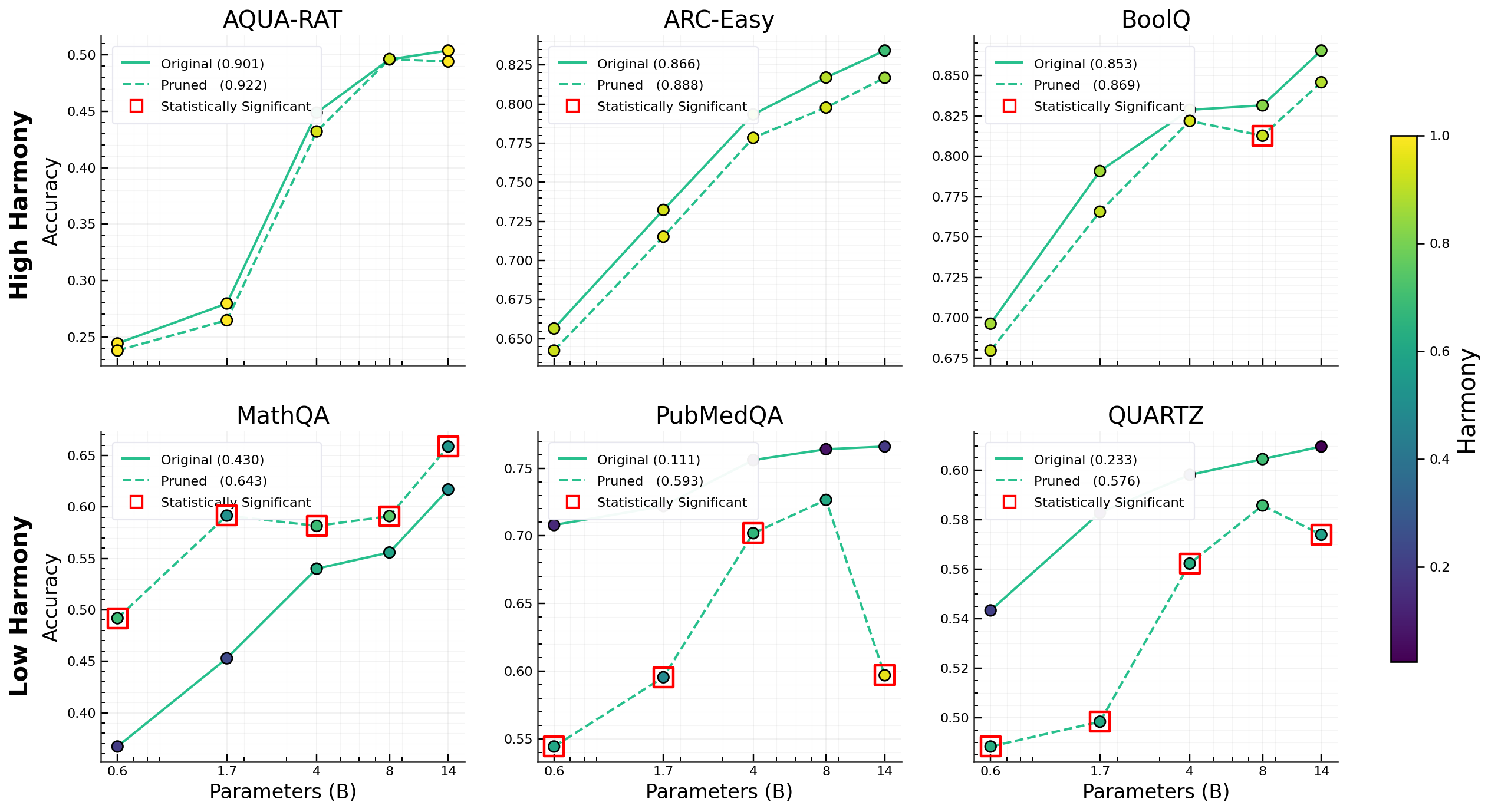}
    \caption{Pruning experiments replicated with pruning ratio $p=0.4$ for all benchmarks.}
    \label{fig:p04}
\end{figure*}

\begin{figure*}
    \centering
    \includegraphics[width=\linewidth]{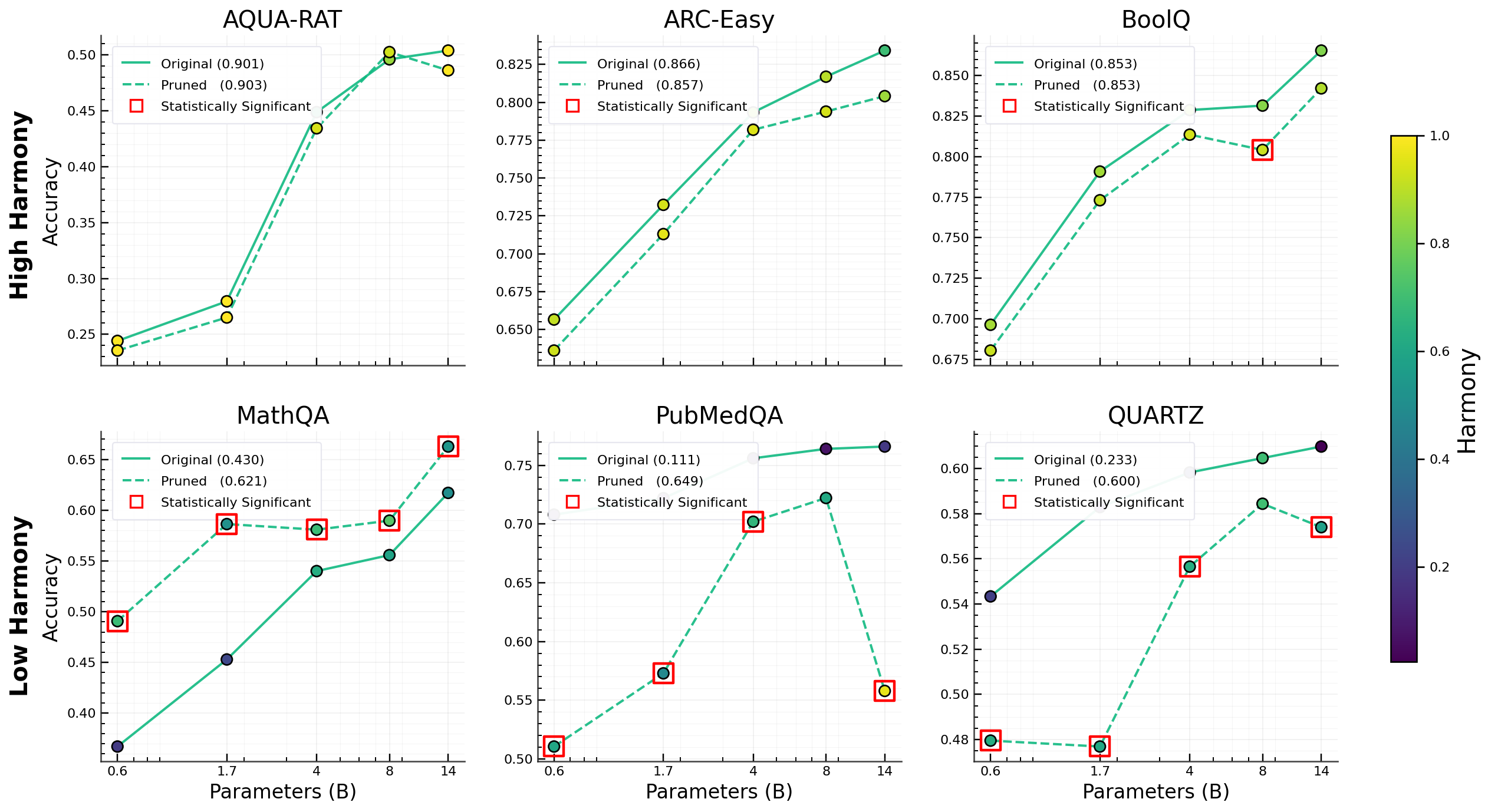}
    \caption{Pruning experiments replicated with pruning ratio $p=0.5$ for all benchmarks.}
    \label{fig:p05}
\end{figure*}

To explicitly rule out pruning ratio as a potential confound in the experiments of Sec.~\ref{sec:pruning}, we repeat the pruning procedure while holding the pruning budget fixed across all benchmarks. Concretely, we evaluate five ratios,
$p\in\{0.1,0.2,0.3,0.4,0.5\}$ and report the corresponding results in
Fig.~\ref{fig:p01}-\ref{fig:p05}.

Across all pruning ratios, we observe the same pattern: (i) pruning reliably increases \method, (ii) benchmarks with low \method{} exhibit statistically significant accuracy shifts, and (iii) benchmarks with high \method{} rarely show such shifts, even under more aggressive pruning. Although post-pruning accuracies change with the amount of data removed, these effects are stable across pruning budgets. Moreover, the magnitude of the accuracy change tends to saturate, as beyond moderate pruning (around $p=0.3$), further increases in the pruning ratio produce only marginal additional shifts. Taken together, these results indicate that the results in Figure \ref{fig:pruning} are not explained by pruning ratio alone and are robust to the choice of pruning budget.


\section{Details of Statistical Significance Test}
\label{sec:details_of_stat_sig}

We assess whether the subset we keep after pruning has a higher mean than the full set using a nonparametric, \emph{coupled} bootstrap sign test. Let $a_1,\dots,a_N$ be per-example accuracy and $k_i\in\{0,1\}$ indicate membership in the keep subset $K=\{i:\,k_i=1\}$. For $b=1,\dots,B$ with $B=10000$, we draw a bootstrap sample of indices $S^{(b)}$ of size $N$ (with replacement), compute both means on the \emph{same} resample, and then take their difference:
\begin{align}
\bar a_{\mathrm{all}}^{(b)} &= \frac{1}{N}\sum_{i\in S^{(b)}} a_i, \\
n_{\mathrm{keep}}^{(b)} &= \sum_{i\in S^{(b)}} k_i, \\
\bar a_{\mathrm{keep}}^{(b)} &= \frac{1}{n_{\mathrm{keep}}^{(b)}} \sum_{i\in S^{(b)}} k_i\, a_i, \\
\Delta^{(b)} &= \bar a_{\mathrm{keep}}^{(b)} - \bar a_{\mathrm{all}}^{(b)}.
\end{align}
Resamples with $n_{\mathrm{keep}}^{(b)}=0$ are discarded (to avoid degenerate runs we cap total draws at $3B$); let $m\le B$ be the number of valid differences retained. We then form a two-sided $p$-value from the sign statistic with a plus-one small-sample correction:
\begin{align}
r &= \sum_{b=1}^{m} \mathbf{1}\!\left\{\Delta^{(b)} \ge 0\right\}, \\
p &= \min\!\left\{\,1,\, 2\min\!\left(\frac{r+1}{m+1},\, \frac{m-r+1}{m+1}\right)\right\}.
\end{align}
We fix the random seed for reproducibility and declare significance at level $\alpha=0.05$ when $p<\alpha$ (testing $H_0:\mathbb{E}[\Delta]=0$ vs.\ $H_1:\mathbb{E}[\Delta]\neq 0$).

\section{Multi-Dimensional Evaluation}
\label{sec:multi_dim_eval}
Motivated by the skewed aggregate scores in low \method{} benchmarks, we conduct model evaluation at finer granularity. Following recent work on fine-grained evaluation \citep{zeng2025evaltree}, we recursively induce partitions as described in \S\ref{subsec:methodology}. This procedure yields a \emph{labeled tree}, where the root is the full benchmark; each internal node is a subset from the partitioning of its parent node; and leaves are atomic subdomains that admit no further valid split (see App.~\ref{sec:hier_labeling} for details). The resulting hierarchy enables interpretable, multi-dimensional evaluation, where each dimension corresponds to a subdomain of the benchmark.

We illustrate this approach on two examples: \emph{MMLU College Biology} and \emph{MMLU College Physics}. The average \method{} across Qwen3 models is markedly higher for biology (0.8538) than for physics (0.7534). This suggests that the aggregate accuracy for biology is a more representative reflection of the performance across subdomains. Indeed, Table~\ref{tab:mmlu_bio_clusters} shows that rankings in biology subdomains mirror the overall ordering: models that achieve higher overall accuracy also achieve higher accuracy in every subdomain. In other words, with one exception, models with superior overall performance are not surpassed by lower-performing models in any biology subdomain.\footnote{Comparison of 1.7B and 4B in \emph{Molecular \& Cellular Biology} is the only exception.} This alignment underscores that the aggregate score is a consistent and robust summary of subdomain performance in biology.

In contrast, physics exhibits lower \method{} and more notable divergences (Table~\ref{tab:mmlu_physics_clusters}). For example, Qwen3-4B lags behind Qwen3-14B in overall accuracy (58.8\% vs.\ 69.6\%), yet it surpasses it in \emph{Special Relativity} (71.4\% vs.\ 57.1\%) and \emph{Electromagnetism} (83.3\% vs.\ 66.7\%). Similarly, Qwen3-0.6B, despite its weak overall score (24.5\%), achieves competitive performance in \emph{Quantum Mechanics} (56.5\%), outperforming Qwen3-1.7B (34.8\%). These cases highlight how aggregate scores can obscure areas of relative strength, and how fine-grained, multi-dimensional evaluation reveals nuanced interpretation of model competence across subdomains.

We provide extended results for Qwen3 and Gemma 3 model families across 2 MCQA benchmarks and 6 MMLU subtasks in Appendix~\ref{sec:extended_multi_dim}.

\begin{table*}[t]
\centering
\resizebox{\textwidth}{!}{%
\begin{tabular}{lrrrrr}
\toprule
& Qwen3-0.6B & Qwen3-1.7B & Qwen3-4B & Qwen3-8B & Qwen3-14B \\
\midrule
Overall & 47.2 & 68.1 & 82.6 & 86.8 & \textbf{91.0} \\
\midrule
Multicellular Biology (42.4\%) & 45.9 & 68.9 & 88.5 & 88.5 & \textbf{93.4} \\
Evolutionary \& Ecological Processes (25.7\%) & 51.4 & 59.5 & 86.5 & \textbf{91.9} & \textbf{91.9} \\
Molecular \& Cellular Biology (31.9\%) & 45.7 & 73.9 & 71.7 & 80.4 & \textbf{87.0} \\
\midrule
\method{} & 0.951 & 0.918 & 0.757 & 0.784 & 0.859 \\
\bottomrule
\end{tabular}%
}
\caption{Multi-dimensional evaluation results of the Qwen3 model family in MMLU College Biology. \textbf{Bold} implies the best performance.}
\label{tab:mmlu_bio_clusters}
\end{table*}

\begin{table*}[t]
\centering
\resizebox{\textwidth}{!}{%
\begin{tabular}{lrrrrr}
\toprule
& Qwen3-0.6B & Qwen3-1.7B & Qwen3-4B & Qwen3-8B & Qwen3-14B \\
\midrule
Overall & 24.5 & 34.3 & 58.8 & 57.8 & \textbf{69.6} \\
\midrule
Quantum Mechanics Principles and Applications (22.5\%) & 56.5 & 34.8 & 65.2 & 69.6 & \textbf{73.9} \\
Thermodynamics (8.8\%) & 0.0 & 55.6 & 66.7 & 55.6 & \textbf{77.8} \\
Special Relativity Concepts (13.7\%) & 21.4 & 14.3 & \textbf{71.4} & 50.0 & 57.1 \\
Classical Physics Principles and Relationships (20.6\%) & 9.5 & 38.1 & 47.6 & 52.4 & \textbf{61.9} \\
Physics Phenomena and Application (16.7\%) & 0.0 & 11.8 & 11.8 & 17.6 & \textbf{58.8} \\
Electromagnetism (5.9\%) & 16.7 & 16.7 & \textbf{83.3} & \textbf{83.3} & 66.7 \\
Solid State Physics (11.8\%) & 50.0 & 75.0 & \textbf{100.0} & \textbf{100.0} & \textbf{100.0} \\
\midrule
\method{} & 0.686 & 0.712 & 0.765 & 0.815 & 0.789  \\
\bottomrule
\end{tabular}%
}
\caption{Multi-dimensional evaluation results of the Qwen3 model family in MMLU College Physics. \textbf{Bold} implies the best performance.}
\label{tab:mmlu_physics_clusters}
\end{table*}

\section{Extended Results: Multi-Dimensional Evaluation}
\label{sec:extended_multi_dim}

We provide the extended results of multi-dimensional evaluation conducted as described in Appendix~\ref{sec:multi_dim_eval}. Our setup consists of Qwen3 and Gemma 3 model families and ARC-Easy, BoolQ, MMLU Anatomy, MMLU College Biology, MMLU College Computer Science, MMLU College Mathematics, MMLU College Physics, MMLU High School US History.

\subsection{Qwen3 Family}
See Tables~\ref{tab:qwen3_arce}, \ref{tab:qwen3_boolq}, \ref{tab:qwen3_anatomy}, \ref{tab:qwen3_col_bio}, \ref{tab:qwen3_col_cs}, \ref{tab:qwen3_col_math}, \ref{tab:qwen3_col_phy}, and \ref{tab:qwen3_hs_us_hist} for the extended results of the Qwen3 model family.

\begin{table*}[t]
\centering
\resizebox{\textwidth}{!}{%
\begin{tabular}{lrrrrr}
\toprule
 & \textbf{Qwen3-0.6B} & \textbf{Qwen3-1.7B} & \textbf{Qwen3-4B} & \textbf{Qwen3-8B} & \textbf{Qwen3-14B} \\
\midrule
Overall & 60.7 & 72.2 & 80.5 & 83.5 & 84.2 \\
\midrule
Geology and Earth Sciences (15.0\%) & 64.4 & 73.4 & 83.8 & 83.8 & 83.8 \\
Scientific Principles and Processes (32.6\%) & 56.3 & 70.1 & 79.9 & 83.5 & 84.3 \\
Biological Processes and Concepts (24.1\%) & 65.4 & 74.0 & 78.8 & 82.9 & 83.6 \\
Physics Principles in Engineering and Science (9.4\%) & 64.3 & 75.9 & 82.1 & 86.6 & 86.6 \\
Environmental and Energy Assessment (6.5\%) & 60.4 & 63.6 & 78.6 & 81.2 & 79.9 \\
Fundamental Concepts in Astronomy (7.6\%) & 56.1 & 76.7 & 80.6 & 83.3 & 87.2 \\
Fundamentals of Chemical and Material Properties (4.8\%) & 56.1 & 71.9 & 81.6 & 84.2 & 84.2 \\
\bottomrule
\end{tabular}}
\caption{Multi-dimensional evaluation results for the Qwen3 model family on ARC-Easy.}
\label{tab:qwen3_arce}
\end{table*}

\begin{table*}[t]
\centering
\resizebox{\textwidth}{!}{%
\begin{tabular}{lrrrrr}
\toprule
 & \textbf{Qwen3-0.6B} & \textbf{Qwen3-1.7B} & \textbf{Qwen3-4B} & \textbf{Qwen3-8B} & \textbf{Qwen3-14B} \\
\midrule
Overall & 64.1 & 77.5 & 85.0 & 86.6 & 89.3 \\
\midrule
Product Composition, Properties, and Standards (8.8\%) & 66.8 & 80.6 & 83.7 & 86.9 & 88.9 \\
Geographic, Operational, and Temporal Analysis (11.0\%) & 69.3 & 77.3 & 85.3 & 88.4 & 91.7 \\
Media Standards and Analysis (24.8\%) & 61.3 & 77.7 & 86.8 & 88.7 & 91.0 \\
Scientific and Analytical Principles (9.0\%) & 67.9 & 81.2 & 83.3 & 84.6 & 89.1 \\
Sports History and Regulations (10.1\%) & 63.5 & 75.1 & 82.1 & 84.5 & 87.2 \\
Governmental Laws and Regulations (10.6\%) & 64.1 & 72.8 & 80.3 & 84.6 & 86.1 \\
Human Biology and Medical Science (6.5\%) & 60.6 & 81.2 & 87.8 & 85.4 & 87.8 \\
Economic Systems (8.3\%) & 61.6 & 75.3 & 85.2 & 85.6 & 87.5 \\
Sociocultural, Geopolitical, and Linguistic Analysis (7.1\%) & 63.9 & 76.8 & 88.0 & 86.7 & 89.7 \\
Fictional Narrative Analysis and Elements (3.8\%) & 64.8 & 80.8 & 89.6 & 88.0 & 93.6 \\
\bottomrule
\end{tabular}}
\caption{Evaluation results for the Qwen3 model family on BoolQ.}
\label{tab:qwen3_boolq}
\end{table*}

\begin{table*}[t]
\centering
\resizebox{\textwidth}{!}{%
\begin{tabular}{lrrrrr}
\toprule
 & \textbf{Qwen3-0.6B} & \textbf{Qwen3-1.7B} & \textbf{Qwen3-4B} & \textbf{Qwen3-8B} & \textbf{Qwen3-14B} \\
\midrule
Overall & 37.8 & 56.3 & 62.2 & 71.1 & 80.7 \\
\midrule
General Human Anatomy, Physiology, and Terminology (31.9\%) & 51.2 & 67.4 & 79.1 & 76.7 & 86.0 \\
Head and Neck Anatomy (21.5\%) & 20.7 & 34.5 & 34.5 & 62.1 & 82.8 \\
Skeletal Development and Anatomy (14.1\%) & 26.3 & 42.1 & 47.4 & 52.6 & 57.9 \\
Neurological Disorders (10.4\%) & 21.4 & 50.0 & 50.0 & 71.4 & 78.6 \\
Bone Anatomy and Terminology (3.0\%) & 0.0 & 50.0 & 50.0 & 50.0 & 50.0 \\
Anatomy of Circulatory System (2.2\%) & 66.7 & 100.0 & 100.0 & 100.0 & 100.0 \\
Developmental Structures (6.7\%) & 22.2 & 44.4 & 55.6 & 77.8 & 88.9 \\
Nephrology (10.4\%) & 78.6 & 92.9 & 100.0 & 92.9 & 92.9 \\
\bottomrule
\end{tabular}}
\caption{Evaluation results for the Qwen3 model family on MMLU Anatomy.}
\label{tab:qwen3_anatomy}
\end{table*}

\begin{table*}[t]
\centering
\resizebox{\textwidth}{!}{%
\begin{tabular}{lrrrrr}
\toprule
 & \textbf{Qwen3-0.6B} & \textbf{Qwen3-1.7B} & \textbf{Qwen3-4B} & \textbf{Qwen3-8B} & \textbf{Qwen3-14B} \\
\midrule
Overall & 47.2 & 68.1 & 82.6 & 86.8 & 91.0 \\
\midrule
Multicellular Biology (42.4\%) & 45.9 & 68.9 & 88.5 & 88.5 & 93.4 \\
Evolutionary and Ecological Processes (25.7\%) & 51.4 & 59.5 & 86.5 & 91.9 & 91.9 \\
Molecular and Cellular Biology (31.9\%) & 45.7 & 73.9 & 71.7 & 80.4 & 87.0 \\
\bottomrule
\end{tabular}}
\caption{Evaluation results for the Qwen3 model family on MMLU College Biology.}
\label{tab:qwen3_col_bio}
\end{table*}

\begin{table*}[t]
\centering
\resizebox{\textwidth}{!}{%
\begin{tabular}{lrrrrr}
\toprule
 & \textbf{Qwen3-0.6B} & \textbf{Qwen3-1.7B} & \textbf{Qwen3-4B} & \textbf{Qwen3-8B} & \textbf{Qwen3-14B} \\
\midrule
Overall & 28.0 & 41.0 & 66.0 & 72.0 & 68.0 \\
\midrule
Theoretical Foundations of Computation (52.0\%) & 26.9 & 40.4 & 59.6 & 69.2 & 65.4 \\
Computer Architecture and Optimization (7.0\%) & 28.6 & 42.9 & 71.4 & 57.1 & 57.1 \\
Operating Systems (10.0\%) & 50.0 & 80.0 & 90.0 & 90.0 & 80.0 \\
Network Layer Protocols and Technologies (5.0\%) & 60.0 & 60.0 & 80.0 & 100.0 & 100.0 \\
Data Processing (12.0\%) & 8.3 & 8.3 & 66.7 & 50.0 & 66.7 \\
Sorting Algorithms (4.0\%) & 25.0 & 25.0 & 75.0 & 100.0 & 100.0 \\
Graph Algorithms and Data Structures (10.0\%) & 20.0 & 40.0 & 60.0 & 80.0 & 50.0 \\
\bottomrule
\end{tabular}}
\caption{Evaluation results for the Qwen3 model family on MMLU College Computer Science.}
\label{tab:qwen3_col_cs}
\end{table*}

\begin{table*}[t]
\centering
\resizebox{\textwidth}{!}{%
\begin{tabular}{lrrrrr}
\toprule
 & \textbf{Qwen3-0.6B} & \textbf{Qwen3-1.7B} & \textbf{Qwen3-4B} & \textbf{Qwen3-8B} & \textbf{Qwen3-14B} \\
\midrule
Overall & 31.0 & 39.0 & 55.0 & 59.0 & 68.0 \\
\midrule
Advanced Real Analysis (19.0\%) & 26.3 & 26.3 & 42.1 & 52.6 & 52.6 \\
Abstract Algebra (11.0\%) & 27.3 & 45.5 & 90.9 & 63.6 & 81.8 \\
Probability (7.0\%) & 28.6 & 71.4 & 28.6 & 57.1 & 57.1 \\
Properties of Mathematical Operations and Functions (5.0\%) & 20.0 & 20.0 & 40.0 & 80.0 & 80.0 \\
Advanced Mathematical Concepts and Applications (16.0\%) & 31.3 & 37.5 & 75.0 & 56.3 & 75.0 \\
Mathematical Modeling and Algorithms (10.0\%) & 30.0 & 50.0 & 60.0 & 60.0 & 70.0 \\
Multivariable Calculus (27.0\%) & 25.9 & 37.0 & 40.7 & 59.3 & 63.0 \\
Mathematical Optimization Methods (5.0\%) & 100.0 & 40.0 & 80.0 & 60.0 & 100.0 \\
\bottomrule
\end{tabular}}
\caption{Evaluation results for the Qwen3 model family on MMLU College Mathematics.}
\label{tab:qwen3_col_math}
\end{table*}

\begin{table*}[t]
\centering
\resizebox{\textwidth}{!}{%
\begin{tabular}{lrrrrr}
\toprule
 & \textbf{Qwen3-0.6B} & \textbf{Qwen3-1.7B} & \textbf{Qwen3-4B} & \textbf{Qwen3-8B} & \textbf{Qwen3-14B} \\
\midrule
Overall & 24.5 & 34.3 & 58.8 & 57.8 & 69.6 \\
\midrule
Quantum Mechanics Principles and Applications (22.5\%) & 56.5 & 34.8 & 65.2 & 69.6 & 73.9 \\
Thermodynamics (8.8\%) & 0.0 & 55.6 & 66.7 & 55.6 & 77.8 \\
Special Relativity Concepts (13.7\%) & 21.4 & 14.3 & 71.4 & 50.0 & 57.1 \\
Classical Physics Principles and Relationships (20.6\%) & 9.5 & 38.1 & 47.6 & 52.4 & 61.9 \\
Physics Phenomena and Applications (16.7\%) & 0.0 & 11.8 & 11.8 & 17.6 & 58.8 \\
Electromagnetism (5.9\%) & 16.7 & 16.7 & 83.3 & 83.3 & 66.7 \\
Solid State Physics Concepts (11.8\%) & 50.0 & 75.0 & 100.0 & 100.0 & 100.0 \\
\bottomrule
\end{tabular}}
\caption{Evaluation results for the Qwen3 model family on MMLU College Physics.}
\label{tab:qwen3_col_phy}
\end{table*}

\begin{table*}[t]
\centering
\resizebox{\textwidth}{!}{%
\begin{tabular}{lrrrrr}
\toprule
 & \textbf{Qwen3-0.6B} & \textbf{Qwen3-1.7B} & \textbf{Qwen3-4B} & \textbf{Qwen3-8B} & \textbf{Qwen3-14B} \\
\midrule
Overall & 52.5 & 66.2 & 83.3 & 88.7 & 91.7 \\
\midrule
US Sociopolitical Ideologies, Movements, and Issues (46.1\%) & 52.1 & 64.9 & 83.0 & 90.4 & 95.7 \\
Progressive Era Economic and Social Initiatives (3.9\%) & 75.0 & 87.5 & 100.0 & 100.0 & 100.0 \\
United States Governance and Politics (37.3\%) & 46.1 & 61.8 & 81.6 & 85.5 & 88.2 \\
Ideological and Territorial Expansion in the Americas (9.8\%) & 70.0 & 80.0 & 90.0 & 95.0 & 85.0 \\
American History Eras (2.9\%) & 50.0 & 66.7 & 66.7 & 66.7 & 83.3 \\
\bottomrule
\end{tabular}}
\caption{Evaluation results for the Qwen3 model family on MMLU High School U.S. History.}
\label{tab:qwen3_hs_us_hist}
\end{table*}

\subsection{Gemma 3 Family}
See Tables~\ref{tab:llama_arce}, \ref{tab:llama_boolq}, \ref{tab:llama_anat}, \ref{tab:llama_col_bio}, \ref{tab:llama_col_cs}, \ref{tab:llama_col_math}, \ref{tab:llama_col_phy}, and \ref{tab:llama_hs_us_hist} for the extended results of Gemma 3 model family.

\begin{table*}[t]
\centering
\resizebox{\textwidth}{!}{%
\begin{tabular}{lrrrr}
\toprule
& \textbf{Gemma 3 1B} & \textbf{Gemma 3 4B} & \textbf{Gemma 3 12B} & \textbf{Gemma 3 27B}\\
\midrule
Overall & 72.0 & 81.6 & 87.2 & 87.5\\
\midrule
Geology and Earth Sciences (15.0\%) & 74.5 & 84.6 & 87.7 & 89.4\\
Scientific Principles and Processes (32.6\%) & 71.2 & 79.7 & 86.5 & 87.0\\
Biological Processes and Concepts (24.1\%) & 74.3 & 83.0 & 88.5 & 87.8\\
Physics Principles in Engineering and Science (9.4\%) & 70.5 & 83.0 & 88.8 & 88.8\\
Environmental and Energy Assessment (6.5\%) & 68.2 & 80.5 & 83.1 & 84.4\\
Fundamental Concepts in Astronomy (7.6\%) & 72.8 & 79.4 & 86.7 & 87.2\\
Fundamentals of Chemical and Material Properties (4.8\%) & 64.0 & 79.8 & 86.8 & 86.8\\
\bottomrule
\end{tabular}}
\caption{Multi-dimensional evaluation results for the Gemma 3 model family on ARC-Easy.}
\label{tab:llama_arce}
\end{table*}

\begin{table*}[t]
\centering
\resizebox{\textwidth}{!}{%
\begin{tabular}{lrrrr}
\toprule
& \textbf{Gemma 3 1B} & \textbf{Gemma 3 4B} & \textbf{Gemma 3 12B} & \textbf{Gemma 3 27B}\\
\midrule
Overall & 66.5 & 79.0 & 85.3 & 87.1\\
\midrule
Product Composition, Properties, and Standards (8.8\%) & 67.5 & 76.1 & 82.4 & 88.6\\
Geographic, Operational, and Temporal Analysis (11.0\%) & 67.6 & 83.7 & 84.8 & 87.3\\
Media Standards and Analysis (24.8\%) & 66.7 & 80.6 & 88.5 & 89.6\\
Scientific and Analytical Principles (9.0\%) & 65.2 & 75.1 & 82.9 & 86.0\\
Sports History and Regulations (10.1\%) & 63.2 & 78.7 & 83.0 & 86.0\\
Governmental Laws and Regulations (10.6\%) & 64.1 & 75.7 & 80.3 & 82.0\\
Human Biology and Medical Science (6.5\%) & 77.0 & 82.6 & 86.9 & 89.2\\
Economic Systems (8.3\%) & 66.1 & 77.1 & 85.6 & 85.6\\
Sociocultural, Geopolitical, and Linguistic Analysis (7.1\%) & 61.8 & 78.1 & 87.6 & 84.1\\
Fictional Narrative Analysis and Elements (3.8\%) & 71.2 & 79.2 & 91.2 & 90.4\\
\bottomrule
\end{tabular}}
\caption{Multi-dimensional evaluation results for the Gemma 3 model family on BoolQ.}
\label{tab:llama_boolq}
\end{table*}

\begin{table*}[t]
\centering
\resizebox{\textwidth}{!}{%
\begin{tabular}{lrrrr}
\toprule
& \textbf{Gemma 3 1B} & \textbf{Gemma 3 4B} & \textbf{Gemma 3 12B} & \textbf{Gemma 3 27B}\\
\midrule
Overall & 25.9 & 61.5 & 70.4 & 70.4\\
\midrule
General Human Anatomy, Physiology, and Terminology (31.9\%) & 27.9 & 69.8 & 83.7 & 79.1\\
Head and Neck Anatomy (21.5\%) & 27.6 & 41.4 & 55.2 & 48.3\\
Skeletal Development and Anatomy (14.1\%) & 21.1 & 52.6 & 52.6 & 57.9\\
Neurological Disorders (10.4\%) & 42.9 & 64.3 & 57.1 & 71.4\\
Bone Anatomy and Terminology (3.0\%) & 50.0 & 50.0 & 50.0 & 50.0\\
Anatomy of Circulatory System (2.2\%) & 0.0 & 100.0 & 100.0 & 100.0\\
Developmental Structures (6.7\%) & 11.1 & 44.4 & 77.8 & 88.9\\
Nephrology (10.4\%) & 14.3 & 92.9 & 92.9 & 92.9\\
\bottomrule
\end{tabular}}
\caption{Multi-dimensional evaluation results for the Gemma 3 model family on MMLU Anatomy.}
\label{tab:llama_anat}
\end{table*}

\begin{table*}[t]
\centering
\resizebox{\textwidth}{!}{%
\begin{tabular}{lrrrr}
\toprule
& \textbf{Gemma 3 1B} & \textbf{Gemma 3 4B} & \textbf{Gemma 3 12B} & \textbf{Gemma 3 27B}\\
\midrule
Multicellular Biology (42.4\%) & 27.9 & 68.9 & 98.4 & 93.4\\
Evolutionary and Ecological Processes (25.7\%) & 21.6 & 67.6 & 86.5 & 86.5\\
Molecular and Cellular Biology (31.9\%) & 21.7 & 65.2 & 87.0 & 91.3\\
\bottomrule
\end{tabular}}
\caption{Multi-dimensional evaluation results for the Gemma 3 model family on MMLU College Biology.}
\label{tab:llama_col_bio}
\end{table*}

\begin{table*}[t]
\centering
\resizebox{\textwidth}{!}{%
\begin{tabular}{lrrrr}
\toprule
& \textbf{Gemma 3 1B} & \textbf{Gemma 3 4B} & \textbf{Gemma 3 12B} & \textbf{Gemma 3 27B}\\
\midrule
Overall & 30.0 & 48.0 & 57.0 & 63.0\\
\midrule
Theoretical Foundations of Computation (52.0\%) & 36.5 & 38.5 & 51.9 & 61.5\\
Computer Architecture and Optimization (7.0\%) & 28.6 & 57.1 & 71.4 & 57.1\\
Operating Systems (10.0\%) & 20.0 & 60.0 & 80.0 & 80.0\\
Network Layer Protocols and Technologies (5.0\%) & 20.0 & 60.0 & 100.0 & 100.0\\
Data Processing (12.0\%) & 25.0 & 33.3 & 41.7 & 41.7\\
Sorting Algorithms (4.0\%) & 0.0 & 100.0 & 100.0 & 75.0\\
Graph Algorithms and Data Structures (10.0\%) & 30.0 & 70.0 & 30.0 & 60.0\\
\bottomrule
\end{tabular}}
\caption{Multi-dimensional evaluation results for the Gemma 3 model family on MMLU College Computer Science.}
\label{tab:llama_col_cs}
\end{table*}

\begin{table*}[t]
\centering
\resizebox{\textwidth}{!}{%
\begin{tabular}{lrrrr}
\toprule
& \textbf{Gemma 3 1B} & \textbf{Gemma 3 4B} & \textbf{Gemma 3 12B} & \textbf{Gemma 3 27B}\\
\midrule
Overall & 33.0 & 41.0 & 50.0 & 58.0\\
\midrule
Advanced Real Analysis (19.0\%) & 57.9 & 52.6 & 47.4 & 52.6\\
Abstract Algebra (11.0\%) & 27.3 & 72.7 & 63.6 & 54.5\\
Probability (7.0\%) & 14.3 & 57.1 & 42.9 & 57.1\\
Properties of Mathematical Operations and Functions (5.0\%) & 60.0 & 40.0 & 40.0 & 80.0\\
Advanced Mathematical Concepts and Applications (16.0\%) & 25.0 & 31.3 & 56.3 & 68.8\\
Mathematical Modeling and Algorithms (10.0\%) & 20.0 & 30.0 & 50.0 & 60.0\\
Multivariable Calculus (27.0\%) & 29.6 & 25.9 & 40.7 & 55.6\\
Mathematical Optimization Methods (5.0\%) & 20.0 & 40.0 & 80.0 & 40.0\\
\bottomrule
\end{tabular}}
\caption{Multi-dimensional evaluation results for the Gemma 3 model family on MMLU College Mathematics.}
\label{tab:llama_col_math}
\end{table*}

\begin{table*}[t]
\centering
\resizebox{\textwidth}{!}{%
\begin{tabular}{lrrrr}
\toprule
& \textbf{Gemma 3 1B} & \textbf{Gemma 3 4B} & \textbf{Gemma 3 12B} & \textbf{Gemma 3 27B}\\
\midrule
Overall & 20.6 & 41.2 & 52.9 & 63.7\\
\midrule
Quantum Mechanics Principles and Applications (22.5\%) & 8.7 & 39.1 & 43.5 & 73.9\\
Thermodynamics (8.8\%) & 11.1 & 22.2 & 55.6 & 55.6\\
Special Relativity Concepts (13.7\%) & 28.6 & 35.7 & 42.9 & 50.0\\
Classical Physics Principles and Relationships (20.6\%) & 42.9 & 28.6 & 57.1 & 66.7\\
Physics Phenomena and Applications (16.7\%) & 5.9 & 41.2 & 29.4 & 41.2\\
Electromagnetism (5.9\%) & 0.0 & 66.7 & 83.3 & 50.0\\
Solid State Physics Concepts (11.8\%) & 33.3 & 75.0 & 91.7 & 100.0\\
\bottomrule
\end{tabular}}
\caption{Multi-dimensional evaluation results for the Gemma 3 model family on MMLU College Physics.}
\label{tab:llama_col_phy}
\end{table*}

\begin{table*}[t]
\centering
\resizebox{\textwidth}{!}{%
\begin{tabular}{lrrrr}
\toprule
& \textbf{Gemma 3 1B} & \textbf{Gemma 3 4B} & \textbf{Gemma 3 12B} & \textbf{Gemma 3 27B}\\
\midrule
Overall & 26.0 & 75.5 & 88.2 & 91.2\\
\midrule
US Sociopolitical Ideologies, Movements, and Issues (46.1\%) & 26.6 & 80.9 & 87.2 & 92.6\\
Progressive Era Economic and Social Initiatives (3.9\%) & 25.0 & 100.0 & 100.0 & 87.5\\
United States Governance and Politics (37.3\%) & 26.3 & 67.1 & 89.5 & 90.8\\
Ideological and Territorial Expansion in the Americas (9.8\%) & 20.0 & 75.0 & 85.0 & 90.0\\
American History Eras (2.9\%) & 33.3 & 66.7 & 83.3 & 83.3\\
\bottomrule
\end{tabular}}
\caption{Multi-dimensional evaluation results for the Gemma 3 model family on MMLU High School U.S. History.}
\label{tab:llama_hs_us_hist}
\end{table*}

\section{Model List}
\label{sec:model_list}
We list all evaluated models and provide links to their open-source weights.

\begin{itemize}[left=0pt, itemsep=1pt]
  \item Qwen3: \href{https://huggingface.co/Qwen/Qwen3-0.6B-Base}{Qwen3-0.6B-Base}, \href{https://huggingface.co/Qwen/Qwen3-1.7B-Base}{Qwen3-1.7B-Base}, \href{https://huggingface.co/Qwen/Qwen3-4B-Base}{Qwen3-4B-Base}, \href{https://huggingface.co/Qwen/Qwen3-8B-Base}{Qwen3-8B-Base}, \href{https://huggingface.co/Qwen/Qwen3-14B-Base}{Qwen3-14B-Base}, \href{https://huggingface.co/Qwen/Qwen3-0.6B}{Qwen3-0.6B}, \href{https://huggingface.co/Qwen/Qwen3-1.7B}{Qwen3-1.7B}, \href{https://huggingface.co/Qwen/Qwen3-4B}{Qwen3-4B}, \href{https://huggingface.co/Qwen/Qwen3-8B}{Qwen3-8B}, \href{https://huggingface.co/Qwen/Qwen3-14B}{Qwen3-14B}.
  \item Llama 3: \href{https://huggingface.co/meta-llama/Llama-3.2-1B}{Llama-3.2-1B}, \href{https://huggingface.co/meta-llama/Llama-3.2-3B}{Llama-3.2-3B}, \href{https://huggingface.co/meta-llama/Llama-3.1-8B}{Llama-3.1-8B}, \href{https://huggingface.co/meta-llama/Llama-3.1-70B}{Llama-3.1-70B}, \href{https://huggingface.co/meta-llama/Llama-3.2-1B-Instruct}{Llama-3.2-1B-Instruct}, \href{https://huggingface.co/meta-llama/Llama-3.2-3B-Instruct}{Llama-3.2-3B-Instruct}, \href{https://huggingface.co/meta-llama/Llama-3.1-8B-Instruct}{Llama-3.1-8B-Instruct}, \href{https://huggingface.co/meta-llama/Llama-3.1-70B-Instruct}{Llama-3.1-70B-Instruct}.
  \item Olmo 2: \href{https://huggingface.co/allenai/OLMo-2-0425-1B}{OLMo-2-0425-1B}, \href{https://huggingface.co/allenai/OLMo-2-1124-7B}{OLMo-2-1124-7B}, \href{https://huggingface.co/allenai/OLMo-2-1124-13B}{OLMo-2-1124-13B}, \href{https://huggingface.co/allenai/OLMo-2-0325-32B}{OLMo-2-0325-32B}, \href{https://huggingface.co/allenai/OLMo-2-0425-1B-Instruct}{OLMo-2-0425-1B-Instruct}, \href{https://huggingface.co/allenai/OLMo-2-1124-7B-Instruct}{OLMo-2-1124-7B-Instruct}, \href{https://huggingface.co/allenai/OLMo-2-1124-13B-Instruct}{OLMo-2-1124-13B-Instruct}, \href{https://huggingface.co/allenai/OLMo-2-0325-32B-Instruct}{OLMo-2-0325-32B-Instruct}.
  \item Gemma 3: \href{https://huggingface.co/google/gemma-3-1b-pt}{gemma-3-1b-pt}, \href{https://huggingface.co/google/gemma-3-4b-pt}{gemma-3-4b-pt}, \href{https://huggingface.co/google/gemma-3-12b-pt}{gemma-3-12b-pt}, \href{https://huggingface.co/google/gemma-3-27b-pt}{gemma-3-27b-pt}, \href{https://huggingface.co/google/gemma-3-1b-it}{gemma-3-1b-it}, \href{https://huggingface.co/google/gemma-3-4b-it}{gemma-3-4b-it}, \href{https://huggingface.co/google/gemma-3-12b-it}{gemma-3-12b-it}, \href{https://huggingface.co/google/gemma-3-27b-it}{gemma-3-27b-it}.
  \item Phi-3: \href{https://huggingface.co/microsoft/Phi-3-mini-4k-instruct}{Phi-3-mini-4k-instruct}, \href{https://huggingface.co/microsoft/Phi-3-medium-4k-instruct}{Phi-3-medium-4k-instruct}.
\end{itemize}

\section{Extended Results: How does model performance change with increased \method{}?}
\label{sec:extended_balancing}

In this section, we generalize the pruning experiments from \S\ref{subsec:balancing_subsets} beyond the illustrative cases to \emph{all} model families and benchmarks in our setup. Our aim is methodological: we examine how aggregate accuracy and per-subset dispersion evolve as we progressively rebalance a benchmark. Concretely, for each (model, benchmark) pair we sweep a pruning budget (scheduled inversely to baseline \method{}), recompute \method{} and accuracy at each budget, and compare the pruned-set accuracy to the full-set accuracy using the coupled bootstrap significance test detailed in App.~\ref{sec:details_of_stat_sig}. Family-wise plots in this section visualize these trajectories, allowing us to observe whether increased \method{} coincides with stable (or shifting) aggregate scores and tighter per-subset distribution of performance.

\begin{figure*}
    \centering
    \includegraphics[width=\linewidth]{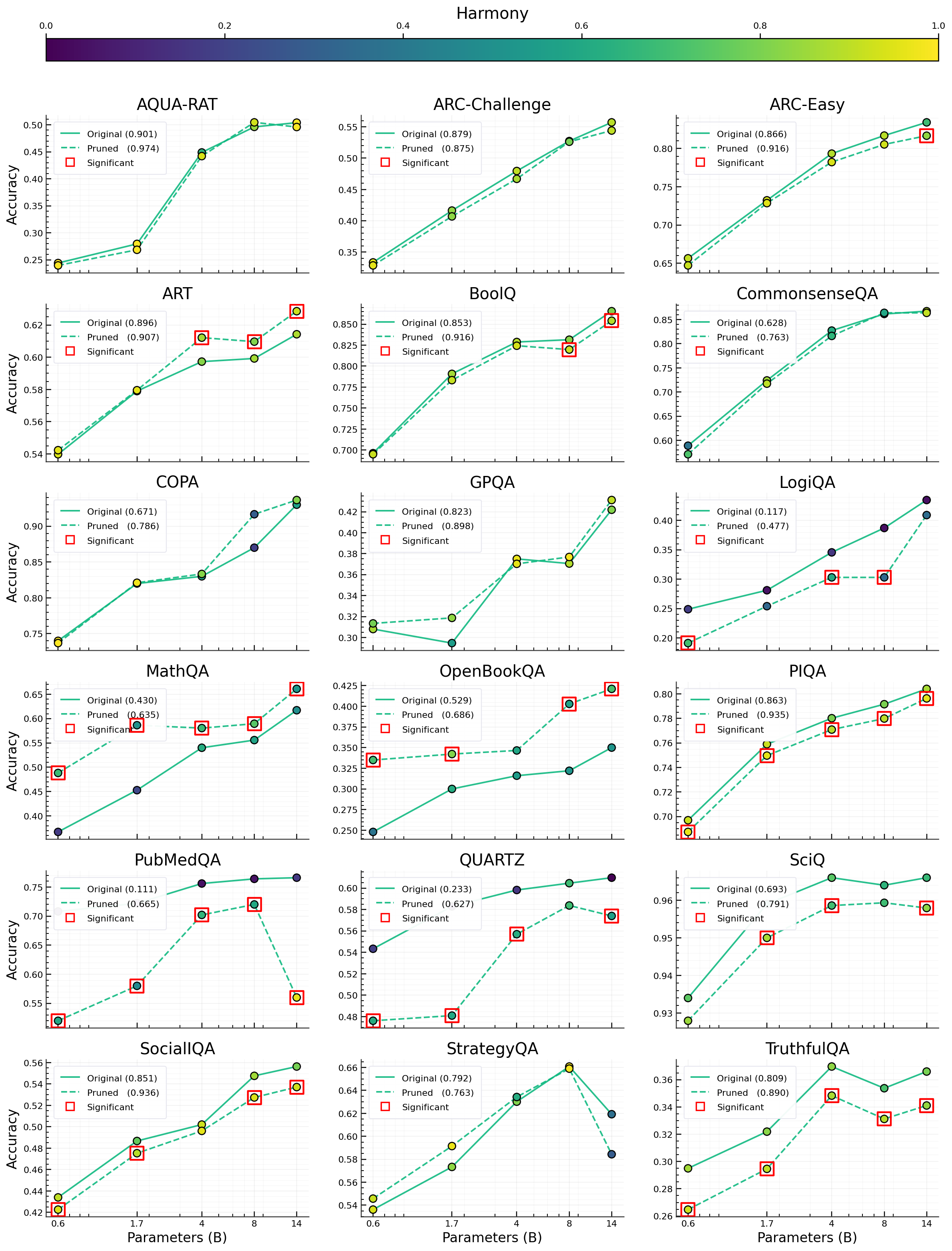}
    \caption{\textbf{Full results of balancing benchmarks via pruning in Qwen3 model family.} We remove overly similar items with a pruning rate inversely proportional to \method{}, which consistently improves \method{}. We find that aggregate scores often change statistically significantly on less harmonious benchmarks, whereas they remain more stable on more harmonious benchmarks.}
    \label{fig:qwen_pruning_base_full}
\end{figure*}

\begin{figure*}
    \centering
    \includegraphics[width=\linewidth]{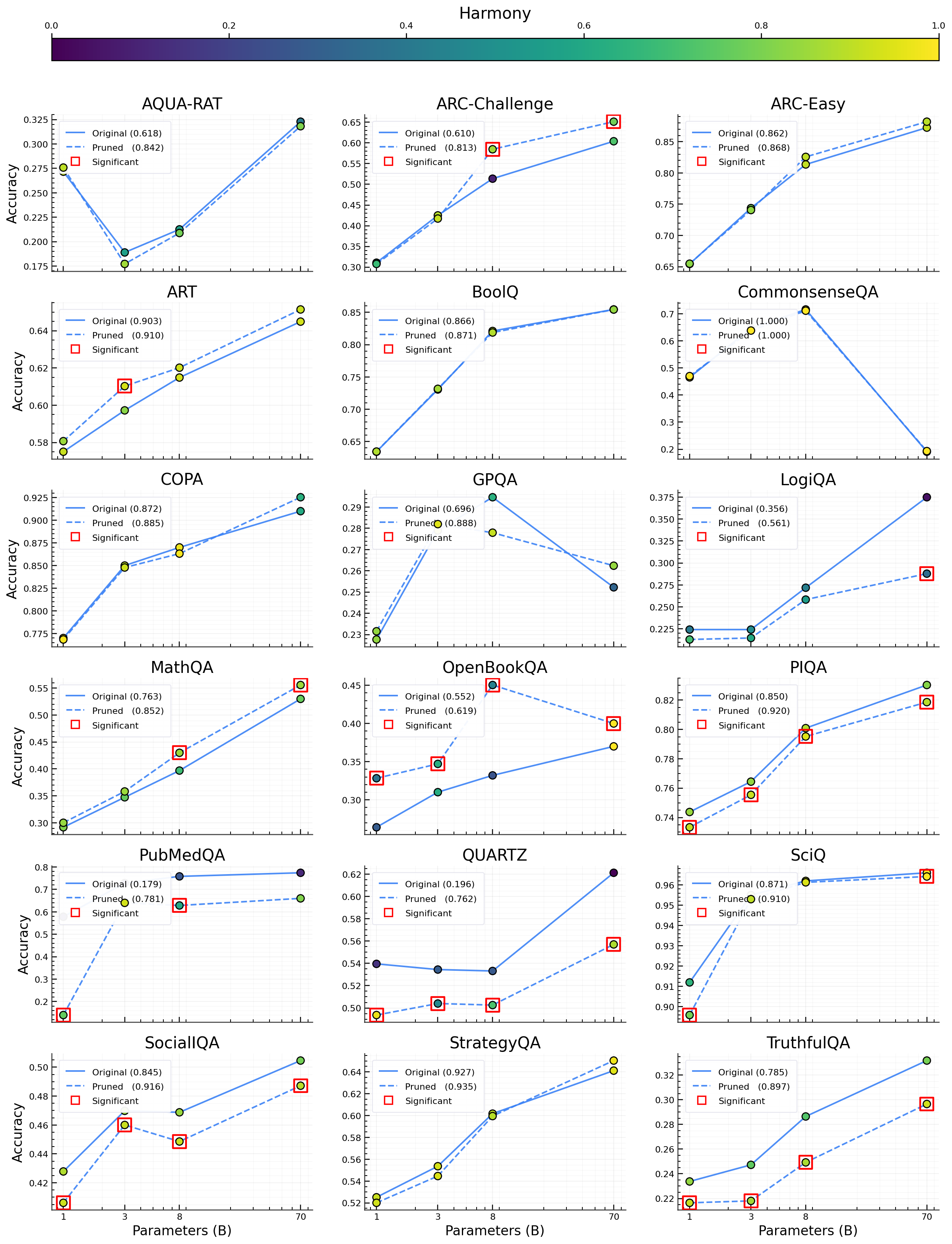}
    \caption{\textbf{Full results of balancing benchmarks via pruning in Llama 3 model family.} We remove overly similar items with a pruning rate inversely proportional to \method{}, which consistently improves \method{}. We find that aggregate scores often change statistically significantly on less harmonious benchmarks, whereas they remain more stable on more harmonious benchmarks.}
    \label{fig:llama_pruning_base_full}
\end{figure*}

\begin{figure*}
    \centering
    \includegraphics[width=\linewidth]{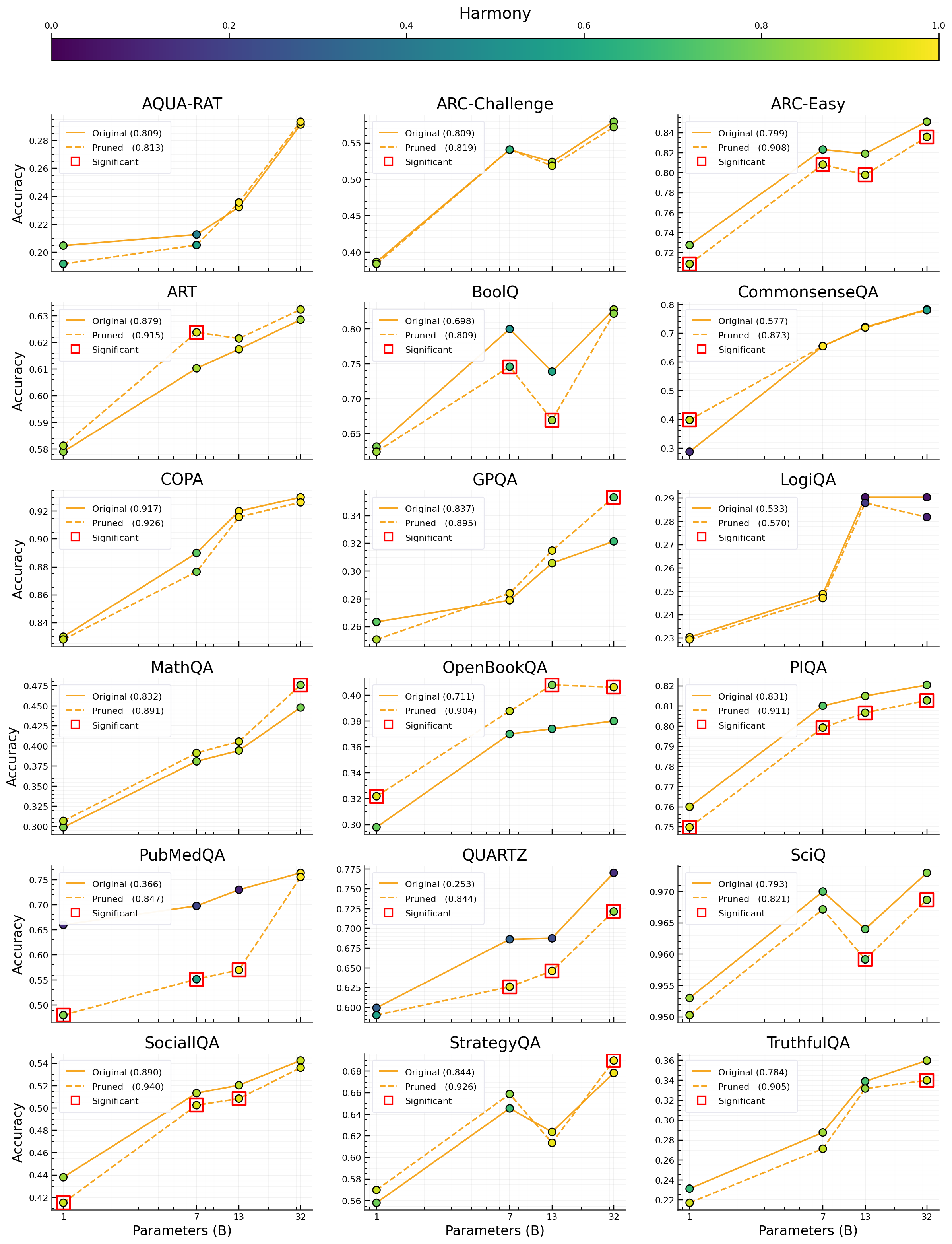}
    \caption{\textbf{Full results of balancing benchmarks via pruning in Olmo 2 model family.} We remove overly similar items with a pruning rate inversely proportional to \method{}, which consistently improves \method{}. We find that aggregate scores often change statistically significantly on less harmonious benchmarks, whereas they remain more stable on more harmonious benchmarks.}
    \label{fig:olmo_pruning_base_full}
\end{figure*}

\begin{figure*}
    \centering
    \includegraphics[width=\linewidth]{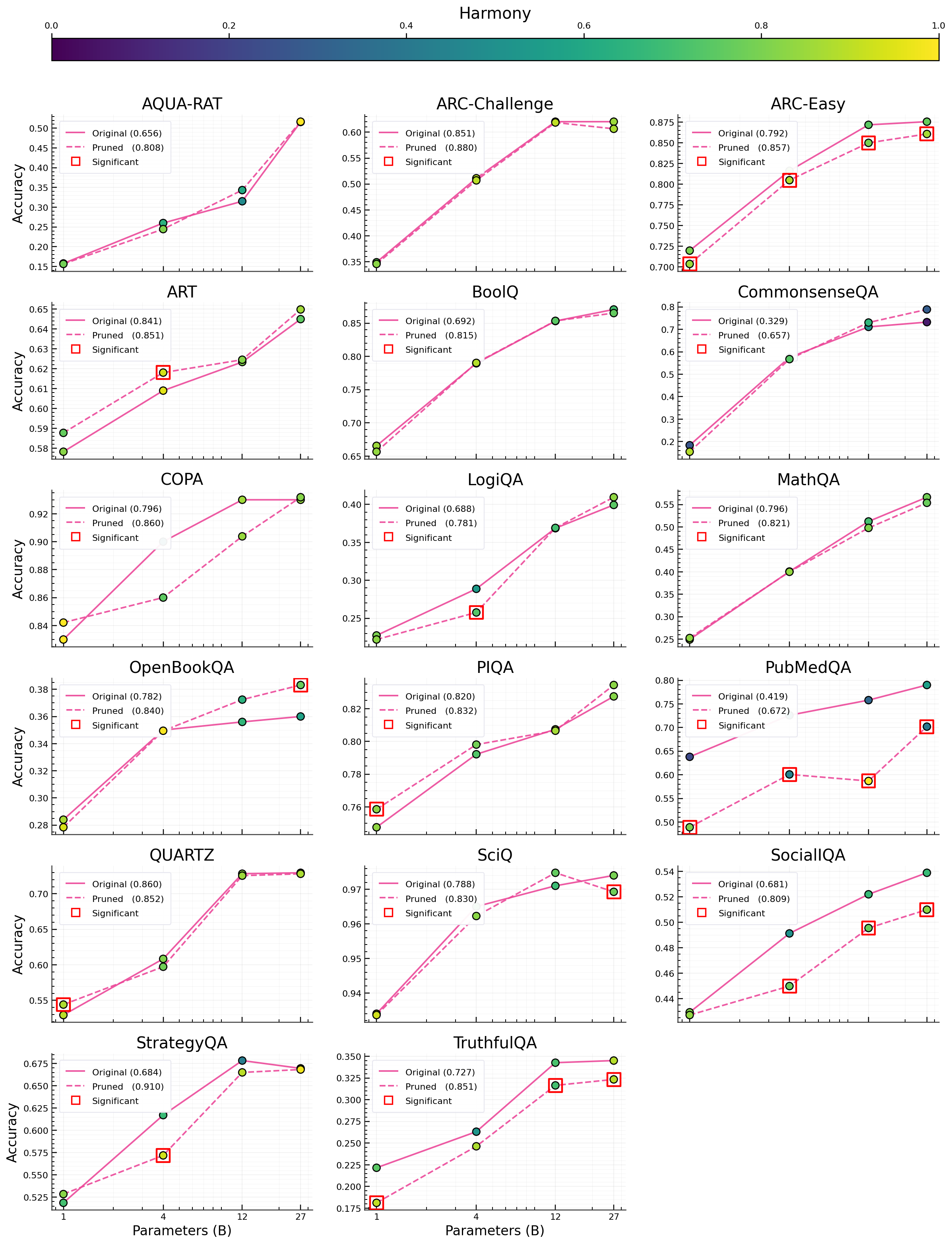}
    \caption{\textbf{Full results of balancing benchmarks via pruning in Gemma 3 model family.} We remove overly similar items with a pruning rate inversely proportional to \method{}, which consistently improves \method{}. We find that aggregate scores often change statistically significantly on less harmonious benchmarks, whereas they remain more stable on more harmonious benchmarks.}
    \label{fig:gemma_pruing_base_full}
\end{figure*}

Across all model families (Fig.~\ref{fig:qwen_pruning_base_full},~\ref{fig:llama_pruning_base_full},~\ref{fig:olmo_pruning_base_full},~\ref{fig:gemma_pruing_base_full}), two patterns consistently hold.
\textbf{(i) Accuracy shifts with increased \method.} As pruning raises \method{}, aggregate accuracy frequently changes in a statistically significant manner (App.~\ref{sec:details_of_stat_sig}), indicating that low \method{} composition can result in a misleading aggregate score.
\textbf{(ii) Low \method{} benchmarks are fragile.} Benchmarks starting with lower \method{} exhibit more instances of significant accuracy change under the pruning procedure than high \method{} benchmarks, underscoring their susceptibility to presenting misleading aggregate scores.

\section{Additional Related Work}
\label{sec:add_rel_work}

\paragraph{Language Model Evaluation.} Reliable evaluation is essential for accurately assessing model capabilities and enabling fair comparisons, which in turn informs future developments. Hence, there has been a surge in the development of benchmarks designed to test various model capabilities, such as reasoning \citep{bisk2019piqareasoningphysicalcommonsense, sap2019socialiqacommonsensereasoningsocial, zellers2019hellaswagmachinereallyfinish, liu2020logiqachallengedatasetmachine}, world knowledge \citep{mihaylov2018suitarmorconductelectricity, hendrycks2020measuring, srivastava2023imitationgamequantifyingextrapolating}, and truthfulness \citep{lin2022truthfulqameasuringmodelsmimic, khatun2024truthevaldatasetevaluatellm}. Beyond individual benchmarks, holistic frameworks have emerged to offer a more comprehensive assessment of model performance \citep{liang2023holisticevaluationlanguagemodels, chiang2024chatbot, eval-harness, lighteval, srivastava2023imitationgamequantifyingextrapolating}. Reciprocally, understanding and improving current benchmarks have been equally important. MMLU Pro \citep{wang2024mmluprorobustchallengingmultitask} and Big-Bench-Hard \citep{suzgun2022challenging} address benchmark saturation by constructing more challenging variants of MMLU \citep{hendrycks2020measuring} and Big-Bench \citep{srivastava2023beyond} respectively. As top models approach ceiling effects on narrow probes, evaluation has shifted toward complex end-to-end tasks and composite suites. HLE and ARC-AGI assess multi-step reasoning, tool use, and robustness across domains \citep{hle, arcagi}. Execution-grounded tasks such as SWE-bench measure real-world software problems and end-to-end correctness \citep{swebench}. Competitive exams like AIME and IMO, and professional exams such as the bar, push systems toward expert-level competence. Another recent practice is evaluation with online leaderboards, which use hidden test sets, fixed prompts, and compute disclosures in order to support fair comparison and consistent progress tracking \citep{chiang2024chatbot}. Yet, these advances rest on a common premise that benchmarks reliably evaluate models on their stated domains. We audit this premise by testing whether benchmarks provide balanced coverage and promote comparable performance across subdomains.

\section{Hierarchical Labeling for Multi-Dimensional Evaluation}
\label{sec:hier_labeling}

We build a tree benchmark over questions, then assign concise, human-readable labels to every node. Leaves summarize the shared evaluation focus of their questions, while internal nodes summarize their children.

To build the tree, we recursively induce partitions as discussed in \S\ref{subsec:methodology}, starting from the root (i.e., the entire benchmark) and ending at leaves (i.e., the clusters that do not admit a valid partition). For labeling leaves, we gather brief question annotations within a leaf and ask a model for one specific noun-phrase label. For labeling internal nodes of the tree, we pass the child labels to the model and ask for a slightly more abstract label that still captures the shared theme. Therefore, this procedure yields a bottom-up label propagation from leaves to internal nodes then to the root. We use \texttt{Gemini-2.0-flash} to annotate individual questions, assign each leaf a label from its question annotations, and propagate labels upward by aggregating child labels.

\section{Example Clusters from Multi-Dimensional Evaluation}
\label{app:example_clusters}
We present qualitative examples of the clusters produced by our hierarchical spectral clustering method, which groups questions using predictive similarity. Tables \ref{tab:arc-easy} and \ref{tab:boolq} illustrate representative clusters for ARC-Easy and BoolQ, respectively. Notably, successful hierarchical clustering requires coherence at each depth level, and these examples show that our method produces consistently coherent clusters even at lower depths, which is a particularly stringent and informative criterion.
\begin{table*}[ht]
\centering
\begin{tabular}{@{}cclp{0.6\textwidth}@{}}
\toprule
\textbf{ID} & \textbf{Depth} & \textbf{Topic} & \textbf{Questions} \\
\midrule
2139 & 5 & Reproductive Systems & 
1) Which is the greatest benefit of sexual reproduction?\\
&&& 2) Which of these is the best example of sexual reproduction?\\
&&& 3) Which is an example of asexual reproduction?\\
&&& 4) Which statement describes a characteristic of both sexual and asexual reproduction? \\[4pt]
\midrule
1244 & 5 & Photosynthesis & 
1) In what form do plants store the energy produced from sunlight?\\
&&& 2) Which of the following best describes how plants use the energy they receive from sunlight?\\
&&& 3) Which of the following processes makes it possible for plants to use energy from sunlight to produce their own food?\\
&&& 4) Which form of energy do plants need to capture in order to perform photosynthesis?\\
&&& 5) What is the green pigment that allows plants to change the Sun's energy into chemical energy?\\
&&& 6) Which of the following do plants need to make their own food? \\[4pt]
\midrule
163 & 3 & Waves & 
1) What causes sound?\\
&&& 2) A substance that carries sound waves is called\\
&&& 3) An echo is a sound wave that has been\\
&&& 4) Sound travels as a \\[4pt]
\midrule
131 & 2 & Landform Processes & 
1) A beach is formed when sediment is deposited along a shoreline. What would most likely happen if rivers that empty into the ocean were dammed?\\
&&& 2) As a river enters a larger body of water, sediments are deposited over a wide area. Which of these landforms is likely to be formed at the site of deposition?\\
&&& 3) Deposition of sediment will most likely form a\\
&&& 4) Which statement describes the formation of a delta?\\
&&& 5) A delta at the mouth of a river is the direct result of\\
&&& 6) Rivers leave behind small pieces of rocks after they flood. These small pieces of rock form a floodplain. Which words best describe a floodplain?\\
&&& 7) Rivers and streams can carry sediments long distances before they are deposited. What is formed when sediments are deposited at the mouth of a river?\\
&&& 8) Sediment that is deposited on a beach may come from a local source or be transported by which action?\\
&&& 9) The shape of a riverbed changes over time as a result of which gradual process?\\[4pt]
\midrule
2206 & 3 & Laboratory Equipment & 1) Which tool is best used to observe a soil sample?\\
&&& 2) Which tool would be most useful for observing the details of an insect's wings?\\
&&& 3) Which tools are used to determine the boiling point of water?\\
&&& 4) Which device is used to determine the volume of a liquid?\\
&&& 5) Which tool should be used to measure the stem length of a plant?\\
&&& 6) Which tool is used to observe the cell wall of a leaf?\\
&&& 7) Which tool is best to use when comparing an animal cell to a plant cell?\\
&&& 8) Which object is best seen with a microscope? \\
\bottomrule
\end{tabular}
\caption{Example ARC-Easy clusters with their associated ID, depth, and topic.}
\label{tab:arc-easy}
\end{table*}

\begin{table*}[ht]
\centering
\begin{tabular}{@{}cclp{0.6\textwidth}@{}}
\toprule
\textbf{ID} & \textbf{Depth} & \textbf{Topic} & \textbf{Questions} \\
\midrule
2818 & 2 & Locations & 
1) is st augustine the oldest city in florida\\
&&& 2) is san juan puerto rico in the caribbean\\
&&& 3) is st augustine florida the oldest city in america\\
&&& 4) is there an amtrak station in pensacola florida\\
&&& 5) is key west part of the united states\\
&&& 6) is riviera maya on the gulf of mexico\\
&&& 7) is panama city beach on the gulf of mexico \\[4pt]
\midrule
3026 & 2 & Species Hybridization & 
1) is artificial selection and selective breeding the same thing\\
&&& 2) can a grizzly bear mate with a polar bear\\
&&& 3) can a horse and a donkey have a baby\\
&&& 4) can a polar bear and a grizzly mate\\
&&& 5) is it possible for a zebra and a horse to interbreed \\[4pt]
\midrule
1751 & 3 & Sports & 
1) did kaka play in the 2002 world cup\\
&&& 2) is the uefa champions league final one game\\
&&& 3) did alphonse areola play in world cup 2018\\
&&& 4) has a keeper ever won the ballon d'or\\
&&& 5) has christiano ronaldo ever won the world cup \\[4pt]
\midrule
1149 & 3 & Breaking Bad & 
1) will there be a season six of breaking bad\\
&&& 2) is better call saul set after breaking bad\\
&&& 3) was better call saul filmed before breaking bad\\
&&& 4) is there a spin off from breaking bad \\[4pt]
\midrule
497 & 2 & Constitutional Rights & 
1) is the first amendment in the bill of rights\\
&&& 2) was the right to bear arms in the original constitution\\
&&& 3) does the us constitution protect the right to privacy\\
&&& 4) do corporations have the same free speech rights as persons\\
&&& 5) are fighting words protected under freedom of speech\\
&&& 6) does the first amendment separate church and state\\
&&& 7) does the right to privacy exist in the constitution\\
&&& 8) is there a limit on freedom of speech\\
&&& 9) is the second amendment part of the constitution \\
\bottomrule
\end{tabular}
\caption{Example BoolQ clusters with their associated ID, depth, and topic.}
\label{tab:boolq}
\end{table*}

\section{The Use of Large Language Models (LLMs)}
\label{sec:use_of_llms}

We use large language models (LLMs) in two limited ways in this work. First, we use LLMs for light polishing of author-written text, including grammar, wording, and clarity edits. Second, we leverage LLMs to generate candidate synthetic data for RedundantQA. In particular, as described in detail in Appendix~\ref{sec:redundantqa}, we begin with author-written seed examples and query \texttt{Gemini-2.0-flash} to expand them into candidate question sets, followed by automated consistency checks and manual review before inclusion in the final dataset.

All substantive content, methodological decisions, validation procedures, results, and conclusions are authored and verified by the authors, who take full responsibility for the paper's contents, including the synthetic data construction process and any text edited with LLM assistance.

\end{document}